\RequirePackage{fix-cm}
\documentclass[twocolumn]{svjour3}          % twocolumn
\smartqed  % flush right qed marks, e.g. at end of proof

\usepackage{stmaryrd}
\usepackage{url}
\usepackage{color}
\usepackage[export]{adjustbox}
\usepackage{xspace}
\usepackage{multirow}
\usepackage{dingbat}
\usepackage{csquotes}
\usepackage[table]{xcolor}
\usepackage{epigraph}
\usepackage{amsmath}

\usepackage{xspace}

\newcommand{\latinphrase}[1]{\textit{#1}} 
\newcommand{\etal}{\latinphrase{et~al.}\xspace}
\newcommand{\ie}{\latinphrase{i.e.}\xspace}

\newcommand{\eg}{\latinphrase{e.g.}\xspace}
\newcommand{\etc}{\latinphrase{etc.}\xspace}

\newcommand{\ch}{\checkmark}
\def\I{\mathbf{I}}
\def\U{\mathbf{U}}
\def\T{T}
\def\E{E}
\def\N{N}

\newcolumntype{C}[1]{>{\centering\arraybackslash}p{#1}}

\begin{document}

\title{Diving Deeper into Underwater Image Enhancement: A Survey}

\author{Saeed Anwar$^*$ \and Chongyi Li$^*$}


\institute{	  S. Anwar is a research fellow\at
			  Data61, CSIRO, ACT 2601, AU \\
              Australian National University, Canberra ACT 2600, \\
              \email{saeed.anwar@data61.csiro.au}           
              %  \\
              % \emph{Present address:} of F. Author  %  if needed
           \and
           	  C. Li is a postdoctoral fellow \at
              Department of Computer Science,\\
              City University of Hong Kong, Kowloon, Hong Kong, China \\
              \email{lichongyi@tju.edu.cn}\\
              $^*$ shows equal contribution.
}
\date{Received: date / Accepted: date}

\maketitle

\begin{abstract}
%\CL{ Revision almost done.}

%\CL{We should adjust the order of references starting with [1].}

%\CL{The table 1 is a little over-length. Also, we should make it clear about the loss presented in Table. I mean what is l_w, l_perceptual, l_r, etc.}

%\CL{If possible, we should adjust the layout of Fig.3.}

The powerful representation capacity of deep learning has made it inevitable for the underwater image enhancement community to employ its potential. The exploration of deep underwater image enhancement networks is increasing over time, and hence; a comprehensive survey is the need of the hour. In this paper, our main aim is two-fold, 1): to provide a comprehensive and in-depth survey of the deep learning-based underwater image enhancement, which covers various perspectives ranging from algorithms to open issues, and 2): to conduct a qualitative and quantitative comparison of the deep algorithms on diverse datasets to serve as a benchmark, which has been barely explored before. 

To be specific, we first introduce the underwater image formation models, which are the base of training data synthesis and design of deep networks, and also helpful for understanding the process of underwater image degradation. Then, we review deep underwater image enhancement algorithms, and a glimpse of some of the aspects of the current networks is presented including network architecture, network parameters, training data, loss function, and training configurations. We also summarize the evaluation metrics and underwater image datasets. Following that, a systematically experimental comparison is carried out to analyze the robustness and effectiveness of deep algorithms. Meanwhile, we point out the shortcomings of current benchmark datasets and evaluation metrics. Finally, we discuss several unsolved open issues and suggest possible research directions. We hope that all efforts done in this paper might serve as a comprehensive reference for future research and call for the development of deep learning-based underwater image enhancement.

%Moreover, our survey shows that despite the fast pace developments in deep learning, underwater image enhancement is lagging due to insufficient real-data available for training and the contribution of certainty of the evaluation metrics. The dataset for training are either synthetic or contains an inadequate number of images and visual diversity.} 

%\SA{We present the traditional evaluation metrics and their shortcomings for underwater image enhancement. We then provide details of the underwater specific metrics. Furthermore, a glimpse of some of the aspects of the current networks is also presented. A comparison of the algorithm is performed quantitatively and qualitatively, and the article is concluded with a list of promising research directions.}%\url{https://github.com/saeed-anwar/SRsurvey}
\end{abstract}

\section{Introduction}

\setlength{\epigraphwidth}{23em}
\epigraph{\it `Sit, be still, and listen.'}{Rumi}

Nowadays, developing, exploring, and protecting the ocean's resources have become the strategy center in the international community. Clear underwater images and videos can provide valuable information of the underwater world, which are essential for numerous engineering and research tasks such as underwater archaeology, underwater surveillance, \etc However, the raw underwater images and videos usually suffer from the effects of quality degradation, especially the impact of backscatter in far distances. The issues of quality degradation are mainly introduced by light selective absorption and scattering in  water as well as the use of artificial light in deep water. The degraded underwater images have low contrast and brightness, color deviations,  blurry details, and uneven bright speck, which limit their applications in practical scenarios. As an indispensable processing step, underwater image enhancement methods ranging from the conventional techniques (\eg, physical model-based methods, and histogram equalization-based methods) to the data-driven techniques (\eg, convolutional neural networks, and generative adversarial networks) have been attracting increasing attention.

The past few decades have seen the rapid development of deep learning techniques, which have been extensively applied in various computer vision and image processing tasks \cite{DL}. Deep learning has significantly improved the performance of high-level vision tasks such as object detection \cite{Ren2017} and object recognition \cite{RedidualNet}. Moreover, the low-level vision tasks, such as image super-resolution \cite{Li2018} and image denoising \cite{Zhang2017}, also benefit from the advantages of deep networks and deliver state-of-the-art performance. Unfortunately, we are unable to observe the appealing performance of deep learning-based underwater image enhancement, although lots of researchers have attempted to utilize the deep learning techniques to the underwater image enhancement.

In this paper, we mainly focus on deep learning method, which enhance and restore underwater images. Through this exposition, we provide the latest development and comparison of current deep underwater image restoration and enhancement algorithms. Furthermore, we summarize the existing issues, analyze the potential reasons, and suggest future research directions. The main contributions of this paper are two-fold:
\begin{itemize}
 
\item We summarize the deep learning-based underwater image enhancement algorithms, including network architectures, network parameters, training data, loss function, and training configurations. It provides, to the best of our knowledge, the first comprehensive and in-depth survey for deep learning-based underwater image enhancement, which is helpful for developing more robust and effective deep algorithms. 

\item We conduct the systematic experiments on diverse datasets to qualitatively and quantitatively compare the deep learning-based underwater image enhancement algorithms. Our evaluation and analysis demonstrate the performance of current deep algorithms, point out their limitations, and indicates the bias of existing benchmark datasets and evaluation metrics. As a consequence, we give potential insights for future research directions in this field of study.

\end{itemize}

The rest of the paper is organized as follows. Section 2 introduces the background of underwater image enhancement and restoration, mainly focusing on the imaging models. Section 3 presents the existing deep learning-based underwater image enhancement algorithms and insights into the network. Section 4 gives the experimental quantitative and qualitative results and analysis, evaluation metrics, and datasets. Section 5 suggests future research directions, and Section 6 concludes this paper.   

\section{Background}
In this section, we mainly introduce the commonly-used physical models for underwater image enhancement, including atmospheric scattering model, simplified underwater image formation model, and revised underwater image formation model. These models are the base of training data synthesis and design of deep networks and also helpful for understanding the process of underwater image degradation.

\subsection{Atmospheric Scattering Model}
For an image captured in a scattering medium, only a part of the reflected light from the scene reaches the imaging sensor due to the absorption and scattering effects, typically for hazy image formation. Since underwater images usually have a hazy appearance (similar to the hazy image), the atmospheric scattering model~\cite{Koschmieder1924} is traditionally used to describe the degradation of the underwater image. The atmospheric scattering model~\cite{Koschmieder1924} can be characterized as: 
\begin{equation} 
\label{equ_1} 
\U (x)=\I(x)\T(x)+B(1-\T(x)), 
\end{equation} 
where $x$ denotes the pixel coordinates, $\U(x)$ is the observed image, $\I(x)$ is the haze-free latent image, $B$ is the global atmospheric light which indicates the intensity of ambient light, and $\T(x) \in [0,1]$ is the transmission which represents the percentage of the scene radiance reaching the camera. When the haze is homogenous, $\T(x)$ can be further expressed in an exponential decay term as: 
\begin{equation} 
\label{equ_2} 
\T(x)=\exp(-\beta d(x)), 
\end{equation} 
where $\beta$ is the atmospheric attenuation coefficient and $d(x)$ is the distance from the scene to the camera. In this atmospheric scattering model, the scattering is non-selective, and attenuation is independent of wavelengths.  

\subsection{Simplified Model} 
In fact, there is a significant difference between atmospheric scattering model and real-world underwater image formation model. The real-world underwater imaging is far more complicated due to the optical properties of selective attenuation in water. Thus, in the early stage, most physical model-based methods followed a simplified underwater image formulation model provided by~\cite{Chiang122012}. We denote the captured underwater image by $\U_\lambda(x)$, the clear latent image (also known as scene radiance) as $\I_\lambda(x)$, and the homogeneous global background light as $B_\lambda$, then the degradation model is given as: 

\begin{equation} 
\U_\lambda(x)=\I_\lambda(x)\cdot \T_\lambda(x) + B_\lambda \cdot \big( 1 - \T_\lambda(x) \big), 
\label{eq:UW_synthesis} 
\end{equation} 
where $\lambda$ presents the wavelength of the RGB channels, and $x$ is a point in the underwater scene. Similarly, $\T_\lambda(x)$ is the medium energy ratio, which is the percentage of the scene radiance captured by the camera (the amount of radiance reflected from the point $x$). This phenomenon causes contrast degradation, and color casts. To be precise, $\T_\lambda(x)$ is a function of $\lambda$ and the distance $d(x)$ to the camera from the scene point $x$, expressed as:

\begin{equation} 
\T_{\lambda}(x) = 10^{-\beta_\lambda d(x)} = \frac{\E_\lambda \big( x,d(x) \big)}{\E_\lambda(x, 0)} = \N_\lambda \big( d(x) \big), 
\label{eq:UW_synthesis1} 
\end{equation} 
where $\beta_\lambda$ is the medium attenuation coefficient, which is dependent on the wavelength. Furthermore, $\E_\lambda(x,0)$ is the energy of light from the submerged scene before it passes through the transmission medium from a distance $d(x)$ while $\E_\lambda \big(x,d(x) \big)$ is the strength of light after absorption by the transmission medium. Moreover, $\N_\lambda$ is the normalized residual energy which is the ratio of residual energy to the initial energy per unit of distance and is dependent on the wavelength of light. For example, the bluish tone of the most underwater images is due to the fast attenuation of the red wavelength in open water as it possesses a longer wavelength than blue and green ones.  

\subsection{Revised Model} 
Recent research found that the commonly-used atmospheric scattering model and simplified underwater image formation model ignored some key components in the process of real-world underwater imaging \cite{RevisedModel}. Specifically, the attenuation coefficient for backscatter strongly depends on the veiling light. Moreover, unlike the absorption in the atmosphere, the absorption in water should not be neglected. Most importantly, the attenuation coefficients for the direct signal and the scattering signal are different. 

Based on the findings mentioned above, Akkaynak \& Treibitz  \cite{RevisedModel} proposed a revised underwater image formation model which can be expressed as:
%\cdot \T_\lambda(x) + B_\lambda \cdot \big( 1 - \T_\lambda(x) \big),
\begin{equation}
\U_\lambda(x)=\I_\lambda(x)e^{-\beta_\lambda^{D}(\mathbf{v}_{D})\cdot z}+B_\lambda^\infty \big(1-e^{-\beta_\lambda^{B}(\mathbf{v}_{B})\cdot z}\big),
\label{eq:UW_revise}
\end{equation}
where $B_\lambda^\infty$ is the veiling light, $\beta_\lambda$ is the beam attenuation coefficient, $D$ is the direct transmitted light, $B$ is the backscattered light, the vectors $\mathbf{v}_{d(x)}$ and $\mathbf{v}_{b(x)}$ represent the coefficient dependencies. To be more specific, $\mathbf{v}_{d(x)}$=$\{$z, $\rho$, E, S$_{\lambda}$, $\beta\}$ and $\mathbf{v}_{b(x)}$=$\{$E, S$_{\lambda}$, b, $\beta$ $\}$, where $z$ is the range along LOS, $\rho$ is the reflectance, $E$ is the irradiance, S$_{\lambda}$ is the sensor spectral response, and $b$ is the beam scattering coefficient. Similar to the simplified model, $\U_\lambda(x)$ is the observed underwater image, $\I_\lambda(x)$ is the latent clear underwater image. More details can be found in \cite{RevisedModel}. Moreover, the coefficient associated with the backscatter varies with the sensor, ambient illumination, and water type. Generally, the coefficient of backscatter is different from the coefficient associated with the direct signal.  

In summary, the atmospheric scattering model is suitable for underwater scenarios only in some cases, such as shallow water with low backscatter. Compared to the atmospheric scattering model, the simplified underwater image formation model takes the selective attenuation of different wavelengths into consideration, which extends the generalization of this model. However, the simplified underwater image formation model assumes the attenuation coefficients are only properties of the water, which is inaccurate because the attenuation coefficients vary with the sensor, ambient illumination, \etc Besides, the simplified model ignores the fact that the backscattered light has a different attenuation coefficient from the direct light. Thus, a physically accurate model (\ie, revised underwater image formation model) is proposed, which further completes the model of underwater image formation. Nevertheless, such an accurate model has barely received much attention due to its complexity. Most of the deep learning-based underwater image enhancement algorithms still follow the atmospheric scattering model or simplified underwater image formation model to synthesize their training data and design their network architectures. Inaccurate models tend to happen in unreliable, unstable, and inauthentic results of deep algorithms.   

\section{Deep Underwater Image Enhancement Algorithms}
\begin{figure*}
\centering
\includegraphics[width=\textwidth, clip,trim=1cm 7.5cm 1cm 0]{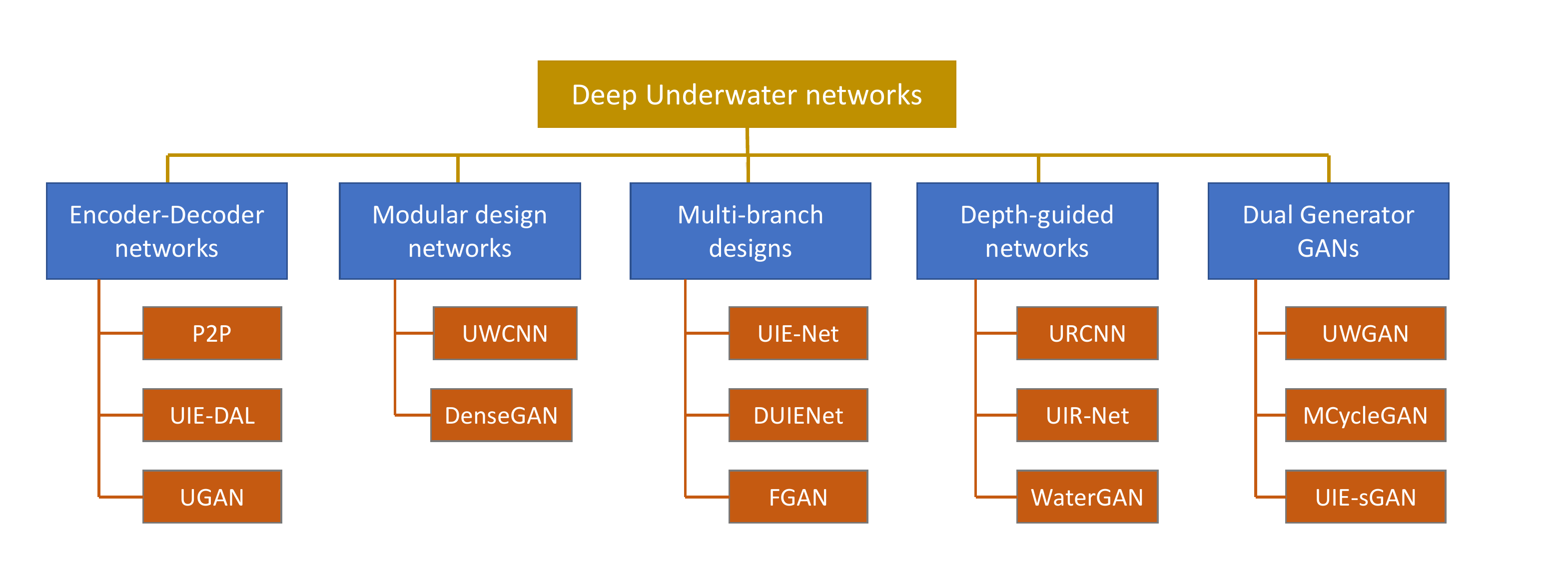}
\caption{\textbf{Categorization of Deep Underwater Networks:} The organization of deep networks based on their essential aspects.}
\label{fig:classify_archs}
\end{figure*}

Deep underwater image enhancement algorithms can ideally be divided into two main categories \ie, CNN-based and GAN-based algorithms. The goal of the CNN algorithms is to be faithful to the original underwater image while the GAN-based algorithms aim to improve the perceptual quality of the images. However, this classification is very naive; therefore, we categorize the networks based on their architectural differences. In Figure~\ref{fig:classify_archs}, the categorization of deep underwater networks is presented, and in the following sections, we list and provide details for each method into different categories based on essential aspects.  

%%%%%%%%%%%%%%%%%%%%%%%%%%%%%%%%%%%%%%%%%%%% Networks %%%%%%%%%%%%%%%%%%%%%%%

%%%%%%%%%%%%%%%%%%%%%%%%%%%%%%%%%%%%%%%% Encoder-decoder %%%%%%%%%%%%%%%%%%%%%%%%%%%%%%%%%%%%%%%%%%%%
\subsection{Encoder-Decoder models}
The following models benefit from the famous encoder-decoder architecture to advance the underwater image enhancement research.

\subsubsection{P2P network}
Recently, Sun~\etal~\cite{sun2018p2p} suggested the use of pixel-to-pixel (P2P) network to enhance underwater images. The proposed model is a \enquote{symmetric} encoder and decoder network similar to REDNet~\cite{mao2016REDNet}. The encoder part is composed of three convolutional layers, while the decoder is made from three deconvolutional layers. ReLU follows each network element except the last one.

This model is trained on 3359 images collected from the real-world environment. To simulate the underwater images, the authors pour milk of 30, 50, and 70 ml into 1m$^{3}$ of water to produce low, medium and high-level degradation, respectively.  Finally, out of these, 10,000 images are selected for training and another 2,000 images for testing. Moreover, the input to the network is a cropped patch of 66$\times$66. The loss function is $\ell_2$ minimized via SGD~\cite{lecun1998SGD} with an initial learning rate of 10$^{-7}$. 

\subsubsection{UIE-DAL}
Underwater Image Enhancement using Domain Adversarial
Learning (UIE-DAL)~\cite{uplavikar2019UIE-DAL} aims to learn agnostic model where it can enhance any underwater-type image. The backbone architecture of the UIE-DAL~\cite{uplavikar2019UIE-DAL} is the famous encoder-decoder UNET~\cite{ronneberger2015unet}.  The novelty of this work is the incorporation of a neural network classifier, named nuisance classifier, which classifies the latent vector extracted from the encoder.  

The authors claim the model to be agnostic considering that nuisance classifier is not aware of the underwater type as it receives the latent vector from the encoder, which is agnostic to the features of the underwater types. The UIE-DAL~\cite{uplavikar2019UIE-DAL} combines three losses \ie $\ell_2$,  nuisance loss, and adversarial loss. The training is achieved in two steps. First, the only encoder-decoder structure is trained, then a nuisance classier is incorporated in the network.

\subsubsection{UGAN}
Recently, Underwater Generative adversarial network (UGAN) \cite{fabbri2018UGAN} is proposed to improve the underwater image quality. For discriminator, UGAN chose WGAN-GP (Wasserstein GAN with gradient penalty)~\cite{gulrajani2017WGANGP} to enforces the soft constraint on the output concerning its input via the Lipschitz on the gradients norms instead of clipping the gradients in some range. The discriminator is fully convolutional and is similar to~\cite{radford2015unsupervised} except batch normalization~\cite{ioffe2015BN} is not applied to the weights of convolutional layers. Furthermore, the discriminator outputs 32$\times$32 feature matrix similar to PatchGAN \cite{PatchGAN2016}. The generator is motivated by CycleGAN~\cite{CycleGAN2017}, comparable to the encoder-decoder network of UNET \cite{ronneberger2015unet}. The encoder of UGAN~\cite{fabbri2018UGAN} is composed of convolutional layers having filter sizes of 4 $\times$ 4 with a stride of two followed by batch normalization~\cite{ioffe2015BN} and leaky ReLU (slope of 0.2). Similarly, the decoder portion consists of deconvolutional layers followed by ReLU~\cite{nair2010ReLU} only except the last layer where TanH is used to restrict distribution between -1 and 1. 

The evaluation and training are achieved on the subsets of ImageNet~\cite{deng2009imagenet}. Moreover, two types of underwater images are collected \ie one set of 6,143 images without distortion and another set of 1,817 images with distortion. The Adam~\cite{kingma2014adam} is used as optimizer with a fixed learning rate of 10$^{-4}$ for 100 epochs. The input to the network is 256$\times$256$\times$3, while loss is a linear combination of $\ell_1$ and Earth-Mover or Wasserstein-1 distance.  
 
%%%%%%%%%%%%%%%%%%%%%%%%%%%%%%%%%%%%%%%%%%%%% Block designs %%%%%%%%%%%%%%%%%%%%%%%%%%%%%%%%%%%%%%%%%%%%%%%
\subsection{Modular designs}
Modular or block designs employ the repetition of the same structure, commonly known as a \enquote{block} or a\enquote{module}, to learn the features. These designs are very successful in computer vision and machine learning tasks. We provide the example of modular or block-based designs for underwater networks below.

\subsubsection{UWCNN}
To deal with the low contrast and distorted color of the degraded underwater images, Anwar~\etal~\cite{UWCNN2018} proposed a CNN underwater image enhancement model, called UWCNN. The UWCNN is an end-to-end model trained by the synthetic underwater image datasets, which includes three densely connected building blocks. Furthermore, each basic building block consists of three densely connected convolutional layers. After the three chained building blocks, a convolutional layer is used to learn the difference (residual) between the degraded underwater image and its clean counterpart. 

To train the UWCNN~\cite{UWCNN2018} model, the authors use the attenuation coefficients of different water types to synthesize various underwater image datasets according to the underwater image formation model resulting in ten types of underwater image datasets which are synthesized by using the RGB-D NYU-v2 dataset~\cite{silberman2012NYU}. These underwater image datasets simulate the open ocean water types and coastal water types ranging from the clearest to the most turbid. Finally, the authors train ten UWCNN models for the ten types of underwater images. The parameters of the UWCNN model are learned by joint optimizing the $\ell_2$ and SSIM loss functions. In the entire UWCNN, the kernel sizes and filter numbers are fixed, \ie, 3$\times3$ and 16, respectively. The learning rate is set to 2$\times$10$^{-4}$ and ADAM~\cite{kingma2014adam} is used for optimization in TensorFlow framework. 

\subsubsection{DenseGAN}
To enhance the underwater images, Guo~\etal~\cite{DenseGAN} introduced a multiscale dense block (MSDB) algorithm, namely, DenseGAN\footnote{The authors' term the model as UWGAN; however, Li~\etal~\cite{UWGAN2018} proposed a model with the same name earlier. To avoid confusion, we call it DenseGAN due to its dense connections.} which employs the use of dense connections, residual learning, and multi-scale network for underwater image enhancement.

\begin{figure*}
\centering
\includegraphics[width=\textwidth, clip,trim=0 0.5cm 0.2cm 0]{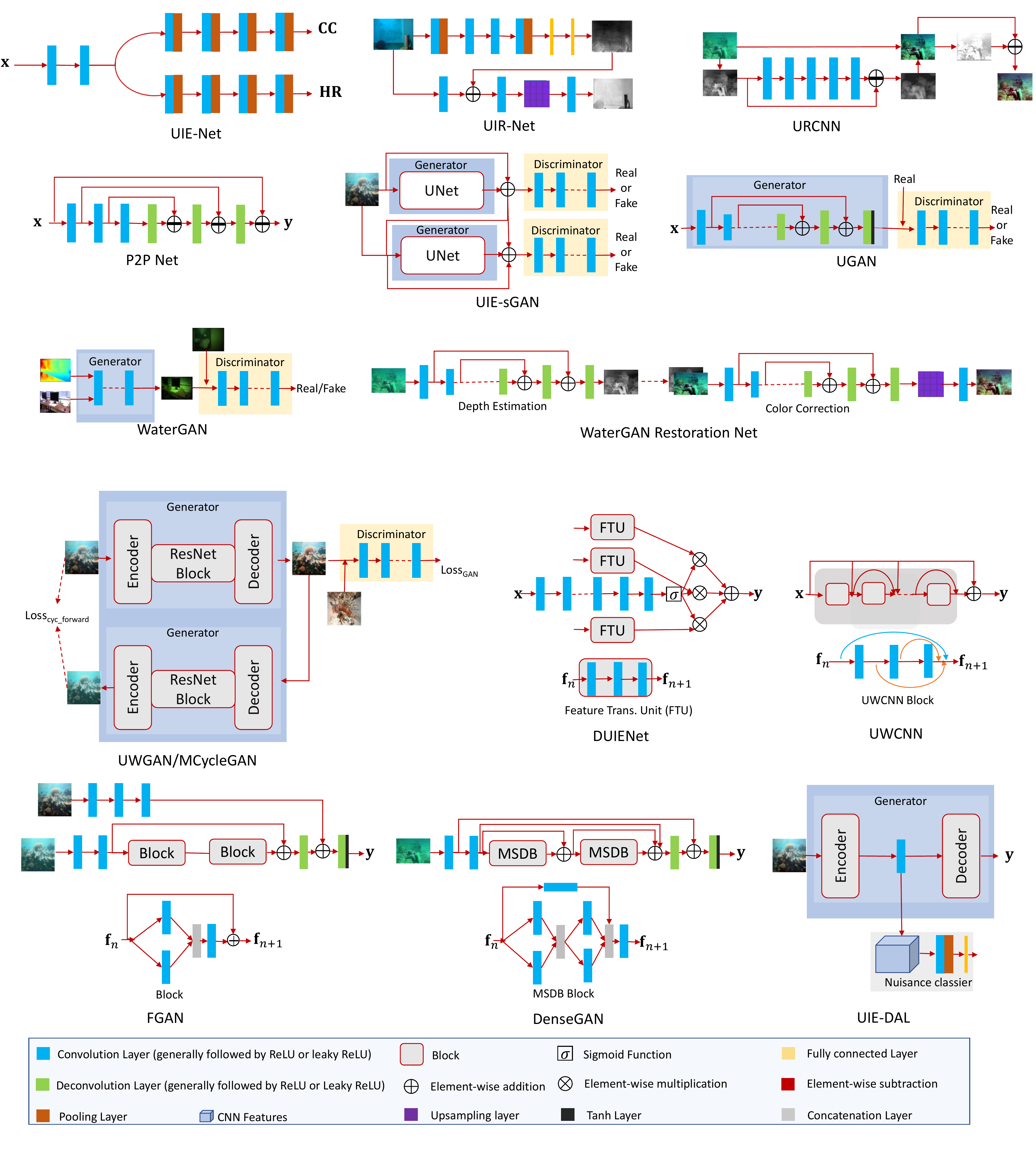}
\caption{\textbf{Network architectures:} A glimpse of network architectures used for underwater image enhancement using CNNs and GANs. \textbf{Best viewed with zoom-in on a digital display.}}
\label{fig:archs}
\end{figure*}

The generator at the start is composed of two convolutional, batch normalization (BN), leaky ReLU (LReLU) sequence then two MSDB blocks followed by sequence Deconvolutional-BN-LReLU, while at the end there is a deconvolutional layer and a TanH layer. The network architecture of the DenseGAN generator and MSDB are shown in Figure~\ref{fig:archs}. In each MSDB block, the input features are passed through two different branches, where each branch has kernels with different dilations. The features from each branch are concatenated half-way through the MSDB block and fed again into the respective branches. At the end of the MSDB block, the features are concatenated again and passed through a  1$\times$1 convolutional layer. The discriminator network is similar to PatchGAN~\cite{PatchGAN2016}; however, it is composed of five layers of spectral normalization~\cite{miyato2018spectral}. Except for the first and last layer, the discriminator is composed of sequences of convolutional-BN-LReLU. 

The first two layers of the generator have 7$\times$7 and 3$\times$3 filter size with 64 and 128 feature maps respectively. The last deconvolution layer outputs the same number of channels as the input. The TanH layer keeps the distribution between -1 and 1. Moreover, the slope of the leaky ReLU is fixed at 0.2, and the network is trained via TensorFlow framework using a learning rate of 10$^{-3}$ with patch size of 256$\times$256$\times$3. The ADAM~\cite{kingma2014adam} is used for optimization, and batch size is set to 32. The losses employed are GAN loss, $\ell_1$, and gradient loss. 

%%%%%%%%%%%%%%%%%%%%%%%%%%%%%%%%%%%%%%%%%%%% Networks %%%%%%%%%%%%%%%%%%%%%%%

%%%%%%%%%%%%%%%%%%%%%%%%%%%%%%%%%%%%%%%%%%%%% Multi-branch designs %%%%%%%%%%%%%%%%%%%%%%%%%%%%%%%%%%%%%
\subsection{Multi-branch designs}
The multiple branch designs aim to either learn different features of the same input at different levels or exploit distinct inputs at separate branches. Following are the examples of such networks.  

\subsubsection{UIE-Net}
Wang~\etal~\cite{wang2017UIENet} presented a deep CNN method for enhancement of underwater images, namely, UIE-Net, which is composed of three subnetworks. The first subnet called sharing network (termed as S-Net) is composed of convolutional layers only. S-Net extracts features from the input image which is then forwarded to the other two subnets (\ie the branches of the network: the color correction network (CC-Net) and the haze removal network (HR-Net).) CC-Net and HR-Net output color corrected image, and transmission map, respectively. Both CC-Net and HR-Net have the same network structure consisting of four convolutional layers, followed by sigmoid activation. The only difference between CC-Net and HR-Net is the number of output channels \ie three channels and one channel, respectively.

The S-Net has two convolutional layers and a consistent filter size of 5$\times$5, while the CC-Net and HR-Net have four convolutional layers with filter sizes of 1$\times$1, 3$\times$3, 5$\times$5 and 7$\times$7 to capture contextual information. Figure~\ref{fig:archs} shows the underlying network architecture of the UIE-Net. The inputs to the network are 32$\times$32 image patches in the procedure of training, and the network is trained on 2$\times$10$^{5}$ image patches synthesized from 200 clear images collected from the internet. The initial learning rate is fixed at 5$\times$10$^{-3}$, which is decreased by half after  5$\times$ 0$^{3}$ until 2.5$\times$10$^{5}$.

The loss employed for learning is $\ell_2$. Moreover, the authors perform smoothing on the input patches to obtain desirable results.  As the last step, the guided image filtering~\cite{He2010Guided} is applied on the transmission map to remove artifacts if any.  It is also to be noted here that UIE-Net is one of the pioneering work in deep learning direction.

\subsubsection{DUIENet}
More recently, Li~\etal~\cite{libenchmark2019} constructed a real-world underwater image enhancement dataset, including 950 underwater images, 890 of which have the corresponding reference images. These potential reference images are produced by 12 image enhancement methods, and the final references are selected by 50 volunteers via majority voting.

Inspired by fusion-based underwater image enhancement method~\cite{Aucti2012CVPR}, Li~\etal~\cite{libenchmark2019} proposed a gated fusion CNN trained by the constructed dataset for underwater image enhancement, called DUIENet. First, three input versions are generated by sequentially applying White Balance, Histogram Equalization, and Gamma Correction algorithms to the raw input image. Then, the DUIENet learns three confidence maps, which determine the most important features remaining in the final result. The DUIENet is a multi-scale FCNN, which consists of 14 convolutional layers followed ReLU except for the last layer (followed by Sigmoid). To reduce the color casts and artifacts introduced by the three pre-processing algorithms, three feature transformation units (FTUs) are used in the DUIENet~\cite{libenchmark2019}. The FTU includes three stacked multi-scale convolutional layers. The input of each FTU is the corresponding preprocessed underwater image, and its output is the transformed image. At last, the transformed three inputs are multiplied by the three learned confidence maps, and then the summation of the three products is the enhanced underwater image.

With the constructed dataset, the authors selected 800 pairs of images randomly to generate the training set. These images are resized to 112$\times$11 and data augmentation is used to obtain seven additional versions of the original 800 pairs of training data. The rest 90 pairs of images are treated as the testing set. To reduce the artifacts induced by pixel-wise loss functions, the authors minimize the perceptual loss (layer relu5$\_$4 of the pre-trained VGG19 network~\cite{simonyan2014vgg}).

\subsubsection{FGAN}
Fusion generative adversarial network, abbreviated as FGAN~\cite{LI2019FGAN}, takes multiple inputs and passes them through different branches in the same network. In the end, the features are summed before the loss of the generator. The architecture of FGAN~\cite{LI2019FGAN} is similar to DenseGAN with slight modifications in the block's architecture. The generator with the fundamental block structure is shown in Figure~\ref{fig:archs}. The discriminator is composed of five convolutional layers employing spectral normalization~\cite{miyato2018spectral}. The discriminator is similar to PatchGAN~\cite{PatchGAN2016}.

A batch-mode learning method with a batch size of $16$ is applied. The RGB images of size 256$\times$256 are used as inputs. Further, the learning rate is set to 10$^{-3}$. The loss function is a combination of relativistic GAN loss~\cite{jolicoeur2018relativistic}, adversarial loss, and $\ell_2$ loss.

%%%%%%%%%%%%%%%%%%%%%%%%%%%%%%%%%%%%%%%%%%%%%%%%%%% Depth-guided networks %%%%%%%%%%%%%%%%%%%%%%%%%%
\subsection{Depth-guided networks}
Depth map or transmission map plays a vital role in restoring the underwater image, which is related to the degradation induced by scattering. Therefore, it is a natural choice to predict the depth map or transmission map of the underwater image to improve the performance of enhancement and restoration. We list the depth-guided networks next.  

\subsubsection{URCNN}
Underwater residual convolutional neural network (URCNN) \cite{hou2018URCNN} is proposed by Hou~\etal, which aims to learn the transmission map. The URCNN, in the first, uses a convolutional layer followed by ReLU to extract features. The batch normalization and ReLU succeed the second Conv layer. This pattern is repeated until the reconstruction layer, where only the convolutional layer is employed to output the transmission map. A global skip connection is used to enforce residual learning. The output transmission map is used to refine the input image. 

The network architecture of the URCNN is a modified version of VGG~\cite{simonyan2014vgg} and the input to the network is 180$\times$180 transmission map instead of the original image. The underwater images are generated from randomly selected 1000 NYU dataset~\cite{silberman2012NYU} images. Furthermore, using random medium attenuation coefficient and background light, a total of 1800 images are generated for training and 200 images for testing. The initial learning rate is selected to be 10$^{-1}$ and reduced to 10$^{-4}$ for 60 epochs. The depth of the network is 25 layers with each layer having 64 feature maps and a filter size of 3$\times$3. Similar to~\cite{wang2017UIENet}, the loss used for learning is $\ell_2$.   

\subsubsection{UIR-Net}
Cao~\etal~\cite{cao2018underwater} lately developed a deep network for underwater image restoration inspired by classical methods where the transmission map and the background light are estimated and computed independently. Consequently, two different network architectures were proposed \ie the light network (BL-Net) and the transmission map network (TM-Net) while collectively, the network is called UIR-Net~\cite{cao2018underwater}. The background light network (BL-Net) is simple and consists of five layers. The initial three layers are convolutional with BN and pooling. The last two layers are fully connected ones. The output of this BL-Net is thresholded to constrain it, in the range of [0,1]. The transmission map network (TM-Net) is more complicated and is based on~\cite{eigen2014depth}, consisting of two subnets, \ie, coarse-global subnet, and refine subnet. The coarse subnet is made of five convolutional layers, with the first two convolutional layers having pooling and batch normalization. The last layers of the coarse-global subnet are fully connected ones. The refined subnet has three convolutional layers and an upsampling layer which lies before the final convolutional layer. The output of this network is the depth map. Using depth maps, the transmission maps are computed. As a last preprocessing step, the guided filter \cite{He2010Guided} is applied to refine the maps further. 

The loss for the BL-Net is Euclidean while for the TM-Net is a scale-invariant minimum square error (MSE) adopted from Eigen~\etal~\cite{eigen2014depth}. Similar to~\cite{wang2017UIENet}, UIR-Net~\cite{cao2018underwater} use NYU-v2 dataset~\cite{silberman2012NYU} to generate 12,000 synthetic underwater images using a total of 29 different underwater ambient lights. The BL-Net is initialized randomly, while TM-Net utilizes the weights from VGG \cite{simonyan2014vgg}.

\subsubsection{WaterGAN}
WaterGAN~\cite{li2018watergan} as the name indicates, is a generative adversarial network, which manipulates RGB-D images to simulate underwater images for color correction. The authors present a two-part solution where the first part in the pipeline is the WaterGAN~\cite{li2018watergan}, and the second part is the image restoration network, composed of a depth estimation network and a color correction network. The WaterGAN has two systems: a generator G and discriminator D. The generator is a noise vector, which is projected, reshaped and passed through several convolutional and deconvolutional layers which output a synthetic image. The discriminator distinguishes between real image (from another dataset) and synthetic (generated by generator). The generator aims to create images which the discriminator classify as real. 

The underwater images generated by~\cite{li2018watergan} are passed through an image restoration network. The network is inspired by an encoder-decoder architecture, particularly, pixel-wise dense learning, and SegNet~\cite{badrinarayanan2015segnet}. The SegNet uses a non-parametric upsampling layer which benefits from the max-pooling index information in the encoder. Furthermore, the authors incorporate the skipping layers in the encoder-decoder architecture to compensate for the high frequencies' loss due to pooling operation. 

The authors collect 7,000 images from Michigan's Marine Hydrodynamics Laboratory. Another 6,500 images are collected from Port Royal, Jamaica. Similarly, 6,083 images are gathered from the coral reef system, Australia~\cite{pizarro2017survey}. Besides, four Kinect datasets \ie the B3DO~\cite{janoch2013B3DO}, the UW RGB-D~\cite{lai2014UWRGBD}, the NYU~\cite{silberman2012NYU} and the Microsoft 7-scenes~\cite{shotton2013Microsoft7scenes}, are utilized to form 15,000 underwater images via WaterGAN, out of which 12,000 are used for training and 3,000 for testing. The depth estimation network is trained separately at a fixed learning rate of 10$^{-6}$ while the color correction network is initially trained with an input resolution of 128 $\times$ 128 having learning rate 10$^{-6}$. After that, the authors refined the color correction network with input images of 512 $\times$ 512 resolution, reducing the base learning rate to 10$^{-7}$. The $\ell_2$ loss is utilized for depth estimation and color correction networks, and further, as a post-processing step, the images are normalized \ie [0,1].  

%%%%%%%%%%%%%%%%%%%%%%%%%%%%%%%% Multiple Generator GANs 
\subsection{Dual Generator GANs}
The dual generator GANs algorithms for underwater image enhancement employ multiple generators to predict the improved image. Currently, the trend is to use two generators with one discriminator or two generators with two discriminators; either the aim is to share the features between the generators or use the prediction of one generator as an input to the other generator. Examples of the dual generator GANs are the following.   

\subsubsection{UWGAN}
Based on the GANs~\cite{GAN2014}, Li~\etal~\cite{UWGAN2018} proposed a weakly supervised color transfer method for underwater image color correction, called UWGAN. The UWGAN model relaxes the need for paired underwater images for training and allows the underwater images to be regarded in unknown locations, which benefits from adversarial learning. Following the CycleGAN~\cite{CycleGAN2017}, the UWGAN model adopts a cycle structure which includes a forward network and a backward network to learn the mapping functions between a source domain (\ie, underwater) and a target domain (\ie, air). The purpose of such a cycle structure is to capture the unique characteristics of one image collection and figure out how these characteristics could be translated into the other image collection. 

The generators used in the UWGAN~\cite{UWGAN2018} have the same architecture as~\cite{Johnson2016}. For the discriminators, the UWGAN uses 70$\times$70 PatchGANs~\cite{PatchGAN2016}. To train the network, 3800 underwater images and 3800 high-quality air images are collected and are resized to 256$\times$256. The final loss function is the linear combination of three-loss functions, including adversarial loss, cycle consistency loss, and SSIM loss. The adversarial loss is to match the distribution of generated images with that of the target domain. The cycle consistency loss is to prevent the learned mappings from contradicting each other. The SSIM loss is to preserve the content and structure of source images.  

\subsubsection{MCycleGAN}
To restore underwater images, Lu~\etal~\cite{lu2019MCycleGAN} proposed a Multi-Scale Cycle Generative Adversarial Network (MCycleGAN), which is a variant of the CycleGAN network~\cite{CycleGAN2017}.  The authors incorporate the multiscale SSIM loss into the CycleGAN \cite{CycleGAN2017} to improve the image restoration task. The aim is to transfer the underwater style to the recovered style image.

As a first step, the dark channel prior (DCP) \cite{he2011DCP} is used to obtain the transmission map of a turbid underwater image.   Additionally, the transmission maps provide depth information in the form of three binary filters. The turbid underwater images are forwarded through the generator network. The turbid and generated clear underwater images are split into R, G, and B channels. The channels are then subjected to different size of sliding windows to compute the SSIM loss between the turbid and generated images. Furthermore, the SSIM maps are multiplied with corresponding filters and added together, which results in the multiscale SSIM map for final loss computation. As a final step, both the real-world underwater image and the computed ones are passed through the discriminator.

CycleGAN~\cite{CycleGAN2017} inspired the generator and discriminator of MCycleGAN~\cite{lu2019MCycleGAN}. More specifically,  the generator is adapted from image superresolution by Johnson~\etal~\cite{johnson2016perceptual} which consists of nine ResNet blocks with training images of size 256$\times$256 while the discriminator is based on 70$\times$70 PatchGANs~\cite{isola2017image,ledig2017photo} to differentiate between real and fake image patches. The loss function is a union of the adversarial loss, the cycle-consistent loss, and the multiscale SSIM loss. The dataset is composed of 1,037 turbid underwater images collected from ImageNet~\cite{deng2009imagenet} and Jiao Zhou Bay, out of which 837 are retained as a training dataset, and the rest 200 are reserved for testing. ADAM~\cite{kingma2014adam} is used as an optimizer adopting a fixed learning rate of 0.0002 until convergence.

\subsubsection{UIE-sGAN}
Yu~\etal~\cite{ye2018underwater} proposed an underwater image enhancement system using stacked conditional generative adversarial networks, abbreviated as UIE-sGAN. The proposed network architecture consists of two subnetworks \ie haze detection subnetwork and color correction subnetwork. Each subnetwork has a generator and discriminator, and the color correction subnetwork is stacked on the haze detection subnetwork. For the haze detection subnet, the generator is similar to UNET~\cite{ronneberger2015unet} consisting of seven convolutional layers and seven deconvolutional layers, both followed by BN and leaky ReLU except the first convolutional layer where only leaky ReLU is employed and the last deconvolutional layer where TanH nonlinear function is realized. While the discriminator is made of four convolutional layers where the initial layer has leaky ReLU purely, and the subsequent ones have batch normalization and leaky ReLU followed by a sigmoid layer. The output of the haze detection network is a haze mask. The structure of haze detection subnet and the color-correction subnet is identical except that color-correction subnet takes the haze mask and RGB images as input and outputs a color corrected underwater image. 

The UIE-sGAN~\cite{ye2018underwater} has three losses \ie the adversarial loss for each network and a consistency loss. The training is accomplished by using WaterGAN~\cite{li2018watergan} to generate underwater images from NYU-v2 dataset~\cite{silberman2012NYU}. Out of 1449 images, 1200 are held for training while the network is evaluated on the remaining ones. The images are resized to 286$\times$286 and then cropped to 256$\times$256 and further applying data augmentation. The network is optimized using ADAM by fixing the learning rate as 5$\times$10$^{-5}$.   

\begin{table*}[thb]
\caption{\textbf{Network Specifics:}  Essential parameters of underwater image enhancement and restoration networks. The losses \ie, $\ell_{gan}$, $\ell_c$, $\ell_{W}$, $\ell_{nui}$, $\ell_r$  and $\ell_g$ represents adversarial,  consistency, Wasserstein, nuisance, relativistic and gradient losses, respectively. The \enquote{-} means information is not available.}
\begin{center}
\begin{adjustbox}{max width=\textwidth}
\begin{tabular}{l||ccccccccl}\hline
&\multicolumn{9}{c}{Network Parameters}\\ \cline{2-10}
                                 		&Patch          & Network   &Feature    &Variable &         &	Residual  & Skip		        &            &									\\
\multicolumn{1}{c||}{Methods}           &Size           & Depth     &maps       &Kernels  & Blocks  &	learning  & connections		    & Framework  & Loss      						\\\hline\hline
UIE-Net~\cite{wang2017UIENet}      		&32$\times$32   & 7         &16-20		&\ch      & 		&			  &						&	-		 & $\ell_2$							\\
UIR-Net~\cite{cao2018underwater}  		&224$\times$224	& 8	        &96-384		&\ch	  &		 	&			  &						&	-	 	 & $\ell_2$							\\
P2P Net~\cite{sun2018p2p}             	&66$\times$66   & 6		    &96-384		&\ch	  &		 	&\ch 	  	  &	\ch 				& Caffe		 & $\ell_2$							\\
UIE-sGAN~\cite{ye2018underwater}  		&256$\times$256 & 16		&64-512		&		  &		 	&\ch		  &	\ch					& TensorFlow &$\ell_{gan}$,$\ell_c$				\\
WaterGAN~\cite{li2018watergan}    		&512$\times$512 & 42      	& 128-512	&		  &		    &\ch		  &	\ch		 			& Caffe		 &$\ell_2$			 				\\
UGAN~\cite{fabbri2018UGAN}        		&256$\times$256 & 9			&64-512		&\ch	  &		 	&\ch		  &	\ch					& TensorFlow &$\ell_1$,$\ell_{W}$ 				\\     
UWCNN~\cite{UWCNN2018}            		&310$\times$230 & 10		&32			&		  &\ch		&\ch		  &	\ch					& TensorFlow &$\ell_2$,$\ell_{SSIM}$			\\
URCNN~\cite{hou2018URCNN}         		&180$\times$180 & 25		& 64		&		  &			& \ch		  & \ch		 			& MatConvNet &$\ell_2$   		 				\\
UWGAN~\cite{UWGAN2018}            		&256$\times$256	&18 		&64-256		&\ch	  &\ch		&			  &						& TensorFlow &$\ell_{gan}$,$\ell_c$,$\ell_{SSIM}$\\
DUIENet~\cite{libenchmark2019}    		&112$\times$112 &8		    &32-128	    &\ch	  &		 	&		      &						& TensorFlow &$\ell_{perceptual}$\\
MCycleGAN~\cite{lu2019MCycleGAN}  		&256$\times$256	& 24		&64-128		&\ch      &\ch	 	&\ch		  & \ch					& TensorFlow &$\ell_{gan}$,$\ell_c$,$\ell_{MSSIM}$\\
DenseGAN~\cite{DenseGAN}          		&256$\times$256 & 10		&64-512		& \ch	  & \ch 	&\ch 		  & \ch					& TensorFlow &$\ell_2$,$\ell_{gan}$,$\ell_g$	\\
FGAN~\cite{LI2019FGAN}            		&256$\times$256 & 8			&64-256		& \ch	  & \ch 	&\ch		  & \ch				    & TensorFlow &$\ell_2$,$\ell_{gan}$,$\ell_r$ 	\\
UIE-DAL~\cite{uplavikar2019UIE-DAL} 	&256$\times$256 & 27		&64-512		&		  &		 	&\ch		  &	\ch					&	-		 &$\ell_2$,$\ell_{gan}$,$\ell_{nui}$ 	\\\hline
\end{tabular}   
\end{adjustbox}
\label{table:parameters}
\end{center}
\end{table*}

%%%%%%%%%%%%%%%%%%%%%%%%%%%%%%%% Network specifics %%%%%%%%%%%%%%%%%%%%%%%%%%%%%%%%%

\subsection{Network Specifics}
After reviewing current deep learning-based underwater image enhancement algorithms, we emphasize the different aspects of the above-mentioned deep models. First, we summarize the network specifics of different models in Table~\ref{table:parameters} and then further analyze network loss, depth, parameters, and input patch size.

\textbf{Network Loss}
Network loss plays an integral part in learning the task underhand.  Here, we discuss the losses employed in deep underwater image enhancement. The most popular type of loss functions are to minimize the per-pixel error between the ground-truth image and the predicted image, commonly known as $\ell_1$ and $\ell_2$. For example, the UIE-Net~\cite{wang2017UIENet}, UIR-Net~\cite{cao2018underwater}, P2P Net~\cite{sun2018p2p}, and URCNN~\cite{hou2018URCNN} only use $\ell_2$ to optimize their networks. Usually, other losses such as SSIM, gradient \etc, are combined with the ones mentioned earlier to improve the performance of the networks, \eg UWCNN~\cite{UWCNN2018}. On the other hand, GANs rely on adversarial loss and perceptual loss to enhance the perceptual quality of the enhanced images, such as DenseGAN~\cite{DenseGAN}, UWGAN~\cite{UWGAN2018}, \etc

\textbf{Network Depth and Paramters}
The network depth and the number of parameters are related. The deeper the network, the more the number of parameters. Unlike other image classification~\cite{he2016deep} and enhancement tasks~\cite{anwar2019densely} where the network depth has exponentially increased and even consists of hundreds of convolutional layers, the underwater image enhancement networks are still very shallow composed of less than 45 layers (deepest network is the WaterGAN~\cite{li2018watergan} with 42 layers); hence comprised of very less number of parameters\footnote[2]{As most of the network models are not publicly available, a fair comparison to determine exact number of parameters is not possible.}.

\textbf{Input Patch Size}
Contrary to low-level vision tasks, most of the underwater image enhancement algorithms operate on full-size images. The reason may be to incorporate the wavelength dissipation of red, green, and blue channels. Furthermore, some algorithms reduce the image to predefined size, which requires upsampling as a post-processing step, such as MCycleGAN~\cite{lu2019MCycleGAN}, DenseGAN~\cite{DenseGAN}, and UWGAN~\cite{UWGAN2018}.  

\begin{figure}
\begin{center}
\begin{tabular}{c@{ }c@{ }c}
    \includegraphics[width=.15\textwidth,height=2.7cm]{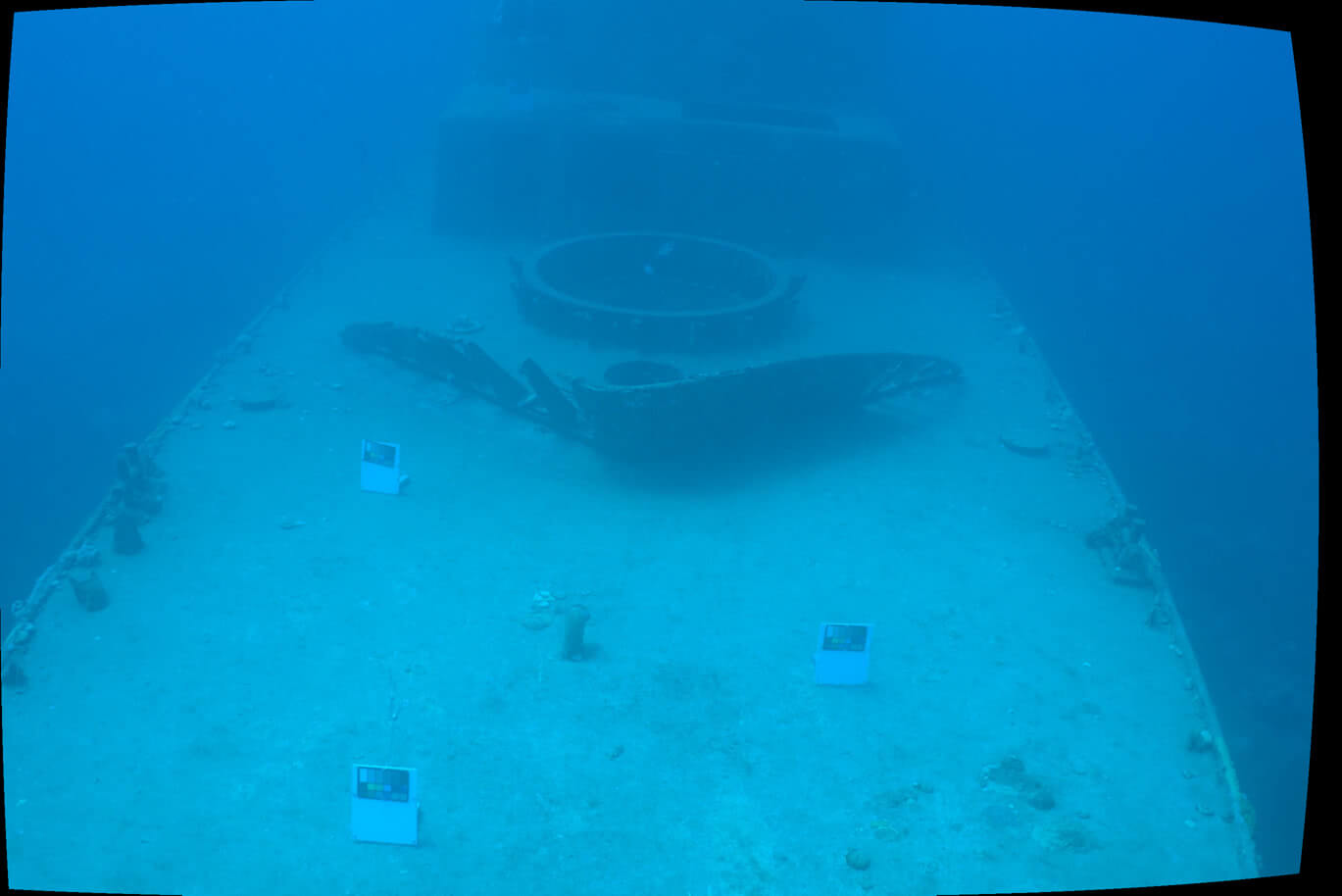}&
    \includegraphics[width=.15\textwidth,height=2.7cm]{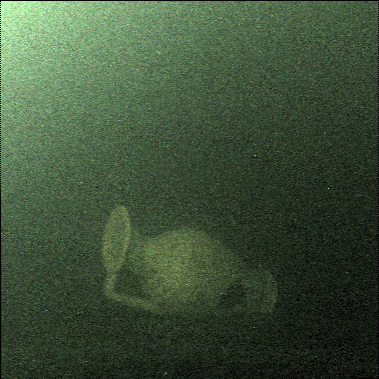}&
    \includegraphics[width=.15\textwidth,height=2.7cm]{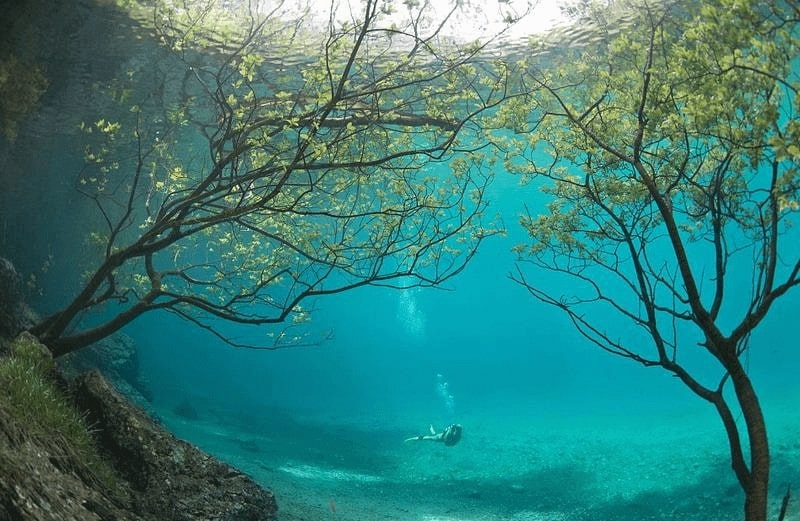}\\

    \includegraphics[width=.15\textwidth,height=2.7cm]{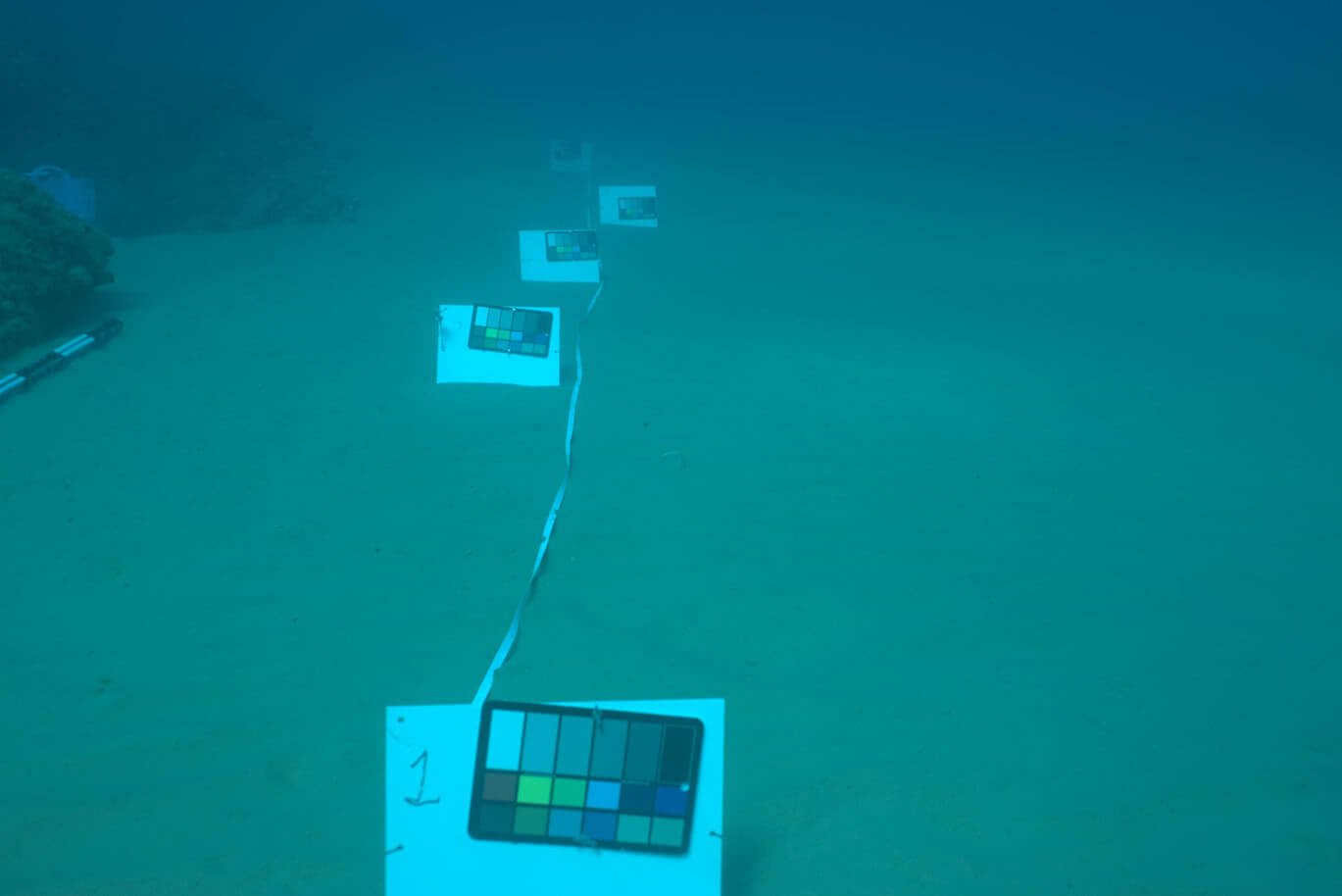}&
    \includegraphics[width=.15\textwidth,height=2.7cm]{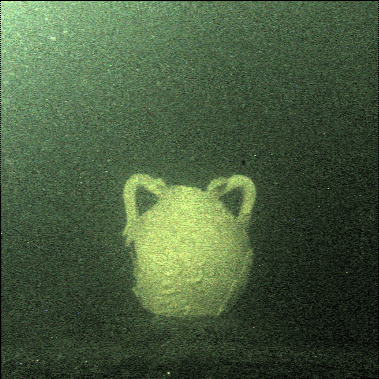}&
    \includegraphics[width=.15\textwidth,height=2.7cm]{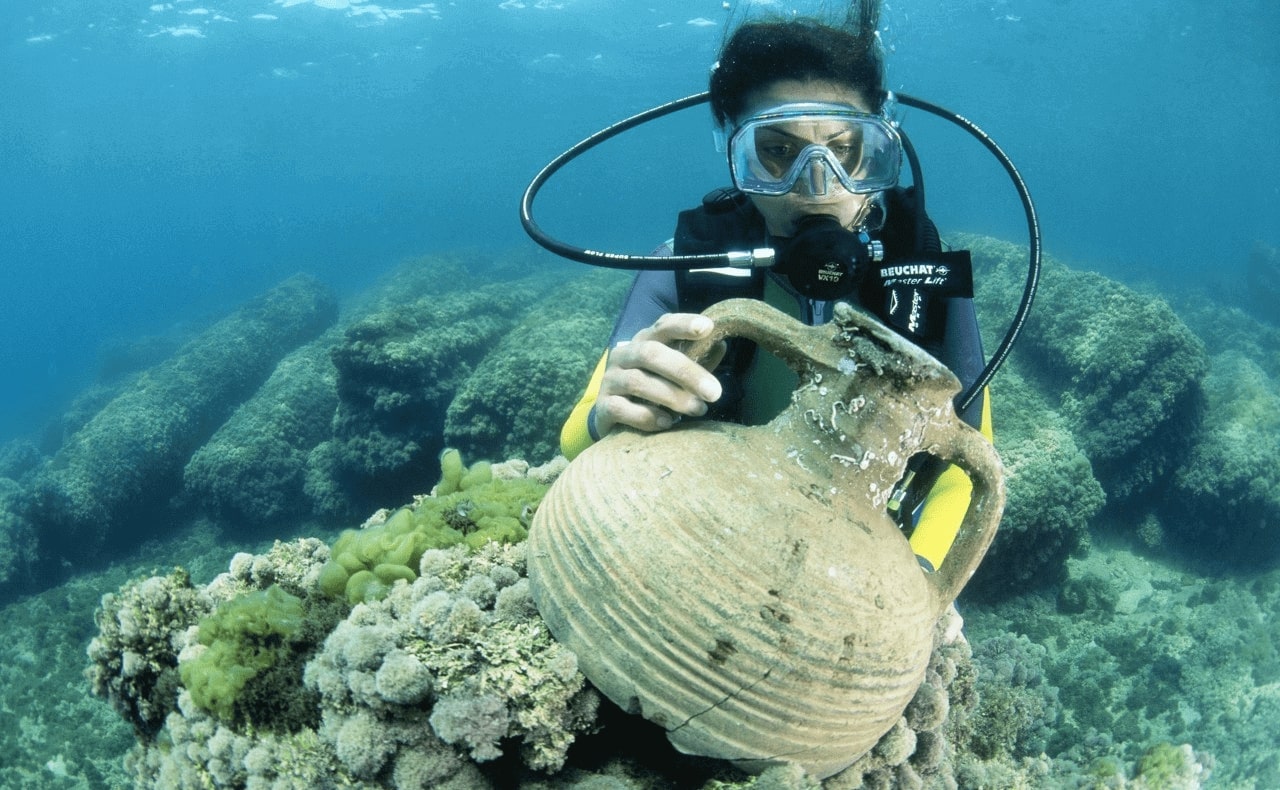}\\

    \includegraphics[width=.15\textwidth,height=2.7cm]{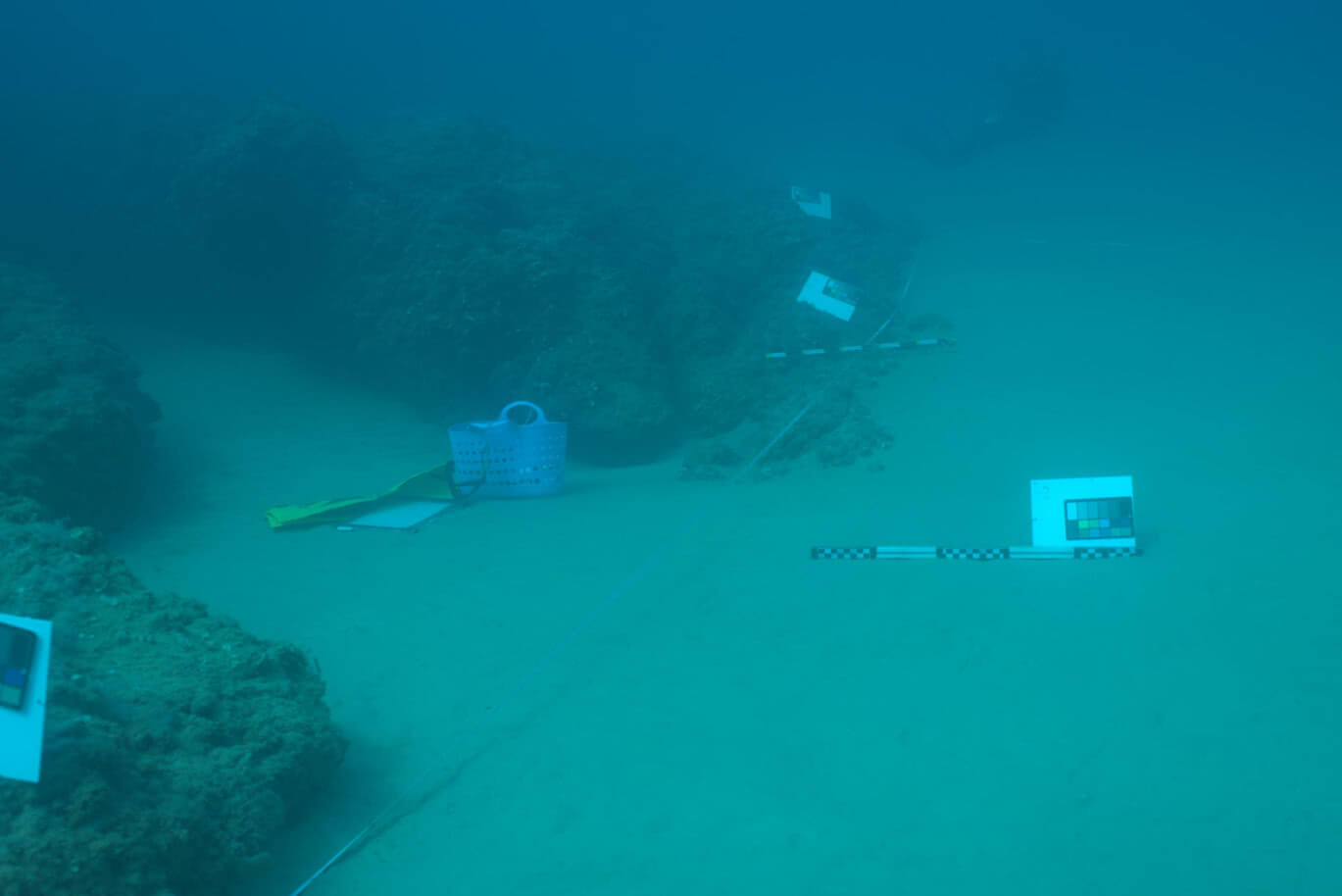}&
    \includegraphics[width=.15\textwidth,height=2.7cm]{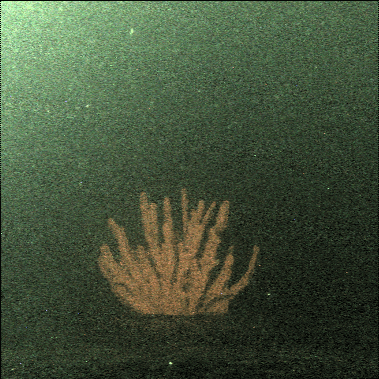}&
    \includegraphics[width=.15\textwidth,height=2.7cm]{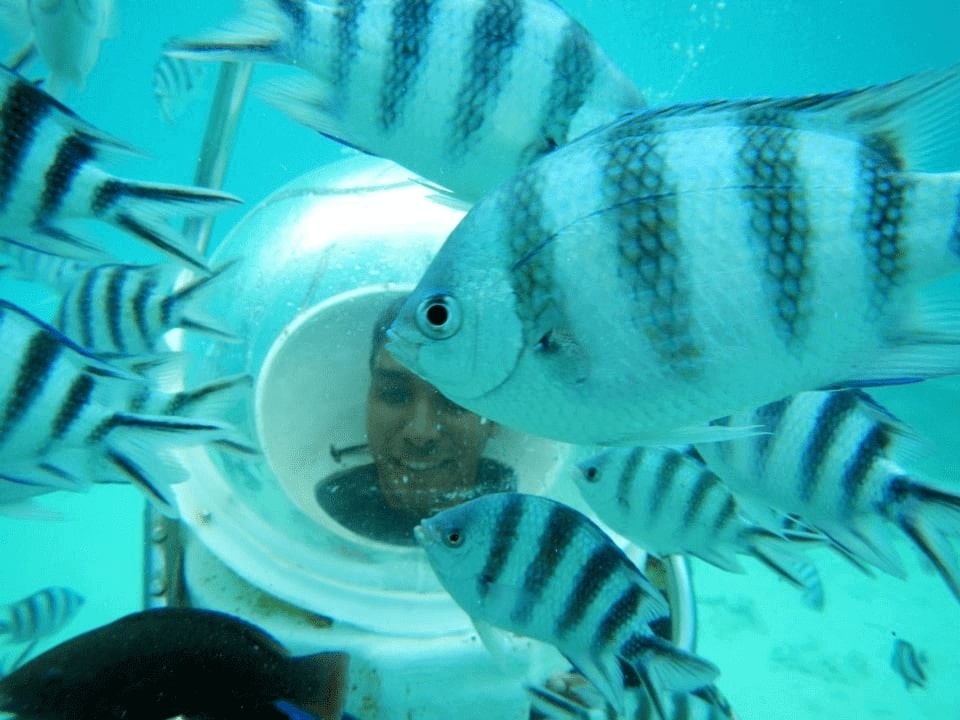}\\
    Haze-line~\cite{haze-line}& ULFID~\cite{ULFID}& UIEBD~\cite{libenchmark2019}\\

\end{tabular}
\end{center}
\caption{\textbf{Representative images:} Three sample images from Haze-line~\cite{haze-line}, ULFID~\cite{ULFID}, and UIEBD~\cite{libenchmark2019} datasets to show the diversity of the underwater images.}
\label{fig:im_sample}
\end{figure}

\section{Experimental Settings} 

\subsection{Real-world Underwater Image Datasets} 
Due to the limitations of synthetic underwater image datasets (\eg, inaccurate formation models, hard assumptions, insufficient images, specific scenes, \etc), we mainly introduce the real-world underwater image datasets in this section.

\begin{itemize}
\itemsep1em 
\item \textbf{Fish4Knowledge}~\cite{Fish4Knowledge} is funded by the European Union Seventh Framework program for the study of marine ecosystems, which provides a video and fish analysis dataset (about 200 Tb in size)\footnote[3]{\url{http://groups.inf.ed.ac.uk/f4k/index.html}}. 

\item \textbf{ULFID:} Underwater Light Field Image Dataset~\cite{ULFID} contains several underwater light field images in pure water and hazy conditions, as well as images taken in the air for reference\footnote[4]{\url{https://github.com/kskin/data}}. 

\item \textbf{MARIS:} Marine Autonomous Robotics for InterventionS~\cite{MARIS} is to advance the development of cooperating AUVs for undersea intervention in the offshore industry, in search-and-rescue tasks, and in various flavors of scientific exploration. This project provides several underwater images and videos captured by underwater stereo vision system\footnote[5]{\url{http://rimlab.ce.unipr.it/Maris.html}}. 

\item \textbf{Haze-line Dataset}~\cite{haze-line} collected a dataset of images taken in different locations with varying water properties, showing color charts in the scenes (about 33GB in size). Moreover, the 3D structure of the scene was calculated based on stereo imaging\footnote[6]{\url{http://csms.haifa.ac.il/profiles/tTreibitz/datasets/ambient_forwardlooking/index.html}}.

\item \textbf{UIEBD:} Underwater Image Enhancement Benchmark Dataset~\cite{libenchmark2019} includes 950 real-world underwater images, 890 of which have the corresponding reference images where each reference image is selected from 12 enhanced results. The rest 60 underwater images which cannot obtain satisfactory references are treated as challenging data. The UIEBD~\cite{libenchmark2019} contains a large range of image resolution and spans diverse scene/main object categories.\footnote[7]{\url{https://li-chongyi.github.io/proj_benchmark.html}}.

\end{itemize}

The existing real-world underwater image datasets usually have monotonous content and limited quality degradation types. Moreover, these datasets did not provide the corresponding ground truth images because it is impractical to simultaneously obtain the degraded underwater image and the ground-truth of the same scene. The UIEBD~\cite{libenchmark2019} provides the corresponding reference images which can be considered for full-reference image quality assessment. We conduct experimental quantitative and visual comparisons on this dataset. Besides, to validate the generalization of current deep algorithms, we also present the visual results of different methods on another two datasets \ie, Haze-line dataset~\cite{haze-line} and ULFID~\cite{ULFID}. Some representative samples of these three datasets are given in Figure~\ref{fig:im_sample}. 

%%%%%%%%%%%%%%%%%%%%%%%%%%%%%% Metrics %%%%%%%%%%%%%%%%%%%%%%%%%%%%%%%

\subsection{Evaluation Metrics}
Evaluations performed for underwater image enhancement can be broadly categorized into automatic evaluation metrics and human visual system (HVS).  The automatic evaluations are performed using six metrics, out of these, four are also most widely used in image enhancement and restoration problems \ie PSNR, MSE, and SSIM~\cite{wang2003SSIM}, and PCQI~\cite{wang2015PCQI} while the other two are specific for underwater image enhancement \ie UCIQE~\cite{yang2015UICQE} and UIQM~\cite{panetta2015UIQM}.  Next, to make the article inclusive, we describe all the evaluation metrics and then detail their limitations and reliability. Moreover, we also provide the report which details the human visual evaluation and its importance.

\subsubsection{Automatic evaluation metrics}
\begin{itemize}
\itemsep2em 
\item \textbf{MSE and PSNR:}
We begin our discussion with Mean Square Error (MSE) as the signal measure. The MSE aims to provide a quantitative score that represents the similarity or distortion between the two signals. Usually, one of the signals is the original signal, and the other one is recovered from some distortion or contamination. Mathematically, the MSE between the two signals can be expressed as:
\begin{equation}
    \text{MSE} = \frac{1}{N}\sum_{i=1}^{N}(x_i - y_i)^2,
\end{equation}
where $x$ and $y$ are two signals, in this case, images and $x_i$ and $y_i$ are the pixels at $i^{th}$ location. Similarly, $N$ are the number of pixels. Furthermore, in the image processing literature, peak signal to noise ratio (PSNR) measure is computed from MSE as:
\begin{equation}
    \text{PSNR} = 10\log_{10}\frac{L^2}{\text{MSE}}
\end{equation}
where $L$ is the dynamic range of image pixel intensities (\ie, 255 for image). The usage of MSE and PSNR has many attractive features \eg 1) it is simple, 2) all norms are valid distance metrics, 3) it has a clear physical meaning, and 4) these are excellent metrics in the context of optimization. The mentioned measures assume that the signal fidelity is independent of the relationship between 1) the original signal, 2) the distorted and original signal, and 3) the signs of the error signal. Unfortunately, none of them even roughly holds in the context of measuring the visual perception of image fidelity~\cite{wang2009MSELoveOrLive}. In the next section, we discuss alternatives to these measures.

%%%%%%%%%%%%%%%%%%%%%%%%%%%%%%%%%%%%%%%%%% Results %%%%%%%%%%%%%%%%%%%%%%%%%%%%%%%%%%%%%
\begin{table*}[thb]
\caption{\textbf{Quantitative results:} The best results are highlighted with \textcolor{red}{red} color while the \textcolor{blue}{blue} color represents the second best.}
\begin{center}
\begin{tabular}{l||cccccc}\hline
&\multicolumn{6}{c}{UWE Dataset}\\ \cline{1-7}
%&&\multicolumn{5}{c||}{UWE Dataset}\\ \cline{1-6}
Method      & PSNR $\uparrow$  &MSE $\downarrow$ & SSIM $\uparrow$    & PCQI $\uparrow$   & UCIQE  $\uparrow$ &  UIQM  $\uparrow$\\ \hline\hline
Original       &  17.36& 1768.90&    \textcolor{blue}{0.6168}&  \textcolor{red}{1.1118}&  0.5196  &1.1571\\
MCycleGAN~\cite{lu2019MCycleGAN}      & \textcolor{blue}{18.33} &\textcolor{blue}{1132.21} & 0.6138   & 0.4521  & 0.5196 & 1.1471 \\
%MCycleGAN(GT:256 x 256)   & 18.52 &1092.69 & 0.6518   & 0.6145  & 0.5155    \\
URCNN~\cite{hou2018URCNN}           &15.94  & 2195.89& 0.5972 & \textcolor{blue}{1.0936}  &    0.5196 & \textcolor{red}{1.5332}\\
%WaterGAN        &       &        &        &         &      \\
UWGAN~\cite{UWGAN2018}           & 16.06 &1853.70 &0.2945  &0.6000   &\textcolor{blue}{0.5921} &1.1099     \\
DUIENet~\cite{libenchmark2019}         & \textcolor{red}{19.29} &\textcolor{red}{1012.20} &\textcolor{red}{0.8093}  &0.9844   &0.5720  &\textcolor{blue}{1.2963}  \\
DenseGAN~\cite{DenseGAN} & 17.56 &1363.60 &0.4239  &0.6697   &\textcolor{red}{0.6291} &1.0952   \\
\hline
UWCNN\_type-1~\cite{UWCNN2018}    & 13.03 &3930.80 &0.4795  &1.0310   &0.4876  & 1.1319  \\
UWCNN\_type-3~\cite{UWCNN2018}    & 13.58 &3297.40 &0.5482  &1.0146   &0.4771  & 1.1035  \\
UWCNN\_type-5~\cite{UWCNN2018}    & 13.29 &3427.20 &0.5102  &0.9223   &0.4303  &1.0122  \\
UWCNN\_type-7~\cite{UWCNN2018}    & 13.30 &3372.60 &0.4287  &0.8693   &0.4533  & 1.0385  \\
UWCNN\_type-9~\cite{UWCNN2018}    &10.58  &6164.80 &0.2598  &0.4958   &0.3636 & 0.7775   \\
UWCNN\_type-I~\cite{UWCNN2018}    & 15.00 &2345.00 &0.5306  &1.0890   &0.4954 &1.1294   \\
UWCNN\_type-II~\cite{UWCNN2018}   & 13.46 &3654.10 &0.4509  &1.0631   &0.4766 & 1.1048   \\
UWCNN\_type-III~\cite{UWCNN2018}  &14.24  &2920.20 &0.4945  &1.0486   &0.4739   &1.0333 \\
\hline
\end{tabular}
\label{table:UWE_Metric}
\end{center}
\end{table*}

\item \textbf{SSIM:}
Another commonly used measure is the Structural SIMilarity (SSIM) index. The main ideas of SSIM were presented by Wang $\&$ Bovik~\cite{wang2002SSIMIdea} and formulated in~\cite{wang2006SSIMimple1,wang2004SSIMimpli2}. Let us consider that $x$ and $y$ are the patches taken from the two different images but locations to be compared against each other. Then SSIM takes three measures into account, which are the similarity of the patch 1) luminance $l(x, y)$, 2) contrasts $c(x, y)$, and 3) the local structures $s(x, y)$. As pointed out in~\cite{wang2004SSIMimpli2}, these similarities are expressed and computed using simple statistics and are combined to produce local SSIM as:
\begin{equation}
\begin{split}
&\text{SSIM} = l(x, y) \cdot c(x, y) \cdot s(x, y),\\
            = &\left(\frac{2\mu_x\mu_y+C_1}{\mu^2_x + \mu^2_y +C_1}\right) \left(\frac{2\sigma_x \sigma_y + C_2}{\sigma^2_x + \sigma^2_y +C_2}\right) \left(\frac{\sigma_{xy}+C_3}{\sigma_x + \sigma_y +C_3}\right),\\
    \end{split}
\end{equation}
where $\mu_x$ and $\mu_y$ are means while $\sigma_x$ and $\sigma_y$ are standard deviations of the patches $x$ and $y$, respectively. Similarly, $\sigma_{xy}$ cross-correlation of the patches after removing their means. The constants $C_1$, $C_2$ and $C_3$ stabilize the terms to avoid near-zero divisions. 

\item \textbf{PCQI:}
Patch-based contrast quality index (PCQI)~\cite{wang2015PCQI} relies on patch-based approach as contrary to relying on global statistics. The PCQI depends on three independent quantities of an image patch \ie mean, signal strength and structure. Mathematically, a patch-based contrast image quality index (PCQI) is given by:

\begin{equation}
\text{PCQI} = q_i(x,y) \cdot q_c(x,y) \cdot  q_s(x,y),
\label{eq:pcqi}
\end{equation}
where $q_i(x,y)$ is to compare mean intensity, $q_c(x,y)$ is to determine the structural distortion and $q_s(x,y)$ is the contrast change. PCQI is mathematically expensive as compared to other metrics. Next, we discuss quantitative measures, which are more specific to underwater image enhancement.
% from A Patch-Structure Representation Method for Quality Assessment of Contrast Changed Images

\item \textbf{UCIQE:}
Underwater color image quality evaluation abbreviated as UCIQE~\cite{yang2015UICQE}, is based chroma, contrast, and saturation of CIELab and is defined as:
\begin{equation}
    \text{UCIQE} = C_1 \times \sigma_c + C_2 \times con_l + C_3 \times \mu_s,
\end{equation}
where $\sigma_c$, $con_l$ and $\mu_s$ are the standard deviation of chroma, the contrast of luminance, and the mean of saturation. It is to be noted here that for underwater images, human perception has a good correlation with the variance of chroma.

\item \textbf{UIQM:}
UIQM~\cite{panetta2015UIQM} stands for underwater image quality measure and is different from earlier defined evaluation metrics. The UIQ employs the HVS model only, and does not require a reference image; hence, a better candidate for evaluation of underwater images. UIQM is dependent on three attribute measures the underwater images, which are 1) image colorfulness measure (UICM), 2) sharpness measure (UISM), and 3) contrast measure (UIConM). Following is the formulation of UIQM: 
\begin{equation} 
\text{UIQM} = c_1 \times \text{UICM} + c_2 \times \text{UISM} + c_3 \times \text{UIConM},
\label{eq:UIQM} 
\end{equation}  
where $c_1$, $c_2$ and $c_3$ are the parameters which are application dependent, \eg, more weight should be given to $c_1$ for underwater color correct while $c_2$ for increasing visibility in the underwater scene.   
\end{itemize}

\subsubsection{Human Visual System}
Due to the lack of real ground-truth data, human subjects are used to evaluate the quality of the predicted images to an attempt to incorporate the perceptual measures. These human inputs may either be crowd-sourced or specialist persons in different competitions. However, none of these methods have shown any significant advantage over the mathematical measure. In other words, mathematically defined measures are still attractive due to the following reasons. 
\begin{itemize} 
\item They are simple to calculate and computationally inexpensive normally. 
\item They are independent of distinct individuals and observing conditions. 
\end{itemize} 

Furthermore, it is thought that viewing conditions play an influential role in human perception of image quality. However, if there are multiple viewing conditions, a method dependent on viewing conditions may produce different estimations that may be inconvenient to utilize. Moreover, it may also be specific to the user observation, and it then becomes the responsibility of each to compute the viewing conditions and provide the output to the measurement systems. On the other hand, a method independent of viewing conditions computes a single quantity that provides a general idea about the image quality. Besides, the experience of volunteers significantly affects human visual perception. The volunteers who understand what the degrading effects of attenuation and backscatter are, and what it looks like when either is improperly corrected can provide more reliable subjective scores of image quality.

\begin{figure*}[t]
\begin{center}
\begin{tabular}{c@{ }c@{ }c@{ }c}
%\begin{tabular}{c@{ }c@{ }c@{ }c@{ }c@{ }c@{ }c@{ }c}
    \includegraphics[width=.24\textwidth]{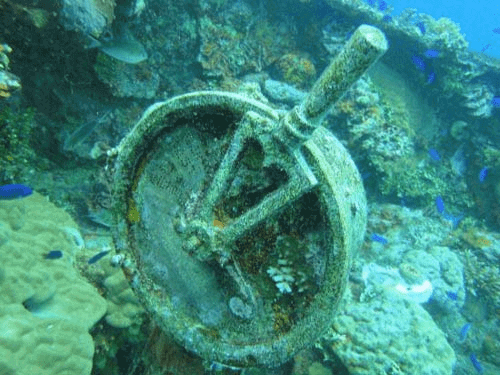}&
    \includegraphics[width=.24\textwidth]{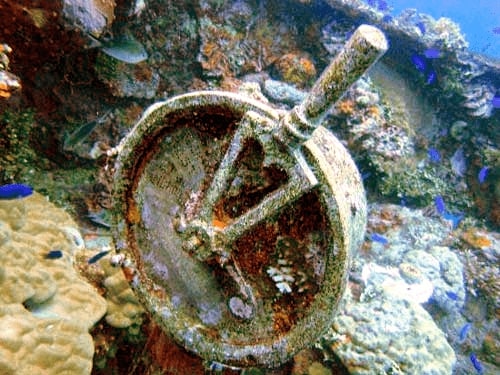}&
    \includegraphics[width=.24\textwidth]{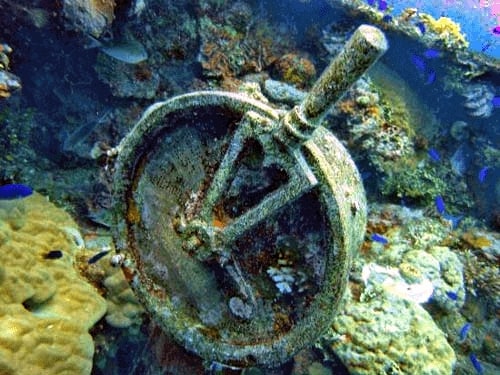}&
    \includegraphics[width=.24\textwidth]{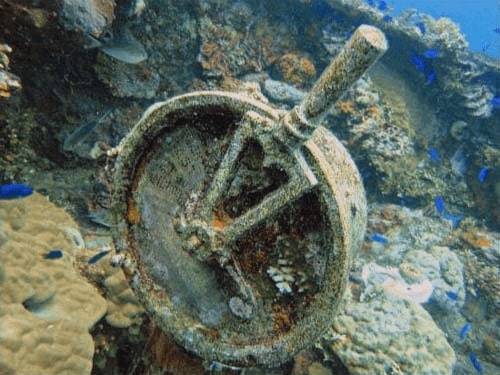}\\
    Underwater& Reference & URCNN~\cite{hou2018URCNN} & DUIENet~\cite{libenchmark2019}\\
    \includegraphics[width=.24\textwidth]{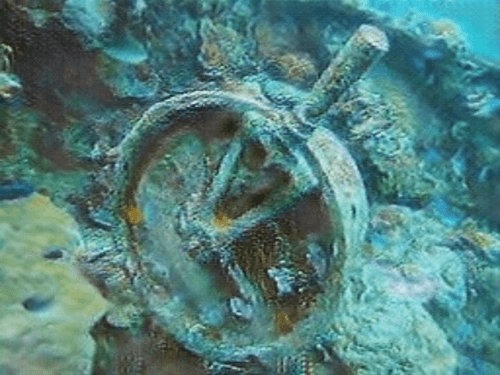}&
    \includegraphics[width=.24\textwidth]{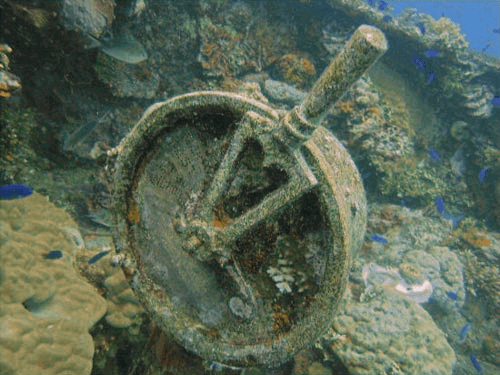}&
    \includegraphics[width=.24\textwidth]{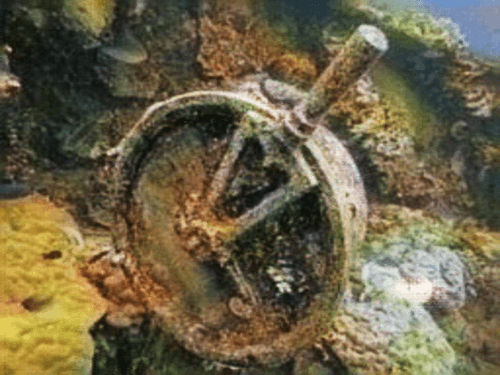}&
    \includegraphics[width=.24\textwidth]{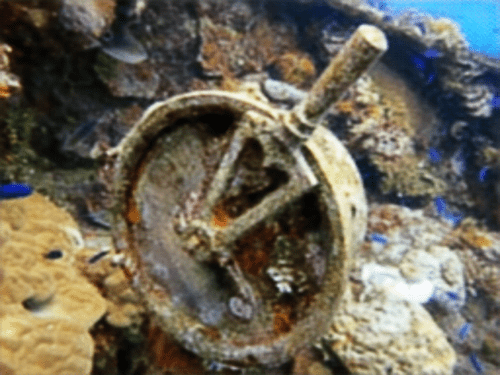}\\
     MCycleGAN~\cite{lu2019MCycleGAN} &UWCNN (type-I)~\cite{UWCNN2018} & UWGAN~\cite{UWGAN2018} & DenseGAN~\cite{DenseGAN} \\
    \\
    \includegraphics[width=.24\textwidth]{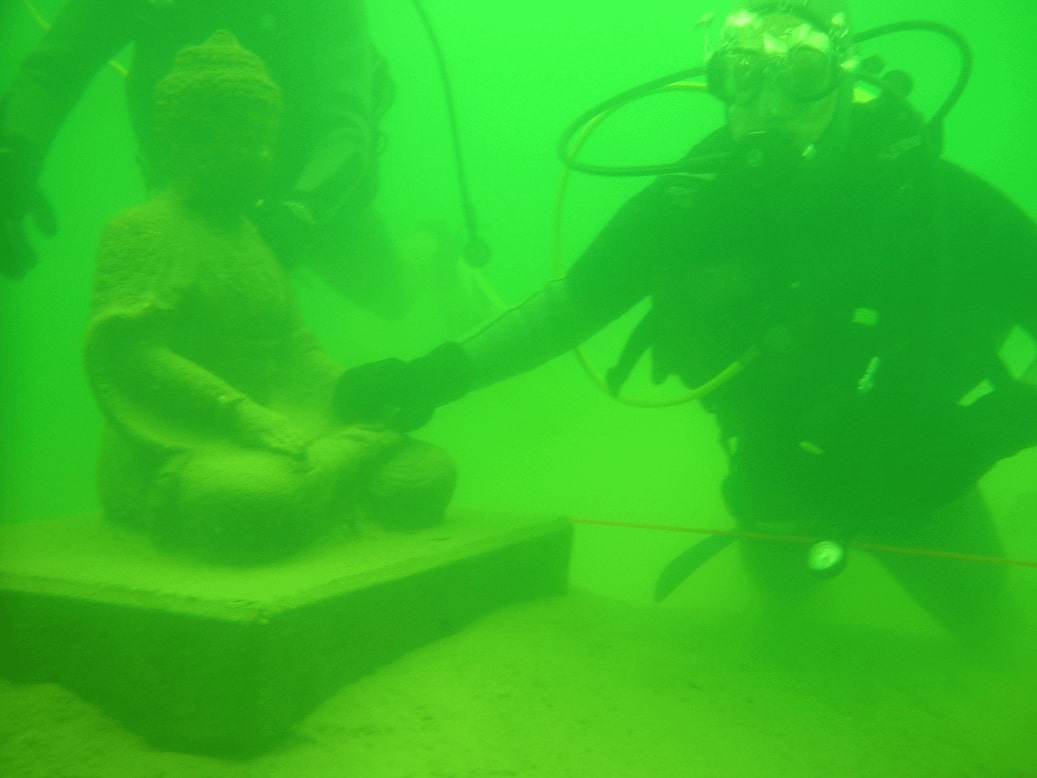}&
    \includegraphics[width=.24\textwidth]{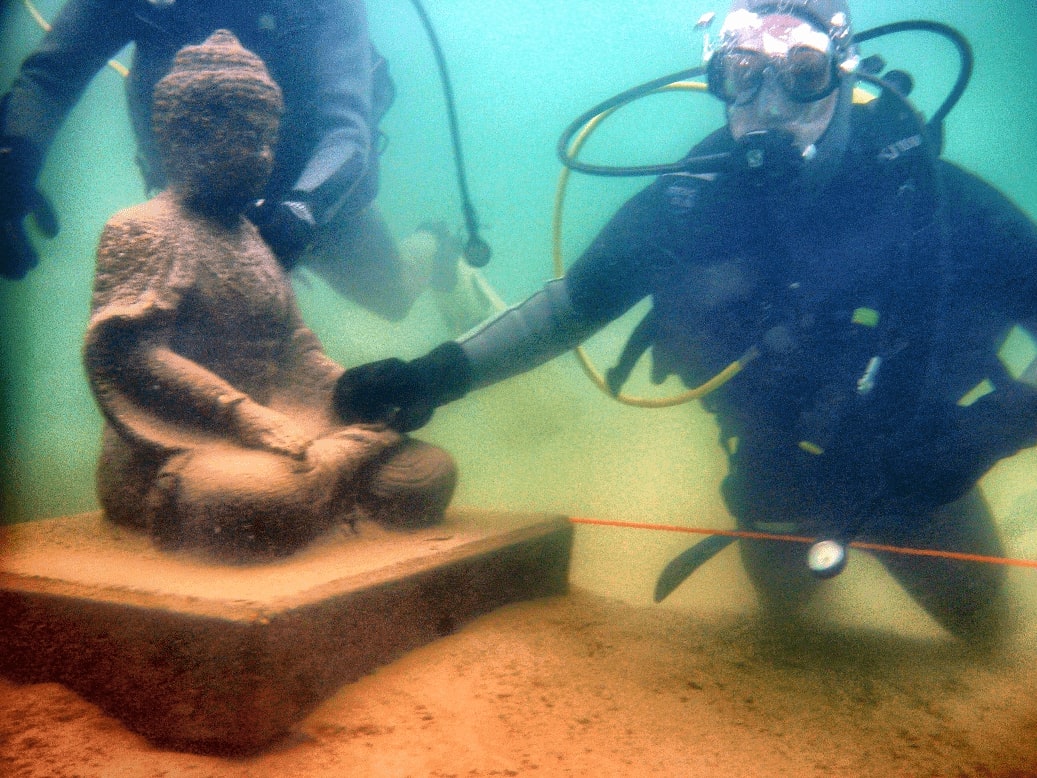}&
    \includegraphics[width=.24\textwidth]{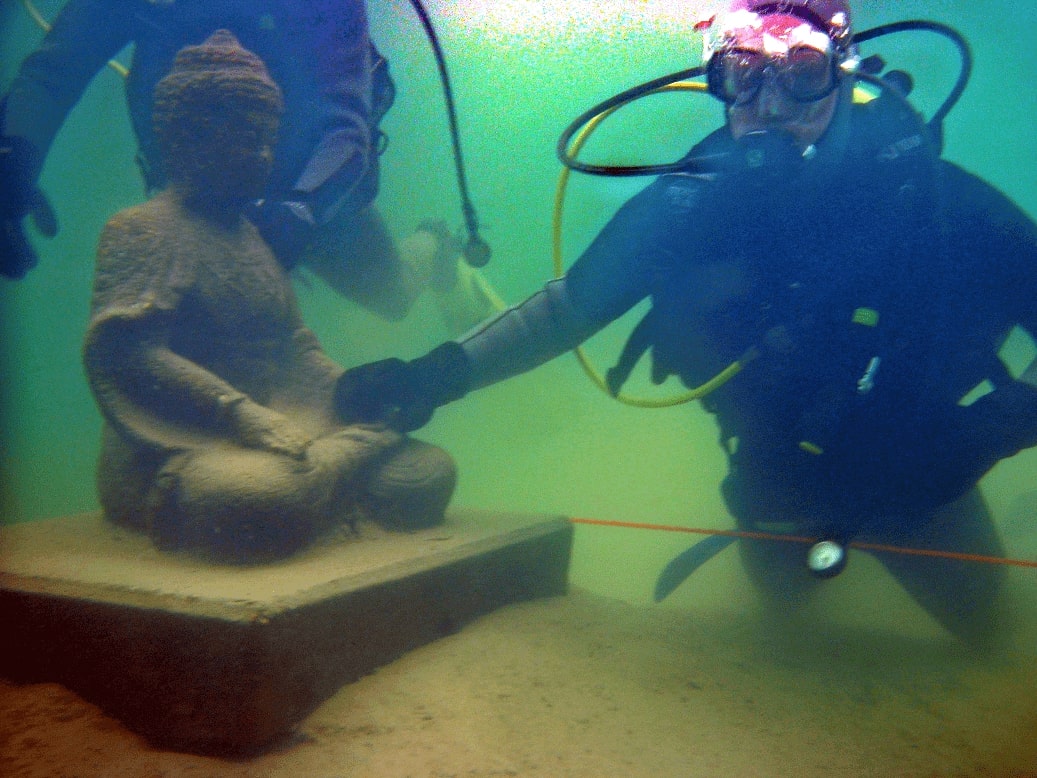}&
    \includegraphics[width=.24\textwidth]{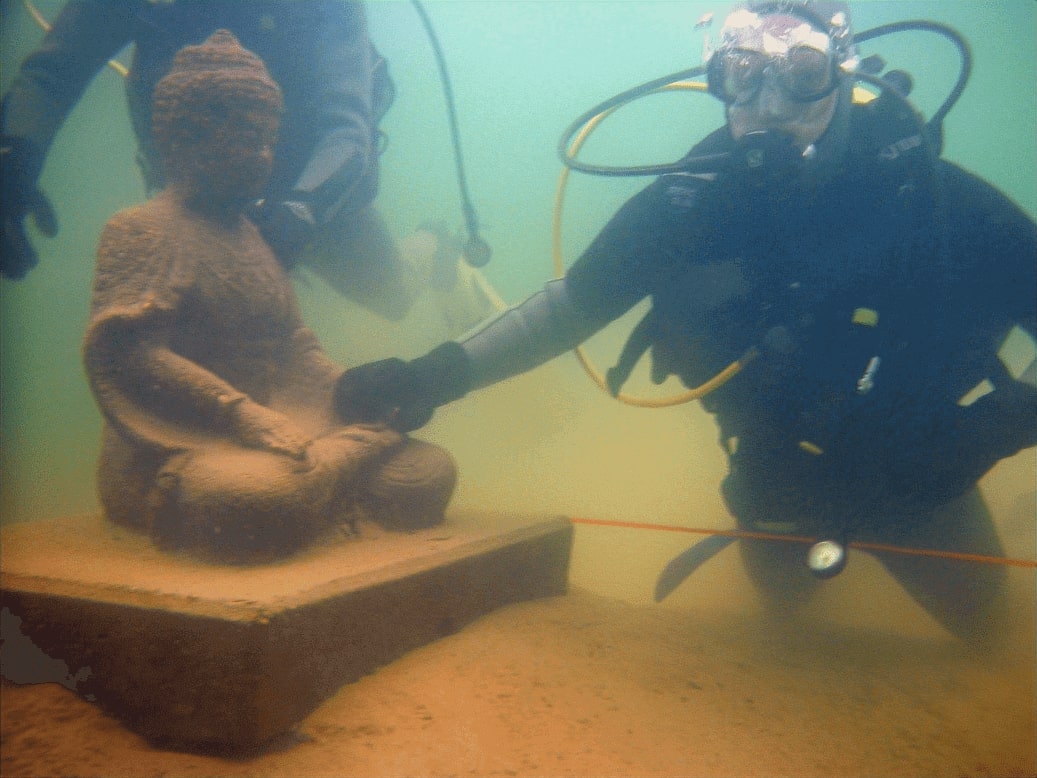}\\
    Underwater& Reference & URCNN~\cite{hou2018URCNN} & DUIENet~\cite{libenchmark2019}\\
    \includegraphics[width=.24\textwidth]{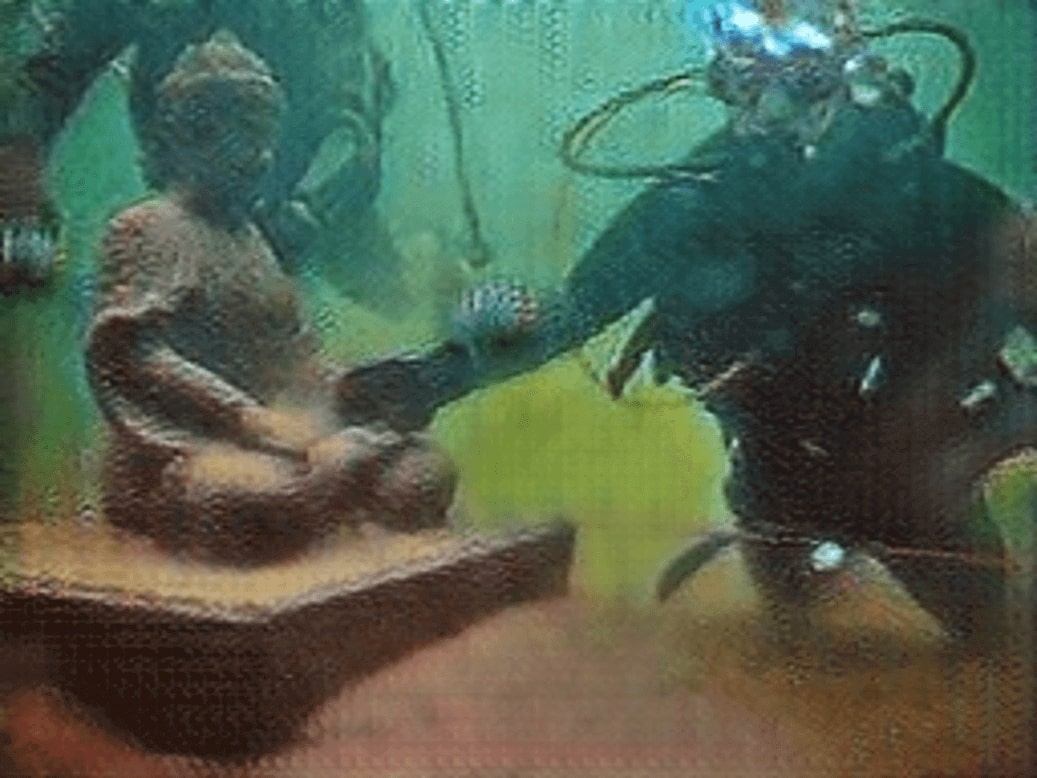}&
    \includegraphics[width=.24\textwidth]{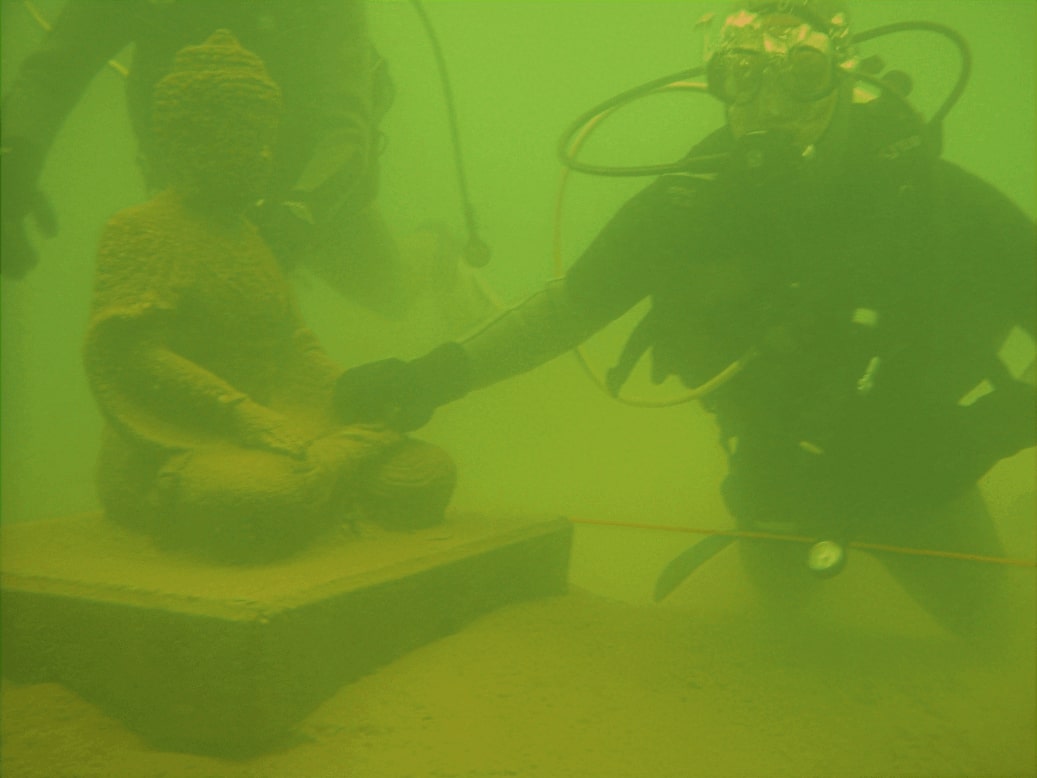}&
    \includegraphics[width=.24\textwidth]{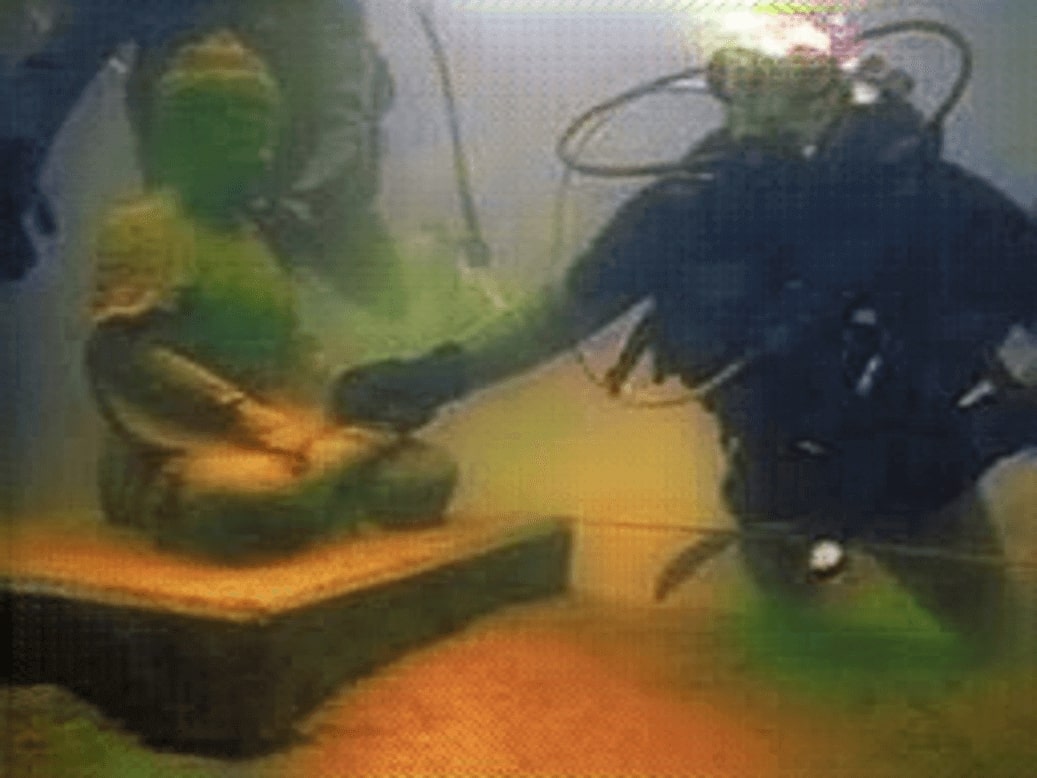}&
    \includegraphics[width=.24\textwidth]{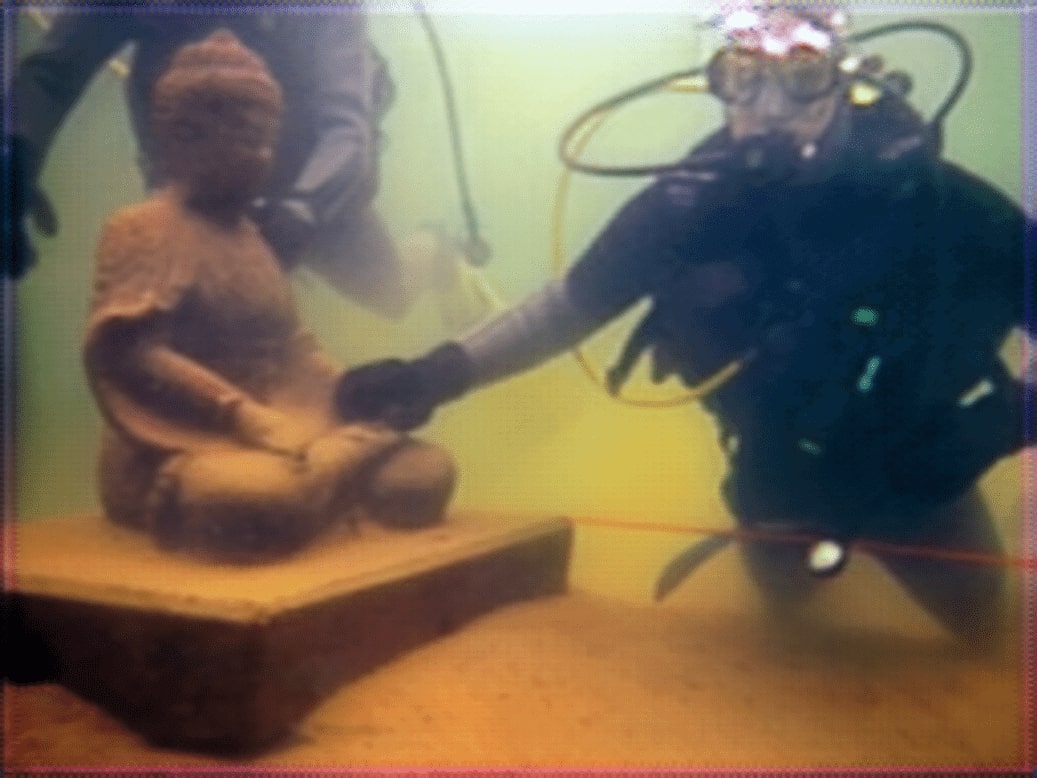}\\
    MCycleGAN~\cite{lu2019MCycleGAN} &UWCNN (type-I)~\cite{UWCNN2018} & UWGAN~\cite{UWGAN2018} & DenseGAN~\cite{DenseGAN} \\
 \end{tabular}
\end{center}
\caption{\textbf{Visual comparison of greenish images:} Comparisons of different methods on the greenish underwater samples from UIEBD~\cite{libenchmark2019}. Here, UWCNN-type-I represents the model trained by synthetic type-I training data.}
\label{fig:im_greenish}
\end{figure*}

\begin{figure*}[t]
\begin{center}
\begin{tabular}{c@{ }c@{ }c@{ }c}
%\begin{tabular}{c@{ }c@{ }c@{ }c@{ }c@{ }c@{ }c@{ }c}
    \includegraphics[width=.24\textwidth]{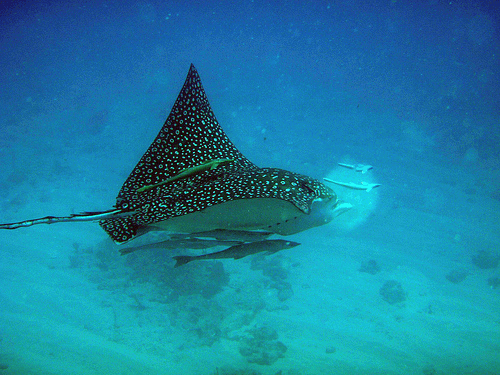}&
    \includegraphics[width=.24\textwidth]{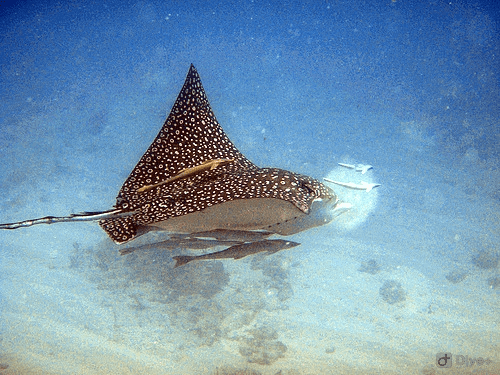}&
    \includegraphics[width=.24\textwidth]{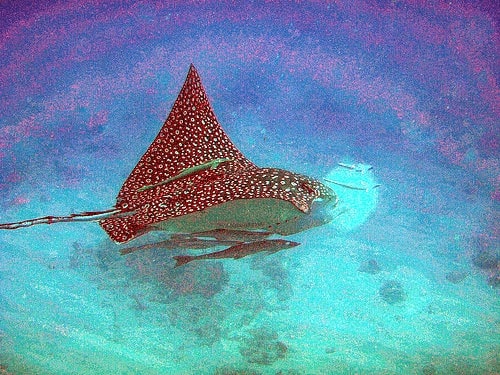}&
    \includegraphics[width=.24\textwidth]{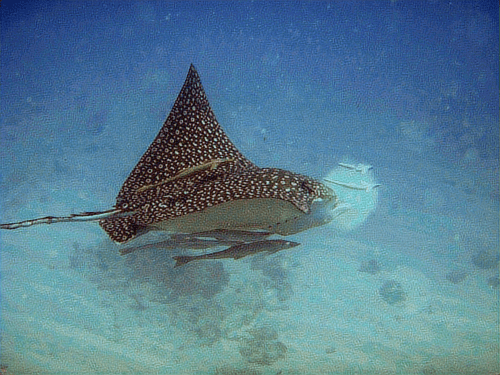}\\
    Underwater& Reference & URCNN~\cite{hou2018URCNN} & DUIENet~\cite{libenchmark2019}\\
    \includegraphics[width=.24\textwidth]{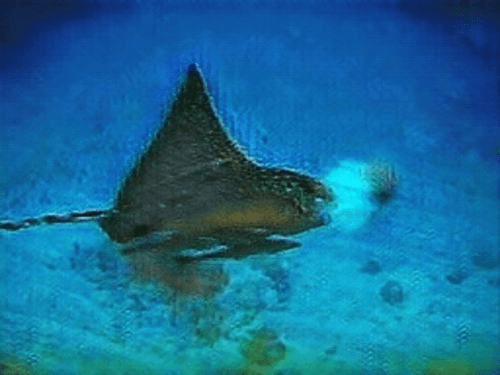}&
    \includegraphics[width=.24\textwidth]{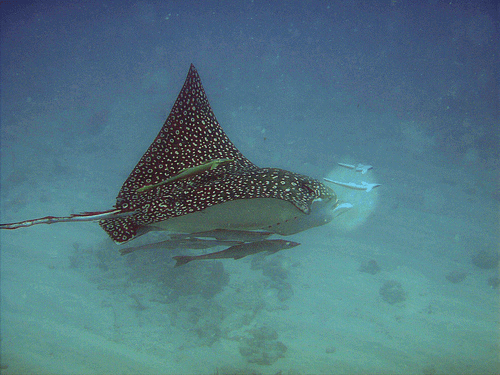}&
    \includegraphics[width=.24\textwidth]{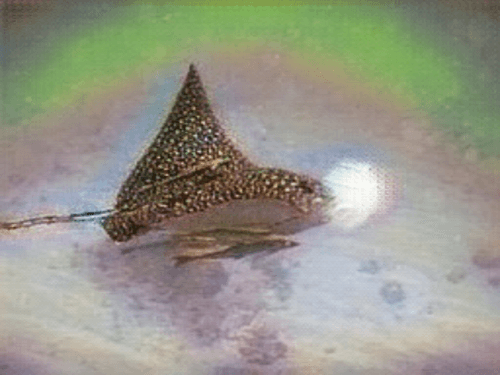}&
    \includegraphics[width=.24\textwidth]{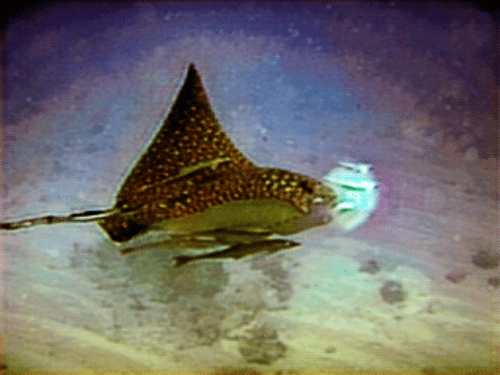}\\
     MCycleGAN~\cite{lu2019MCycleGAN} &UWCNN (type-I)~\cite{UWCNN2018} & UWGAN~\cite{UWGAN2018} & DenseGAN~\cite{DenseGAN} \\
    \\
    \includegraphics[width=.24\textwidth]{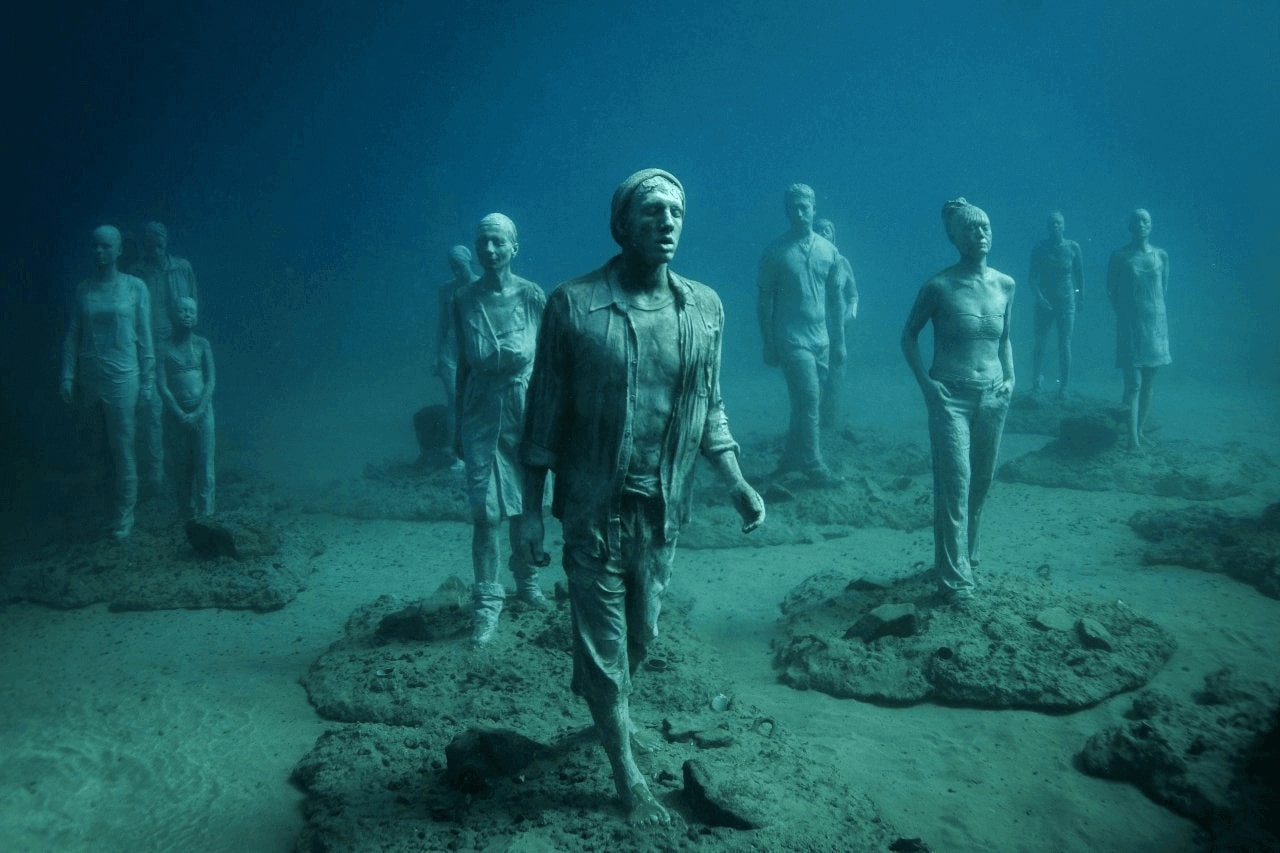}&
    \includegraphics[width=.24\textwidth]{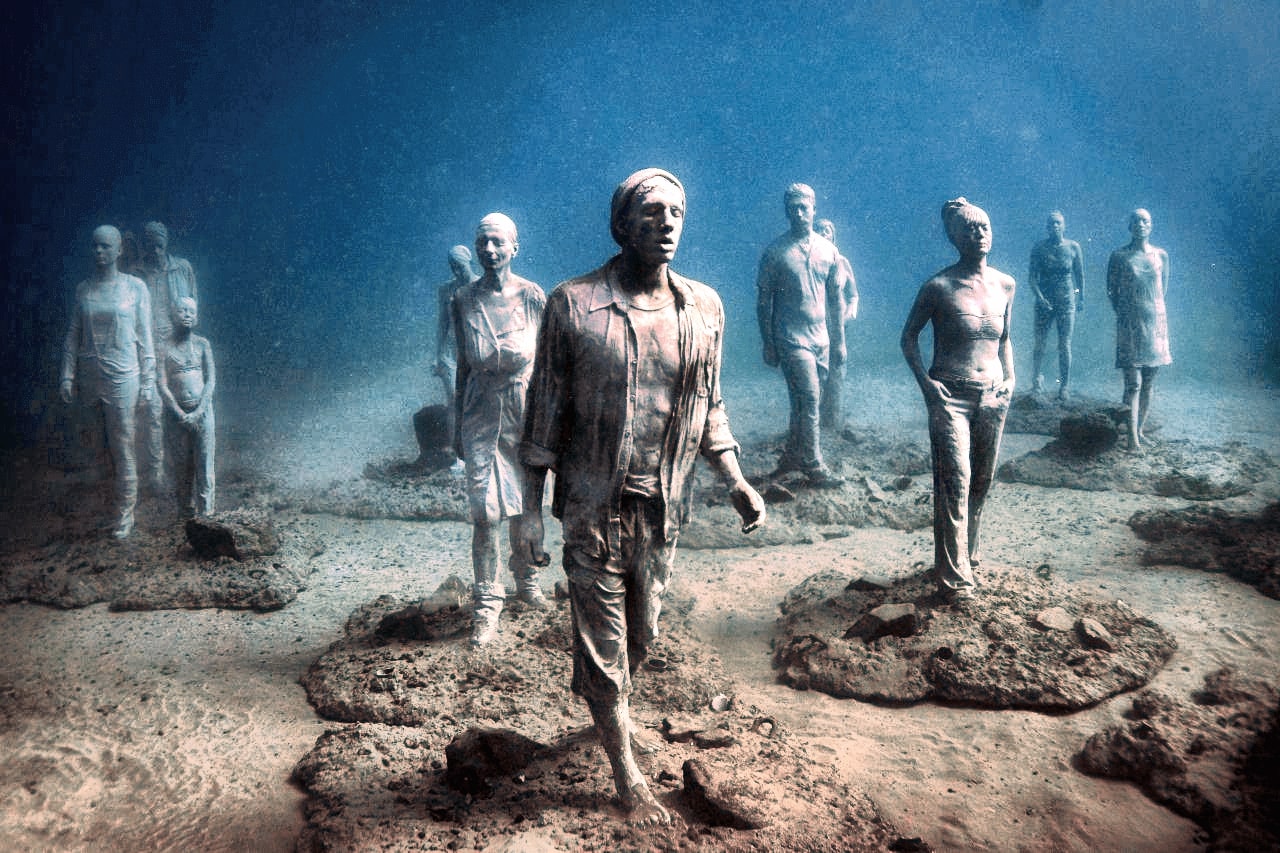}&
    \includegraphics[width=.24\textwidth]{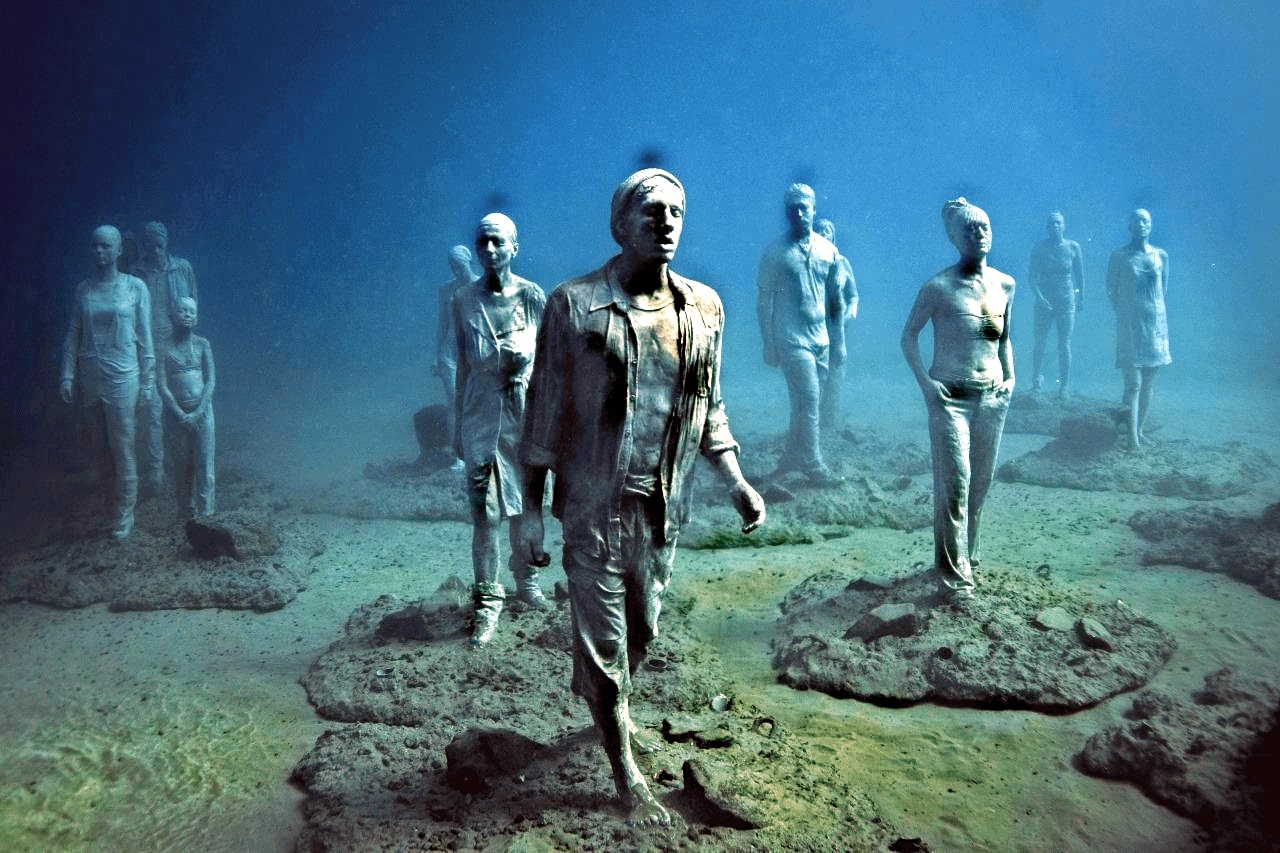}&
    \includegraphics[width=.24\textwidth]{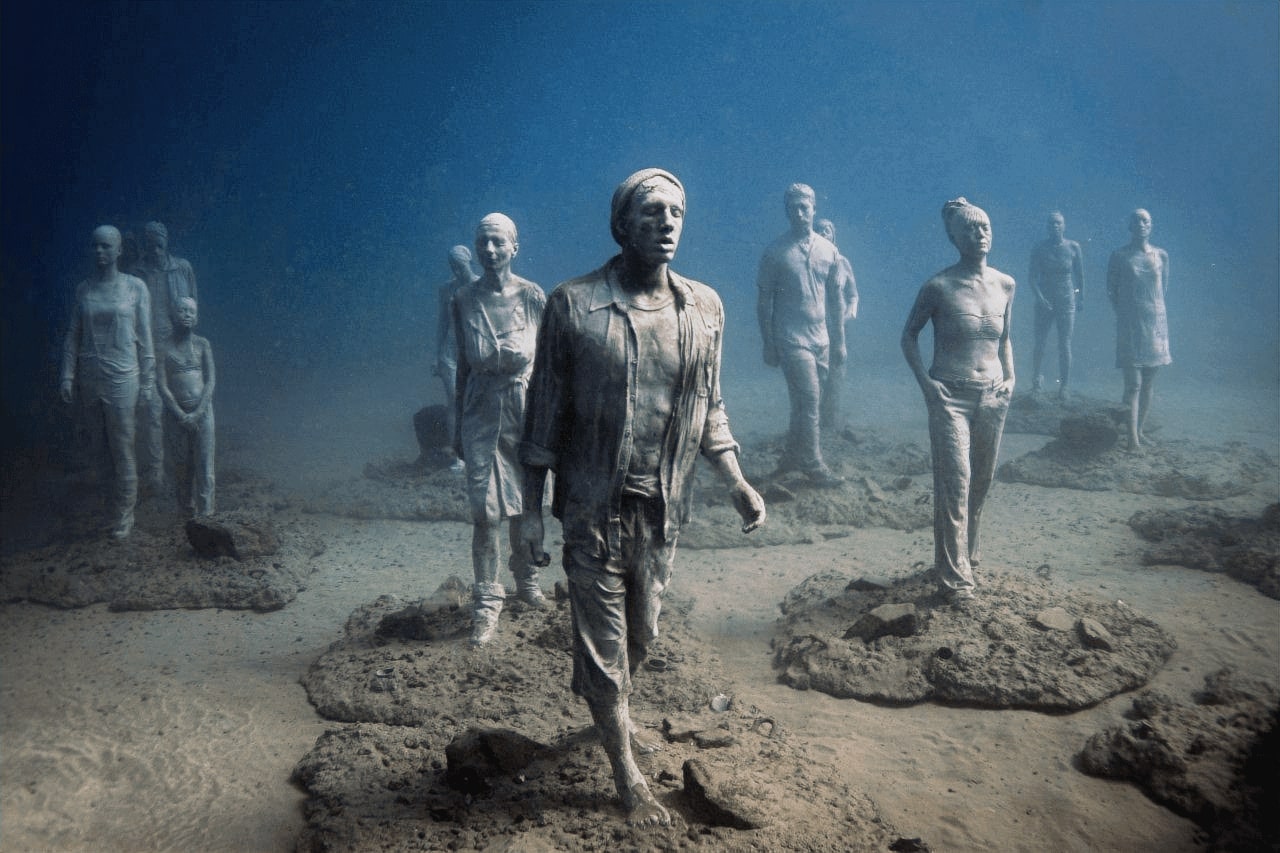}\\
    Underwater& Reference & URCNN~\cite{hou2018URCNN} & DUIENet~\cite{libenchmark2019}\\
    \includegraphics[width=.24\textwidth]{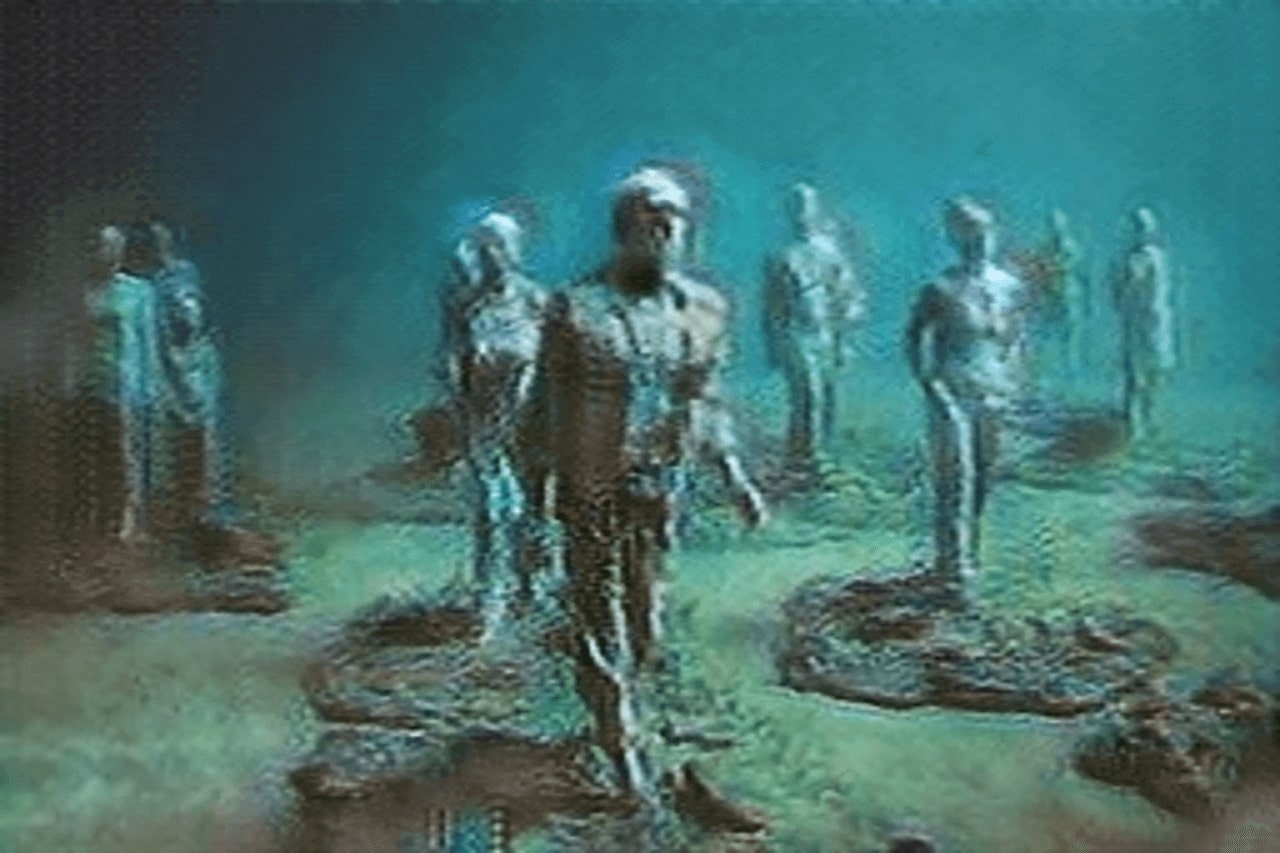}&
    \includegraphics[width=.24\textwidth]{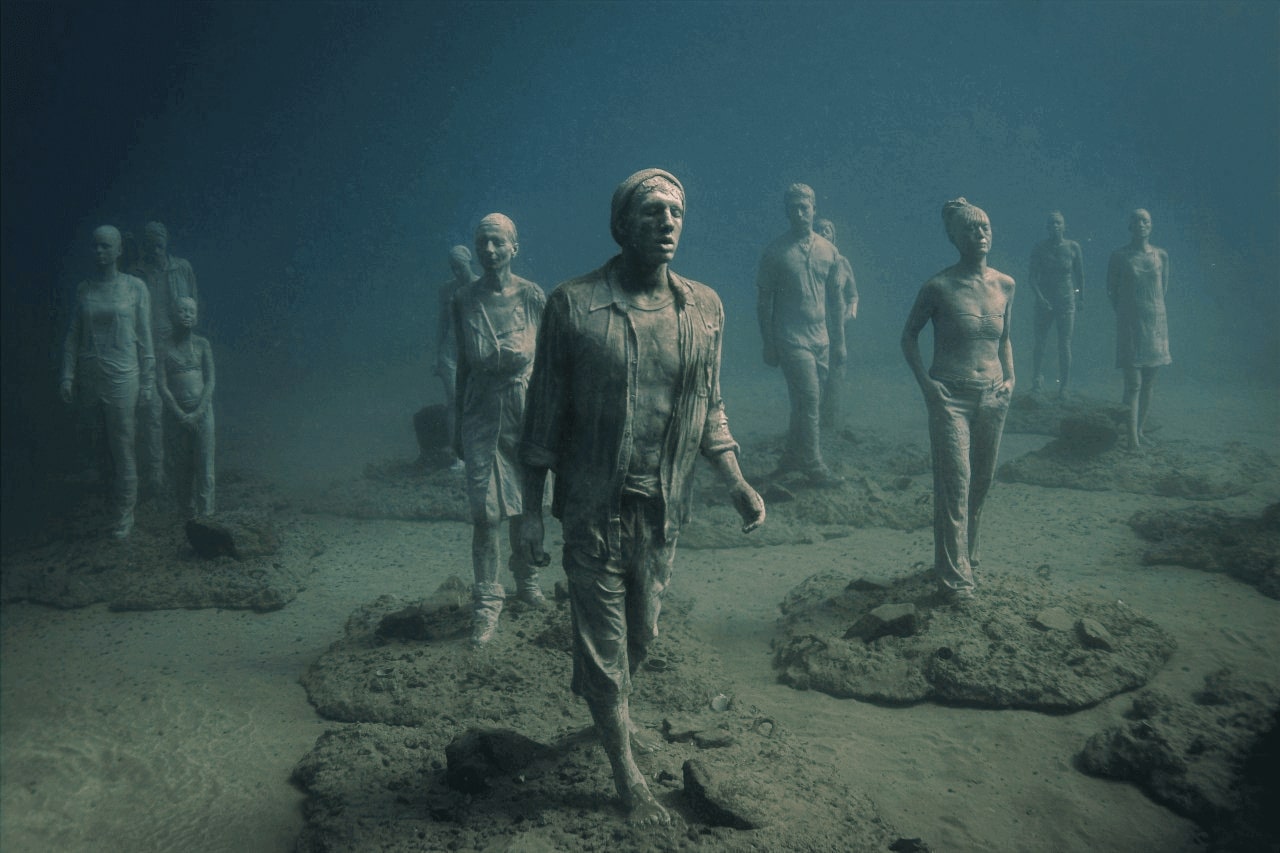}&
    \includegraphics[width=.24\textwidth]{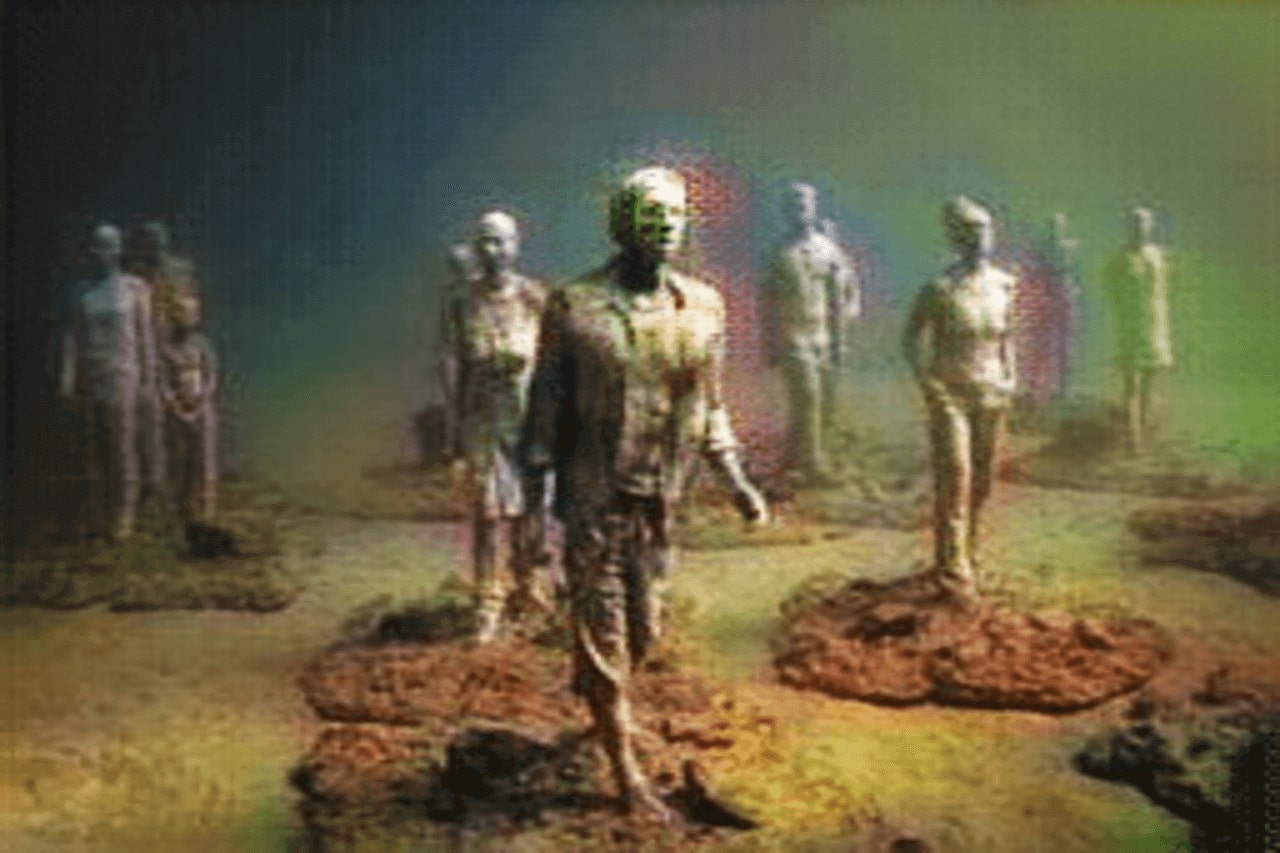}&
    \includegraphics[width=.24\textwidth]{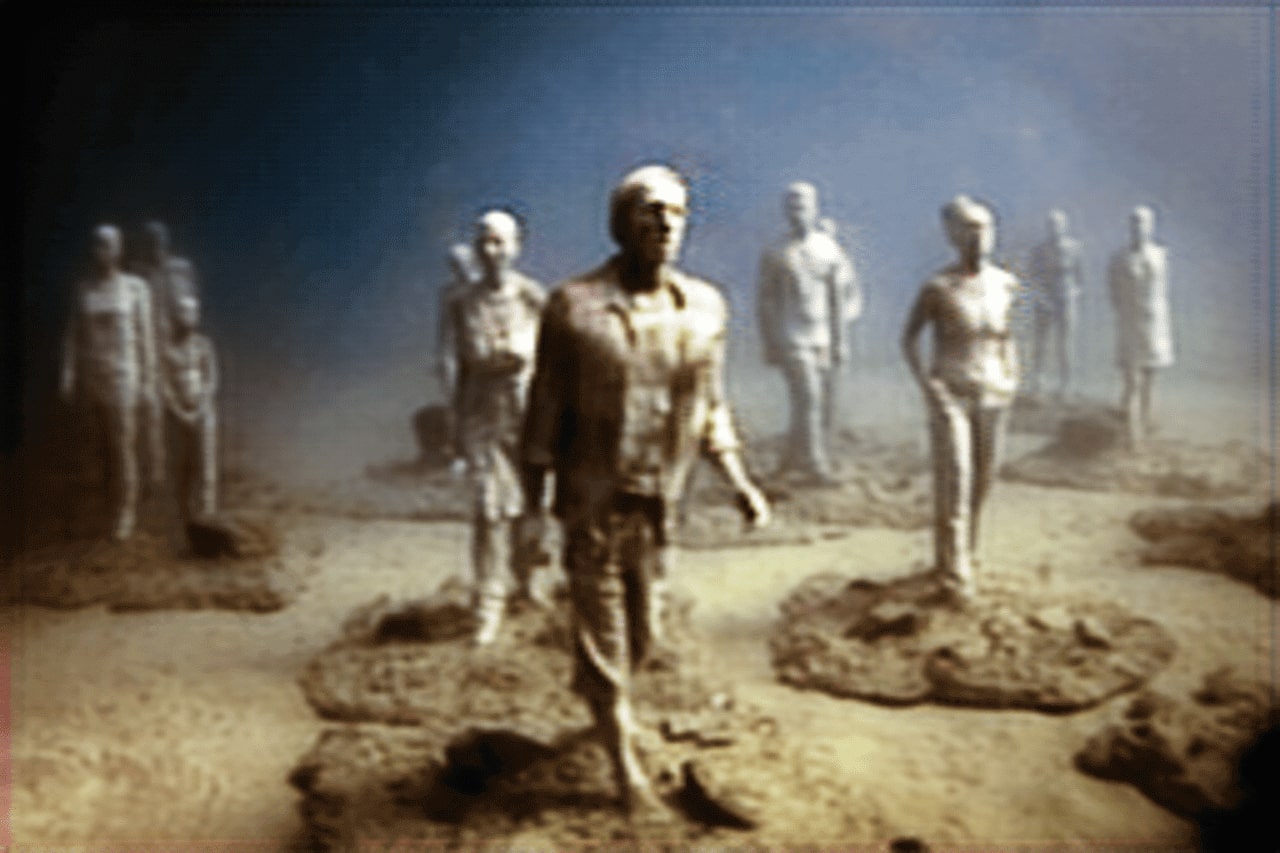}\\
    MCycleGAN~\cite{lu2019MCycleGAN} &UWCNN (type-I)~\cite{UWCNN2018} & UWGAN~\cite{UWGAN2018} & DenseGAN~\cite{DenseGAN} \\
 \end{tabular}
\end{center}
\caption{\textbf{Qualitative comparisons on bluish images:} The results of various CNN-based and GAN-based methods on the sample underwater images from UIEBD \cite{libenchmark2019}.}
\label{fig:im_bluish}
\end{figure*}

\begin{figure*}[t]
\begin{center}
\begin{tabular}{c@{ }c@{ }c@{ }c}
%\begin{tabular}{c@{ }c@{ }c@{ }c@{ }c@{ }c@{ }c@{ }c}
    \includegraphics[width=.24\textwidth]{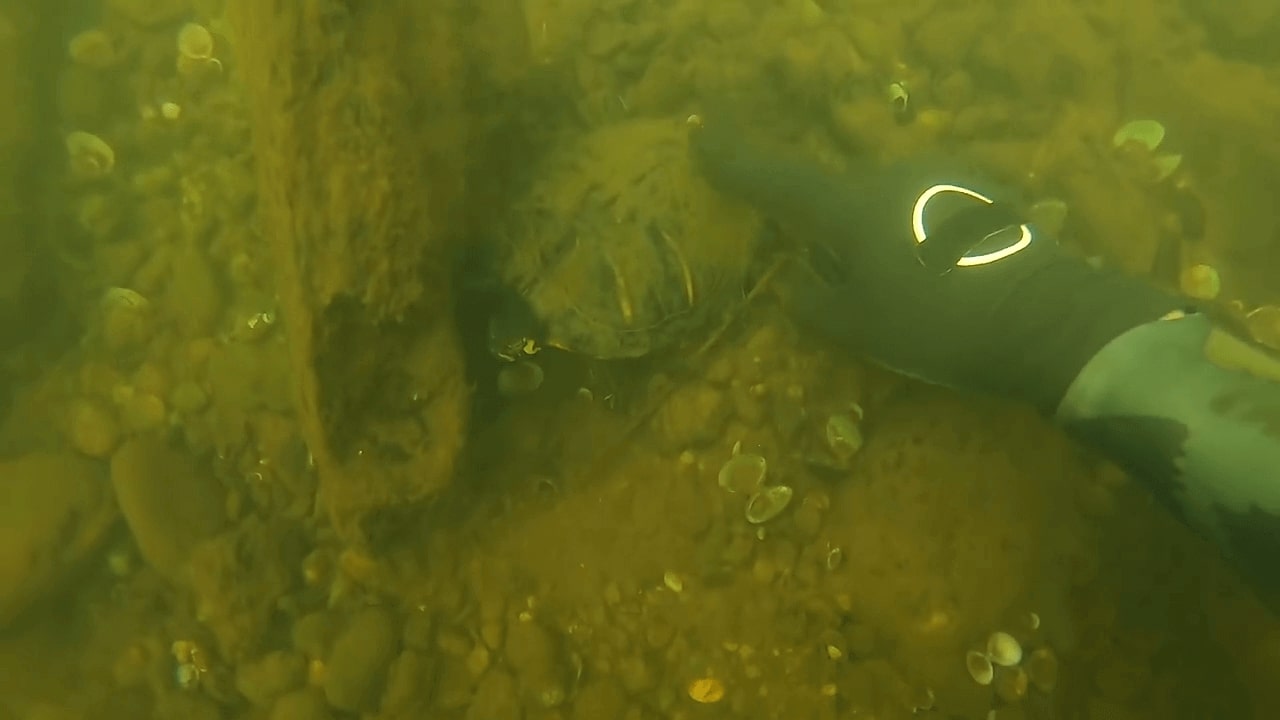}&
    \includegraphics[width=.24\textwidth]{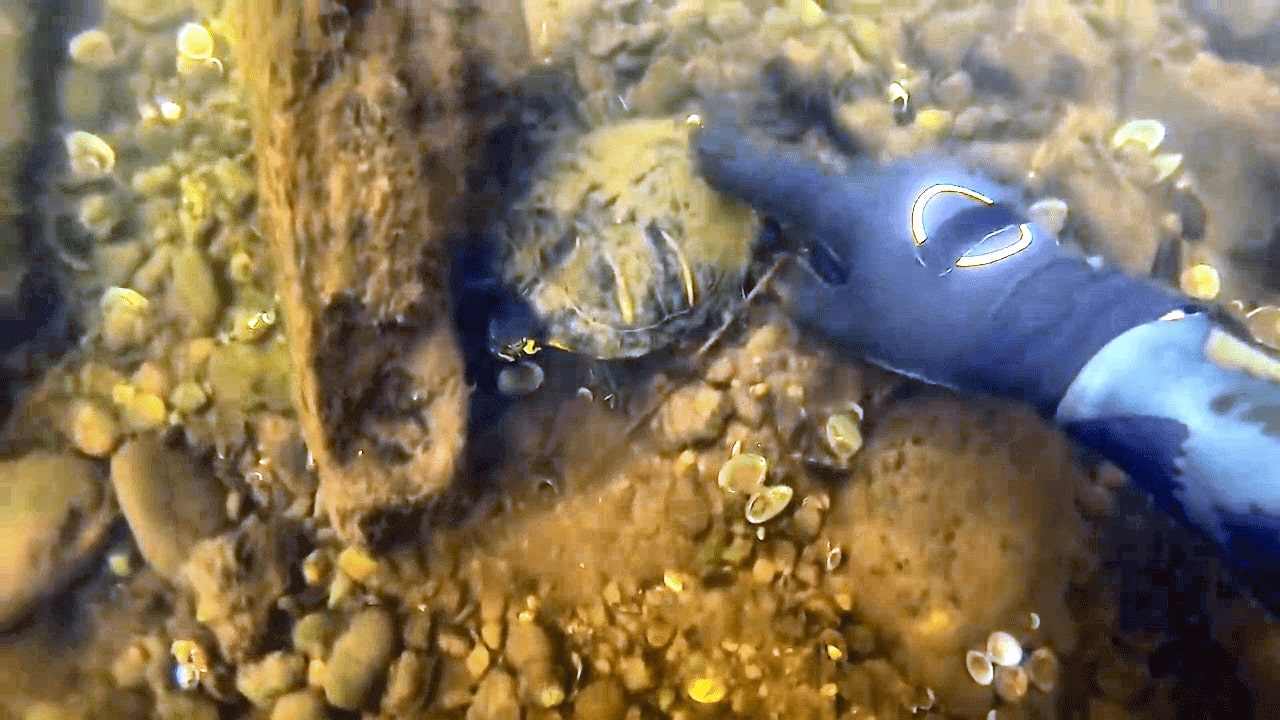}&
    \includegraphics[width=.24\textwidth]{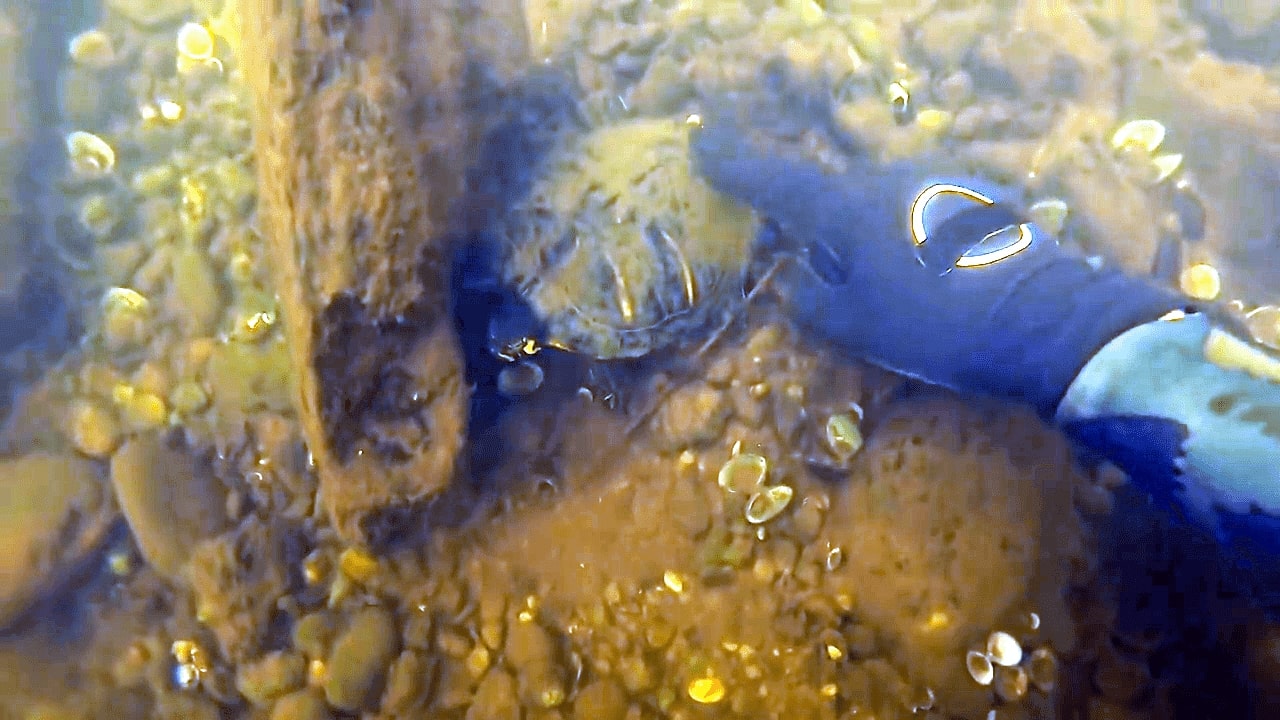}&
    \includegraphics[width=.24\textwidth]{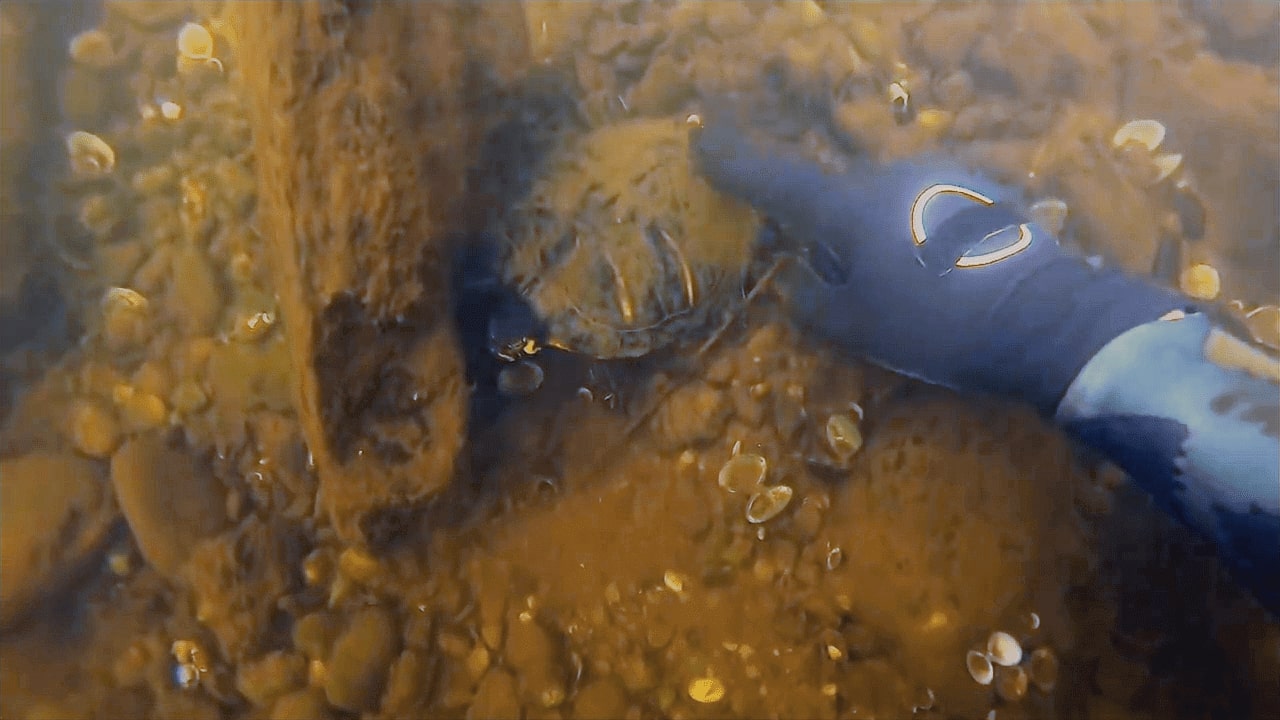}\\
    Underwater& Reference & URCNN~\cite{hou2018URCNN} & DUIENet~\cite{libenchmark2019}\\
    \includegraphics[width=.24\textwidth]{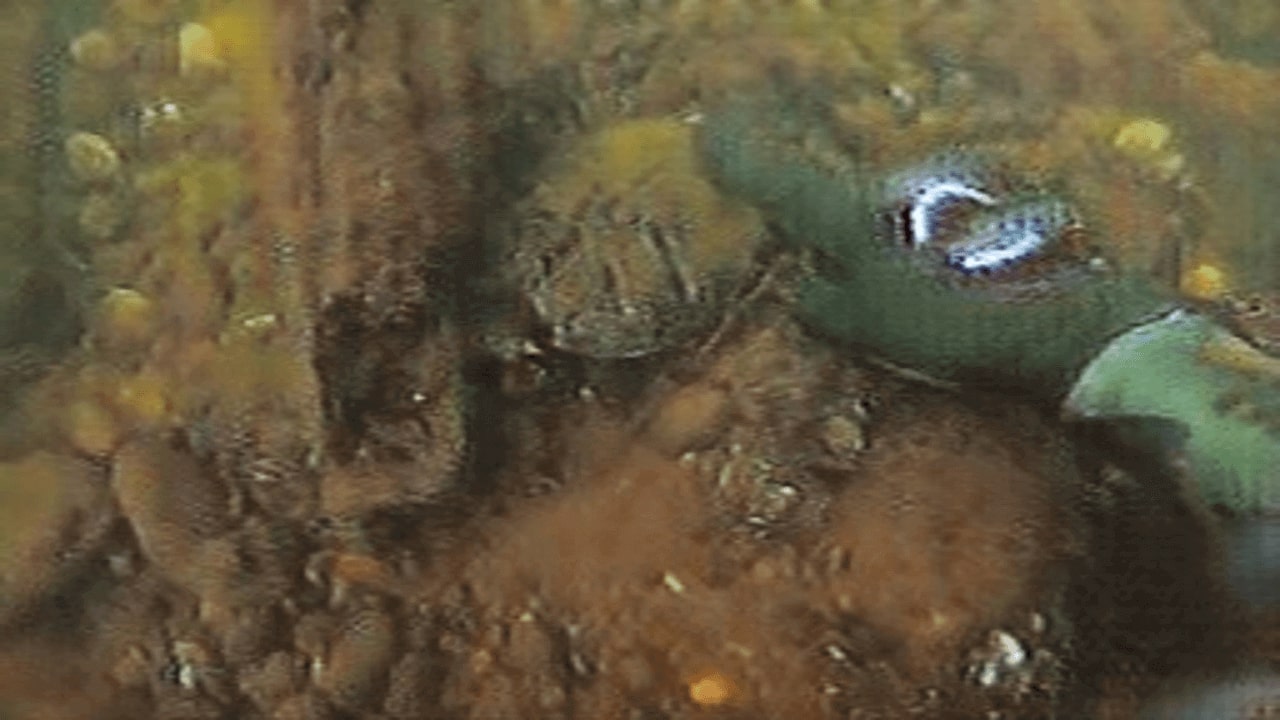}&
    \includegraphics[width=.24\textwidth]{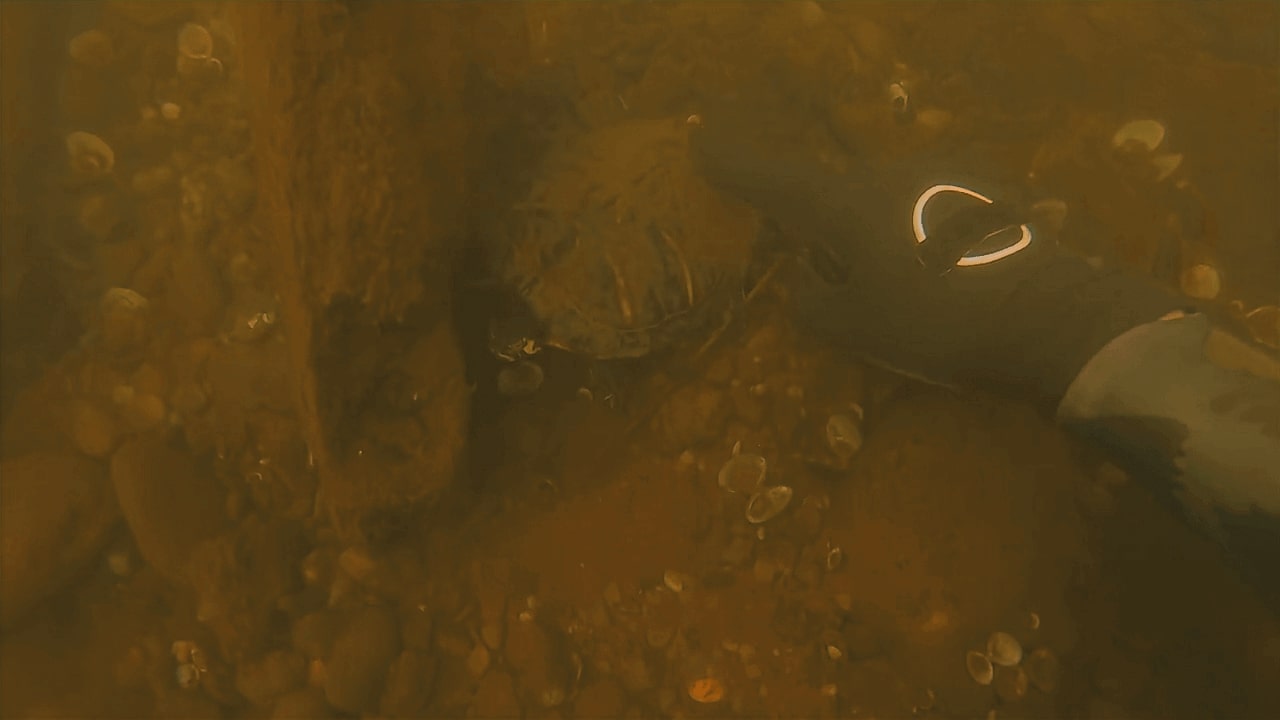}&
    \includegraphics[width=.24\textwidth]{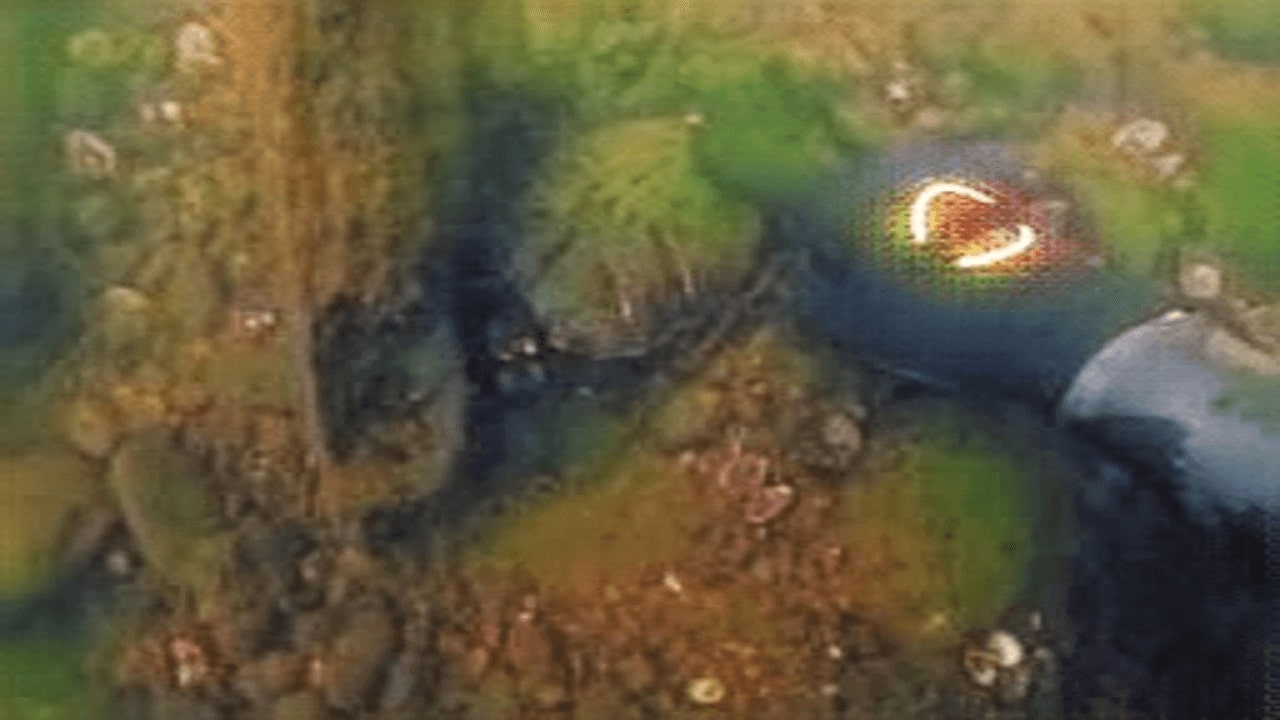}&
    \includegraphics[width=.24\textwidth]{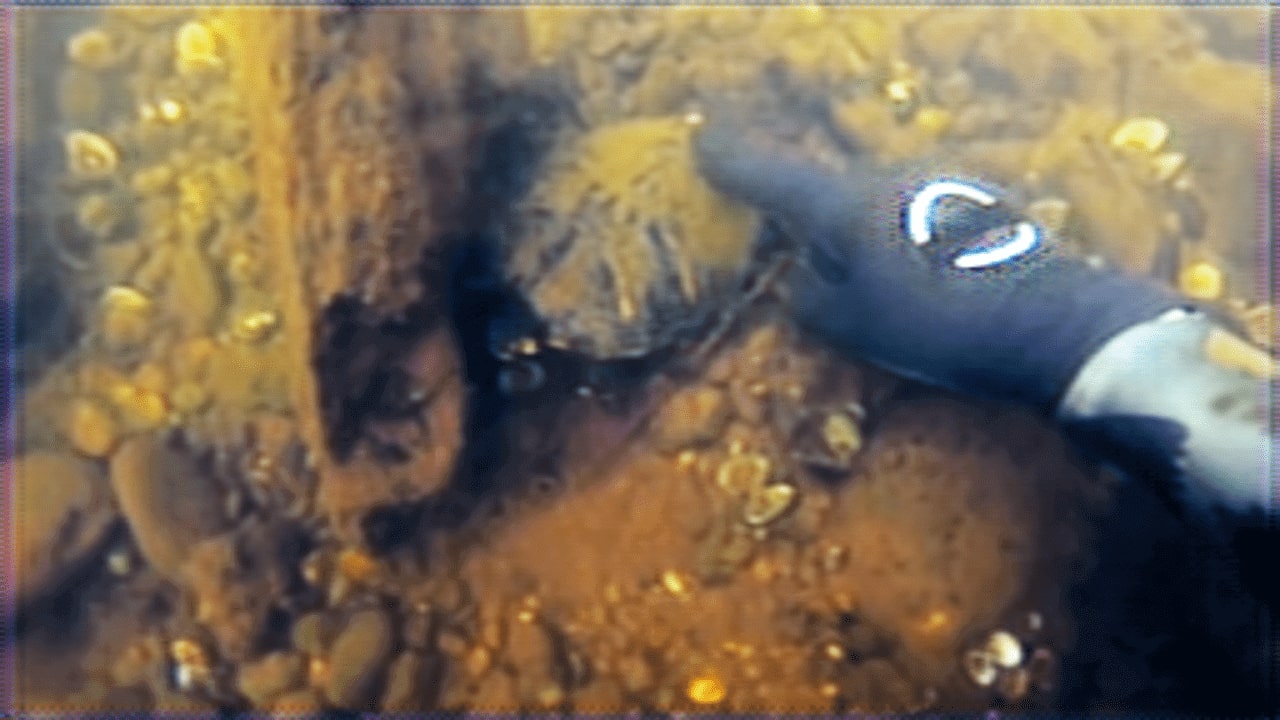}\\
     MCycleGAN~\cite{lu2019MCycleGAN} &UWCNN (type-I)~\cite{UWCNN2018} & UWGAN~\cite{UWGAN2018} & DenseGAN~\cite{DenseGAN} \\
    \\
    \includegraphics[width=.24\textwidth]{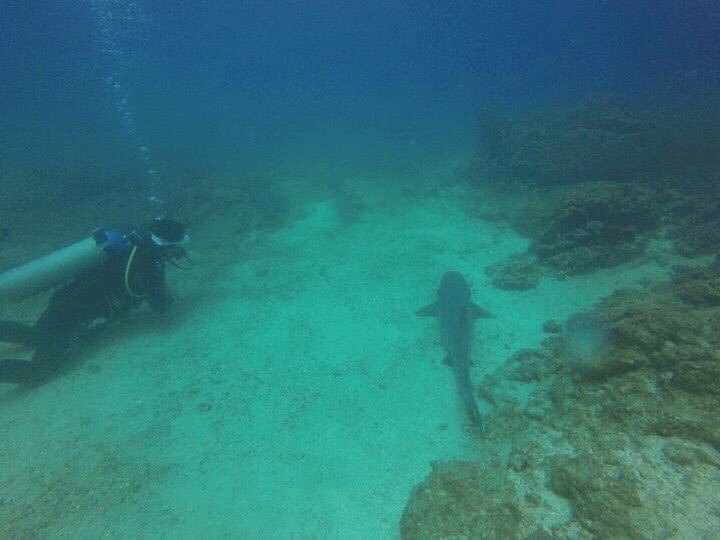}&
    \includegraphics[width=.24\textwidth]{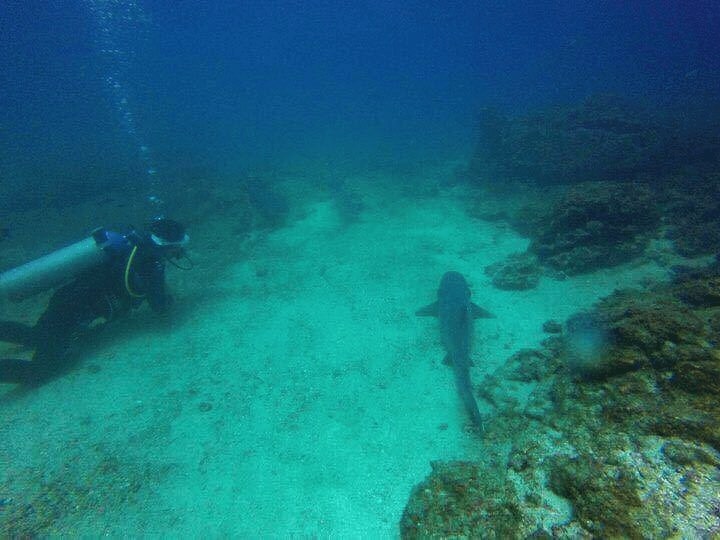}&
    \includegraphics[width=.24\textwidth]{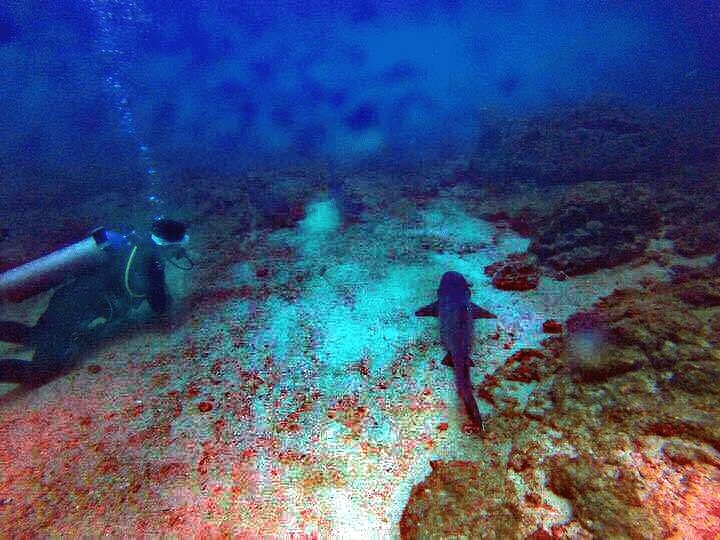}&
    \includegraphics[width=.24\textwidth]{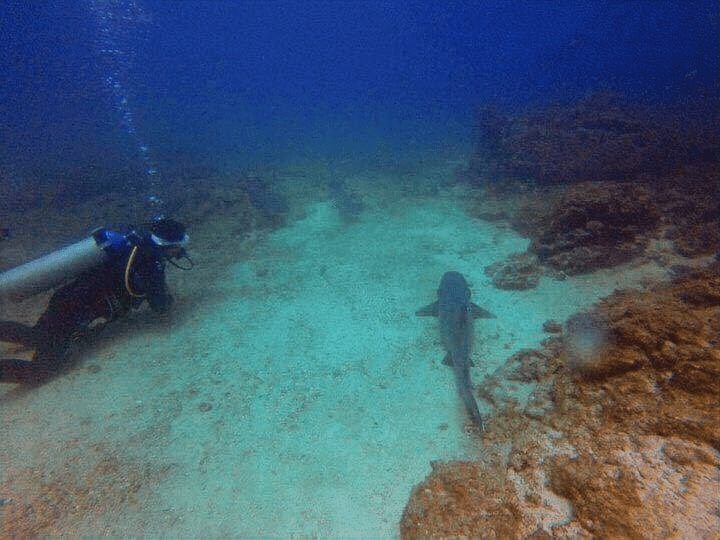}\\
    Underwater& Reference & URCNN~\cite{hou2018URCNN} & DUIENet~\cite{libenchmark2019}\\
    \includegraphics[width=.24\textwidth]{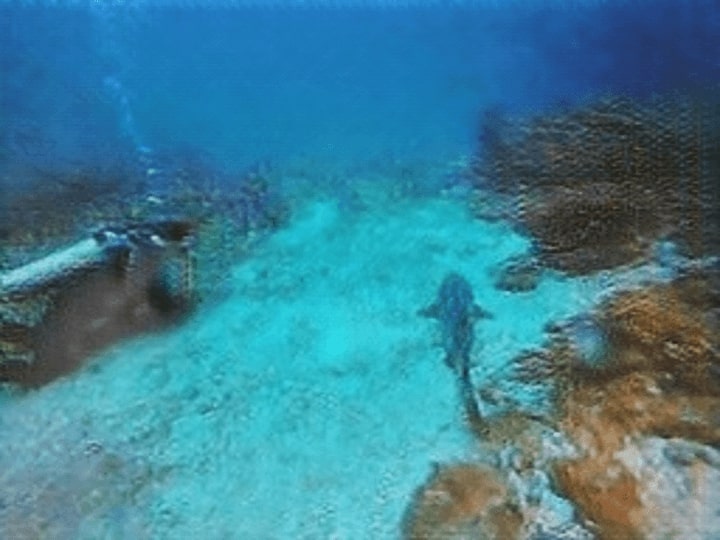}&
    \includegraphics[width=.24\textwidth]{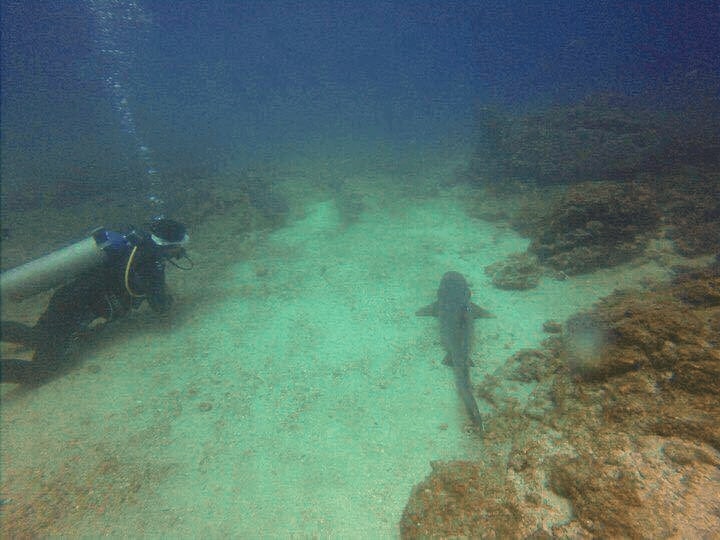}&
    \includegraphics[width=.24\textwidth]{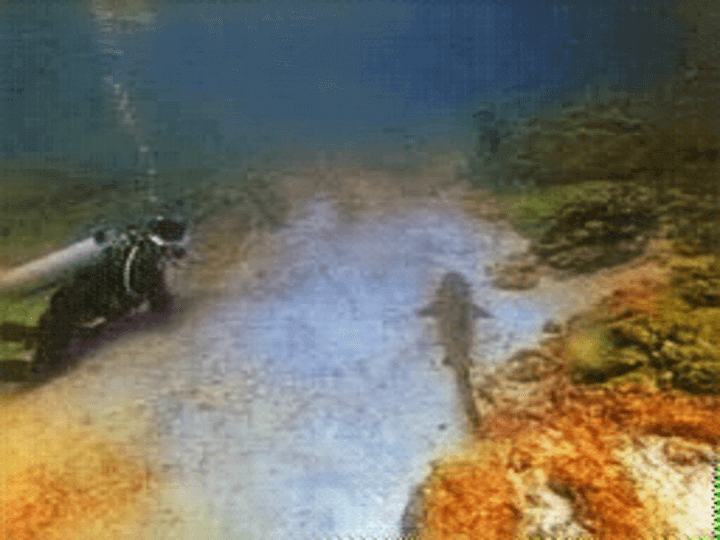}&
    \includegraphics[width=.24\textwidth]{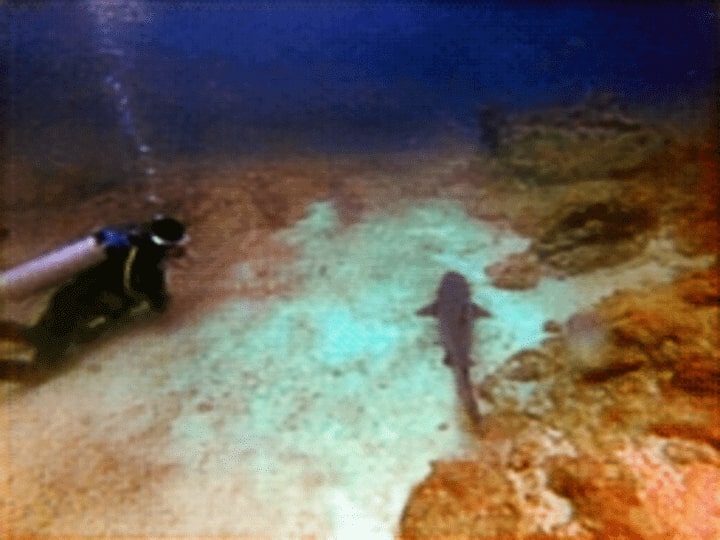}\\
   MCycleGAN~\cite{lu2019MCycleGAN} &UWCNN (type-I)~\cite{UWCNN2018} & UWGAN~\cite{UWGAN2018} & DenseGAN~\cite{DenseGAN} \\
 \end{tabular}
\end{center}
\caption{\textbf{The low and high backscatter images:} The challenging images to remove the backscatter. The images are selected from UIEBD~\cite{libenchmark2019} dataset. The top image shows the low backscatter, while the bottom image illustrates the high backscatter.}
\label{fig:im_low_high_backscatter}
\end{figure*}

\begin{figure*}[t]
\begin{center}
\begin{tabular}{c@{ }c@{ }c@{ }c}
%\begin{tabular}{c@{ }c@{ }c@{ }c@{ }c@{ }c@{ }c@{ }c}
    \includegraphics[width=.24\textwidth]{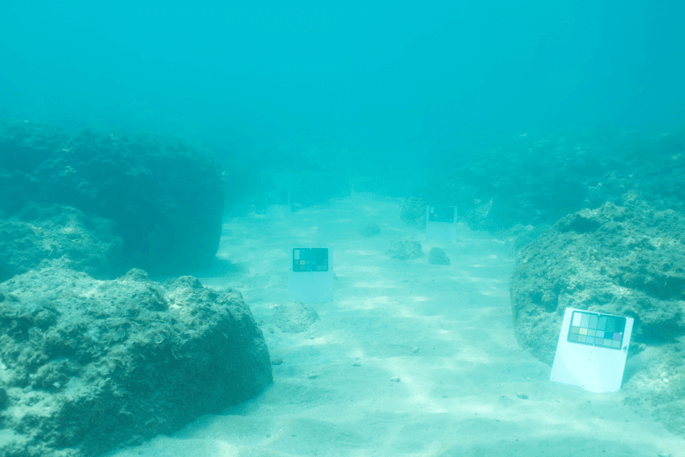}&
    \includegraphics[width=.24\textwidth]{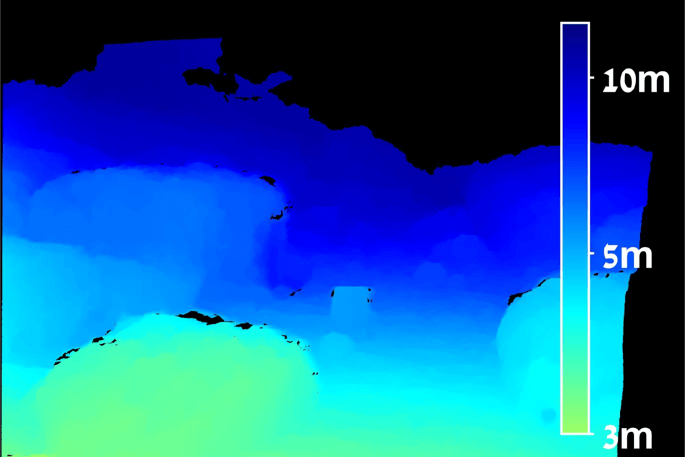}&
    \includegraphics[width=.24\textwidth]{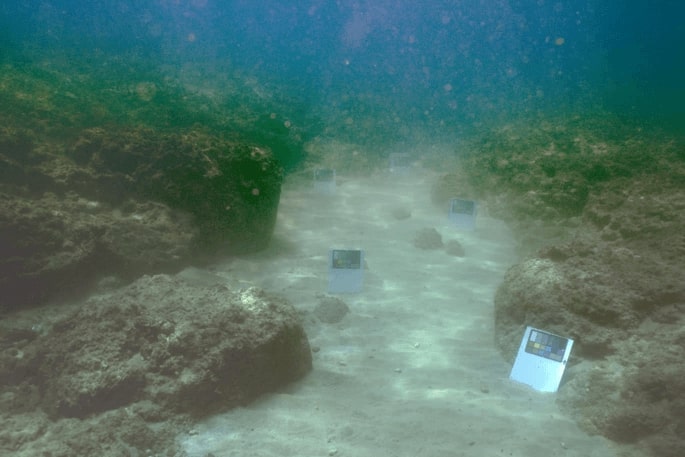}&
    \includegraphics[width=.24\textwidth]{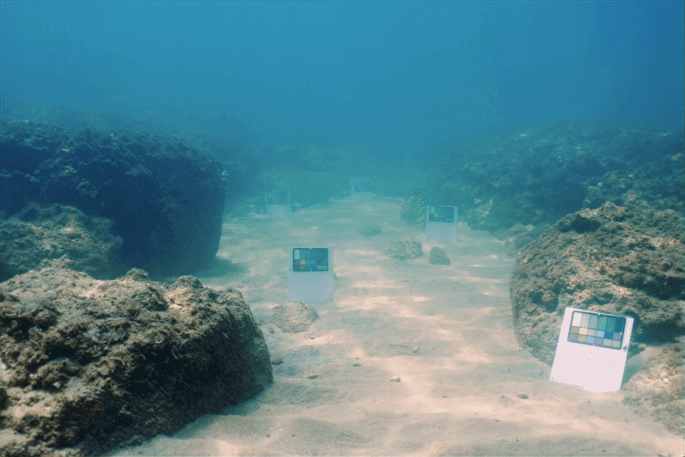}\\
    Underwater  & Distance & Haze-line~\cite{haze-line} & DUIENet~\cite{libenchmark2019} \\
    
    \includegraphics[width=.24\textwidth]{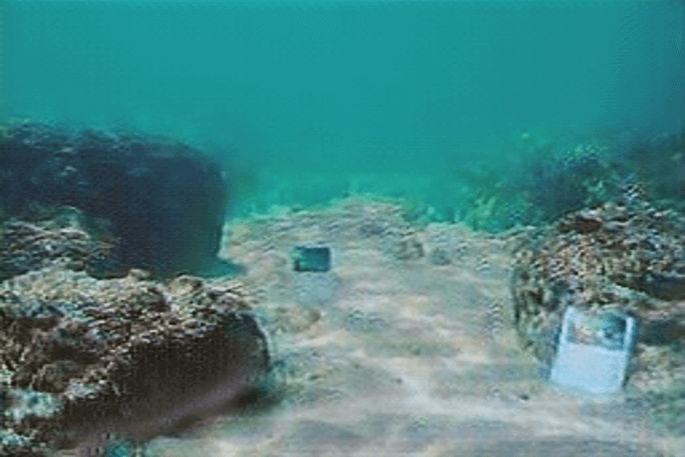}&
    \includegraphics[width=.24\textwidth]{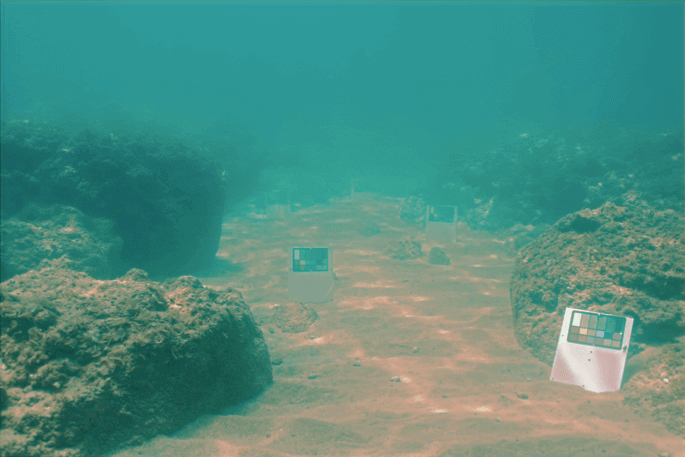}&
    \includegraphics[width=.24\textwidth]{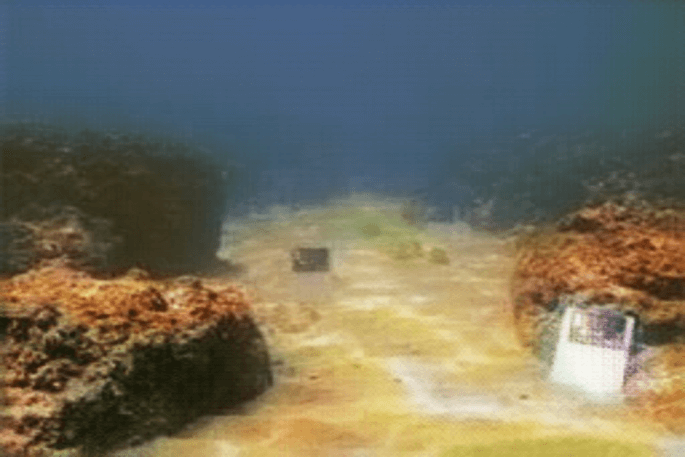}&
    \includegraphics[width=.24\textwidth]{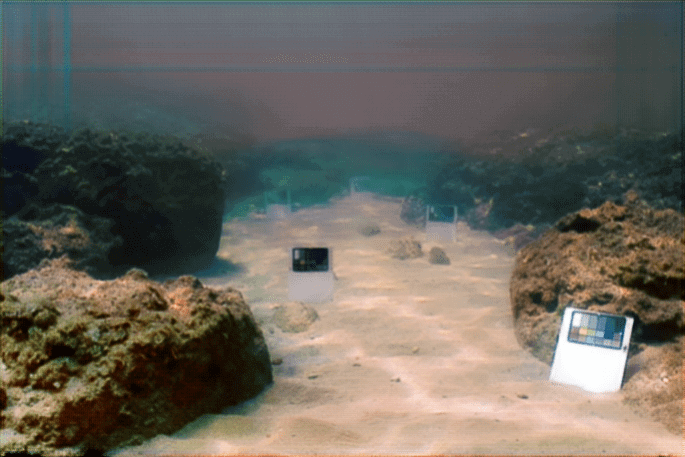}\\
    MCycleGAN~\cite{lu2019MCycleGAN} & UWCNN (type-I)~\cite{UWCNN2018} & UWGAN~\cite{UWGAN2018} & DenseGAN~\cite{DenseGAN} \\
    
    \includegraphics[width=.24\textwidth]{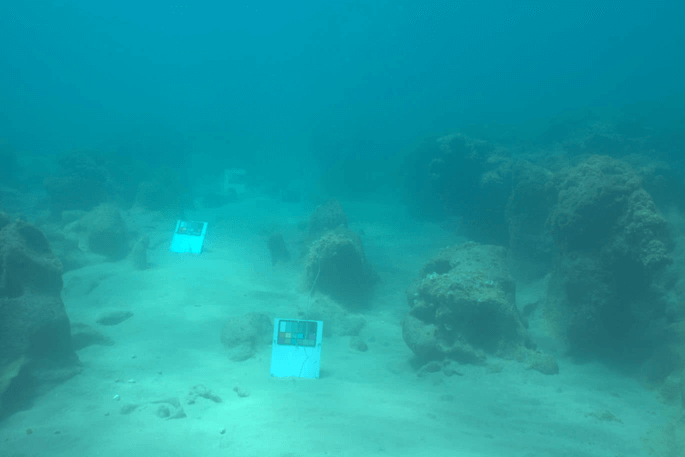}&
    \includegraphics[width=.24\textwidth]{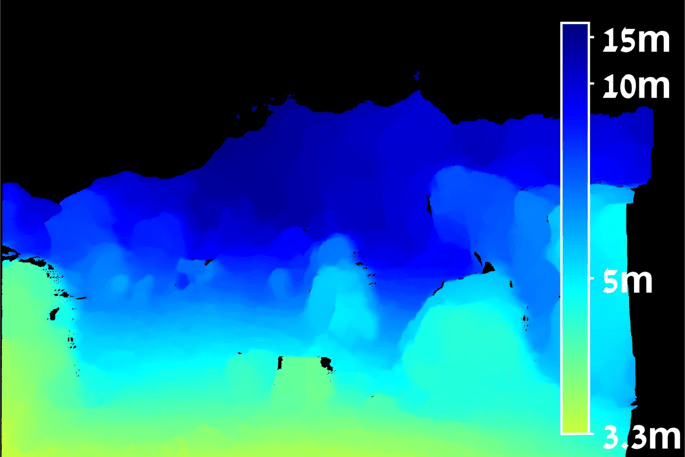}&
    \includegraphics[width=.24\textwidth]{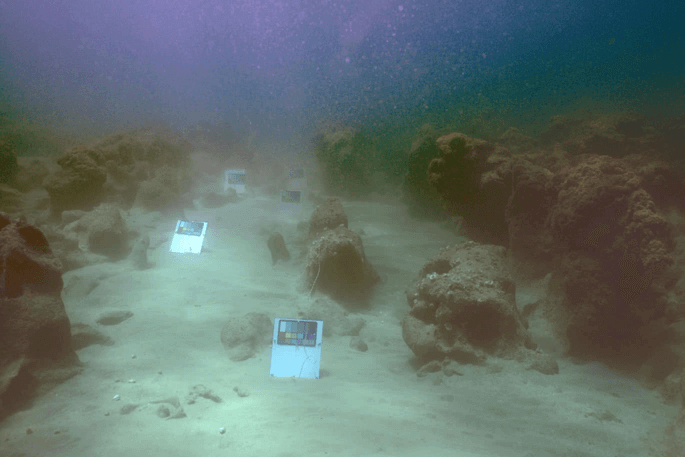}&
    \includegraphics[width=.24\textwidth]{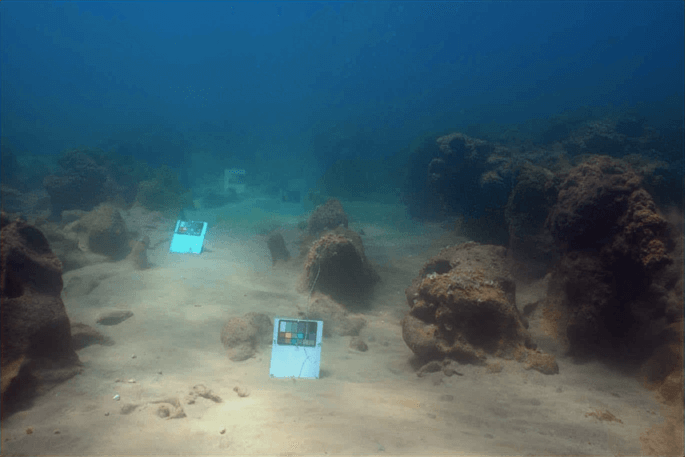}\\
    Underwater  & Distance & Haze-line~\cite{haze-line} & DUIENet~\cite{libenchmark2019} \\
    
    \includegraphics[width=.24\textwidth]{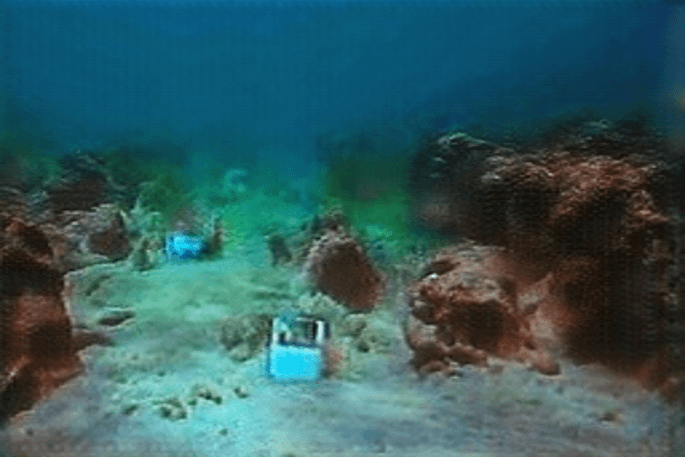}&
    \includegraphics[width=.24\textwidth]{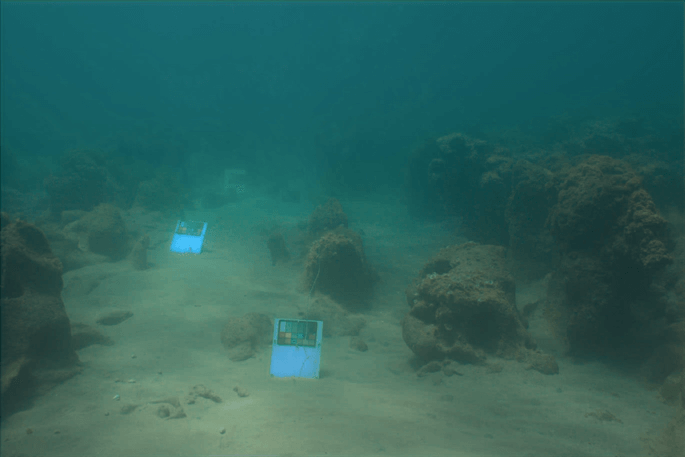}&
    \includegraphics[width=.24\textwidth]{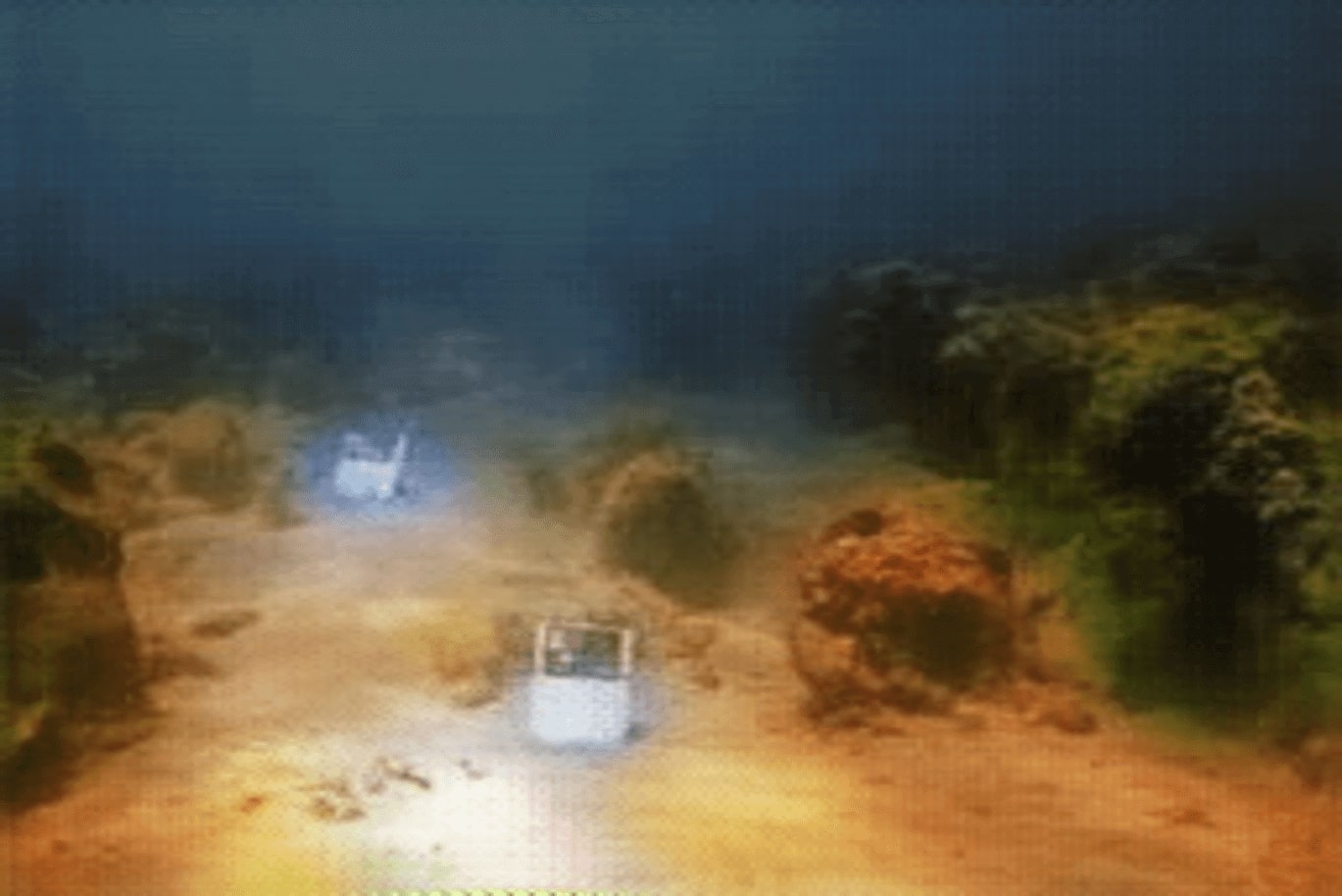}&
    \includegraphics[width=.24\textwidth]{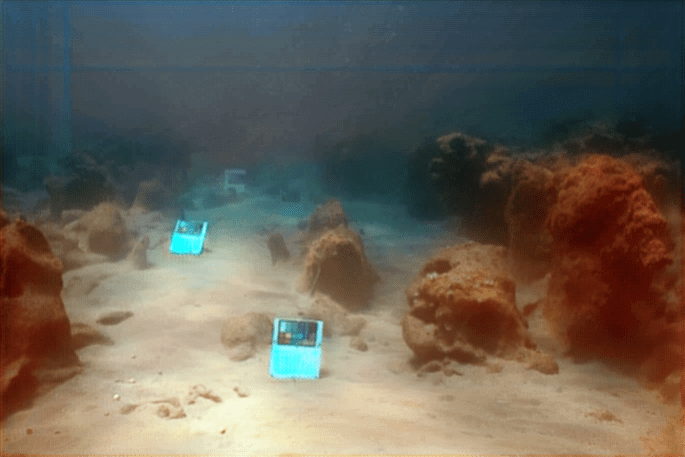}\\
    MCycleGAN~\cite{lu2019MCycleGAN} & UWCNN (type-I)~\cite{UWCNN2018} & UWGAN~\cite{UWGAN2018} & DenseGAN~\cite{DenseGAN} \\
 \end{tabular}
\end{center}
\caption{\textbf{Visual comparisons on Haze-line~\cite{haze-line}:} The Haze-line dataset provides an accurate distance based on the stereo. To be fair to the authors of Haze-line~\cite{haze-line}, we have also included the results of the best performer (\ie, Haze-line~\cite{haze-line}, a conventional method) on this dataset.}
\label{fig:im_HazeLine}
\end{figure*}

\begin{figure*}[t]
\begin{center}
\begin{tabular}{c@{ }c@{ }c@{ }c}
%\begin{tabular}{c@{ }c@{ }c@{ }c@{ }c@{ }c@{ }c@{ }c}
    \raisebox{5\normalbaselineskip}[0pt][0pt]{\multirow{3}{*}{\includegraphics[width=.24\textwidth]{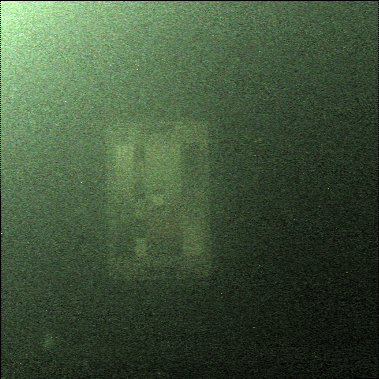}}}&
    \includegraphics[width=.24\textwidth]{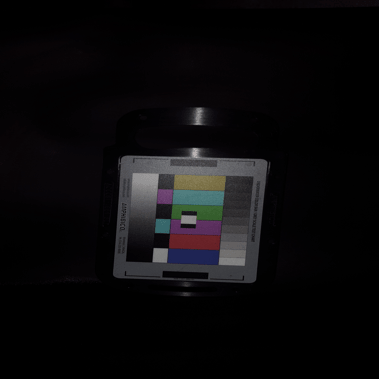}&    \includegraphics[width=.24\textwidth]{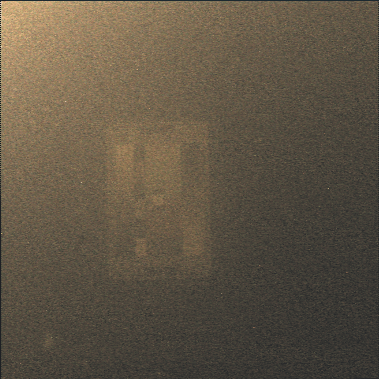}&
    \includegraphics[width=.24\textwidth]{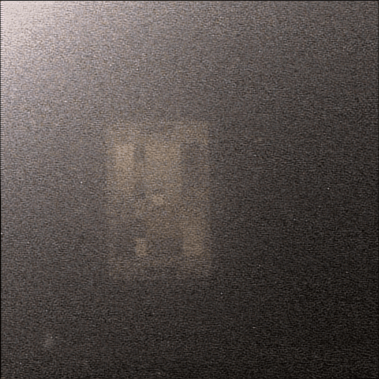}\\
    & In-air  & UWCNN (type-I)~\cite{UWCNN2018} & DUIENet~\cite{libenchmark2019} \\
    &
    \includegraphics[width=.24\textwidth]{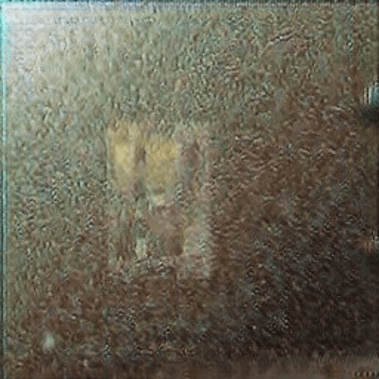}&
    \includegraphics[width=.24\textwidth]{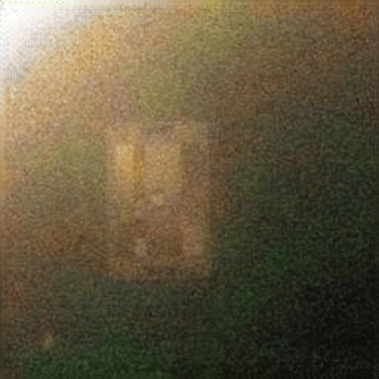}&
    \includegraphics[width=.24\textwidth]{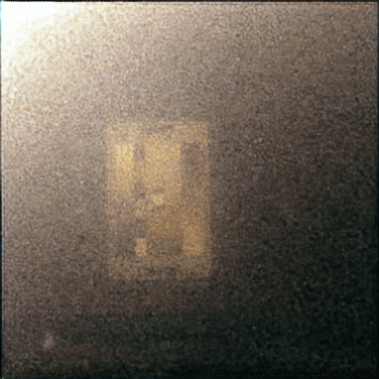}\\
     \raisebox{6.8\normalbaselineskip}[0pt][0pt]{Underwater} & MCycleGAN~\cite{lu2019MCycleGAN} & UWGAN~\cite{UWGAN2018} & DenseGAN~\cite{DenseGAN} \\
    \\
    \raisebox{5\normalbaselineskip}[0pt][0pt]{\multirow{3}{*}{\includegraphics[width=.24\textwidth]{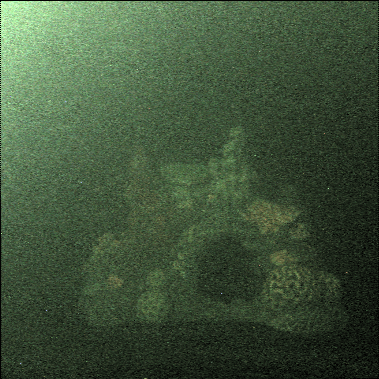}}}&
    \includegraphics[width=.24\textwidth]{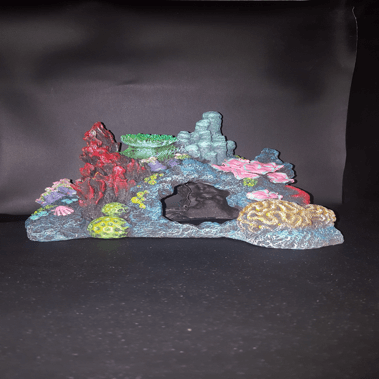}&
    \includegraphics[width=.24\textwidth]{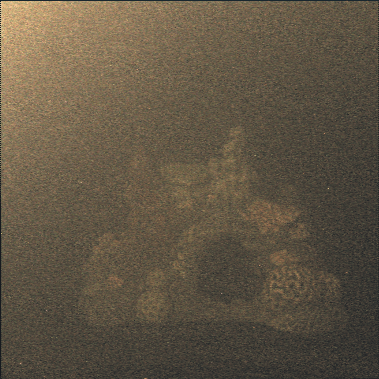}&
    \includegraphics[width=.24\textwidth]{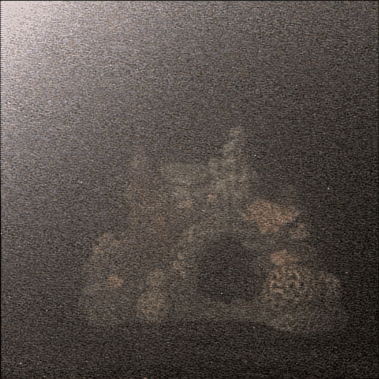}\\
    & In-air &  UWCNN (type-I)~\cite{UWCNN2018} & DUIENet~\cite{libenchmark2019} \\
    &
    \includegraphics[width=.24\textwidth]{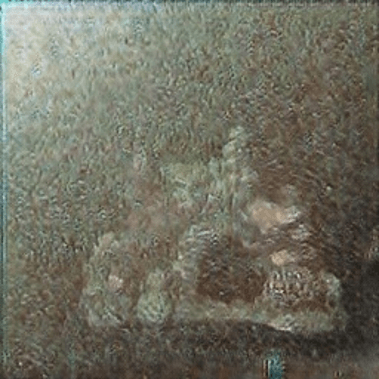}&
    \includegraphics[width=.24\textwidth]{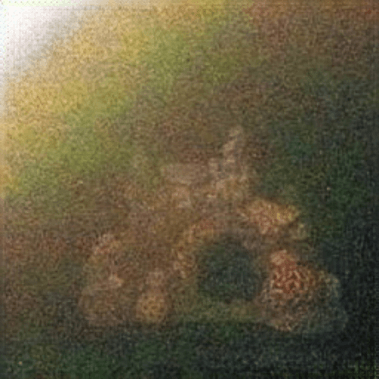}&
    \includegraphics[width=.24\textwidth]{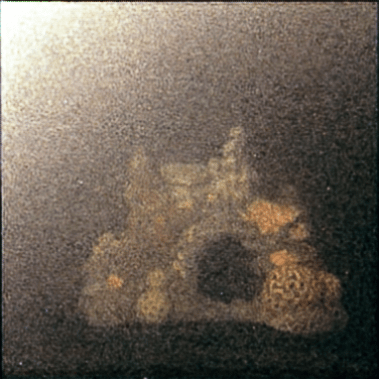}\\
     \raisebox{6.8\normalbaselineskip}[0pt][0pt]{Underwater}  & MCycleGAN~\cite{lu2019MCycleGAN} & UWGAN~\cite{UWGAN2018} & DenseGAN~\cite{DenseGAN} \\
 \end{tabular}
\end{center}
\caption{\textbf{Images from ULFID~\cite{ULFID}:} A challenging dataset where all the methods fail to provide clean results.}
\label{fig:im_ULFID}
\end{figure*}

%%%%%%%%%%%%%%%%%%%%%%%%%%%%%%%%%%%%%%%%%% Results %%%%%%%%%%%%%%%%%%%%%%%%%%%%%%%%%%%%%

%\subsection{Number of Parameters}
\subsection{Benchmark Results}
The benchmark results for each technique\footnote[8]{The results are reported for the methods having the source code or executables available or the respected authors agreed to provide the results on the dataset.} on UIEBD~\cite{libenchmark2019} dataset are reported in Table~\ref{table:UWE_Metric}. The quantitative experiments are conducted on UIEBD~\cite{libenchmark2019} because it is, to the best of our knowledge, the only one dataset which provides the corresponding reference images for image quality assessment. The results by using reference images can provide realistic feedback on the quality of enhanced results to some extent. Moreover,  in case of multiple variants of the same algorithm, all the results are reported. We encourage the readers to consult the original paper for a detailed analysis of each variant of the same model.

The results are presented via the metrics mentioned earlier. It is to be noted here that the PSNR, SSIM, PCQI, UCIQE, and UIQM, the higher, the better while the MSE, the lower, the better. Also, to be fair amidst all the methods under consideration, we resize the output of the network where the predicted image is a scaled-down version of the underwater scene input. From Table~\ref{table:UWE_Metric}, DUIENet~\cite{libenchmark2019} results are the best among the competitors while the UWCNN~\cite{UWCNN2018} performs worst due to training on the synthesized underwater images which are different from the images in the UIEBD~\cite{libenchmark2019}. However, it is challenging to state the superiority of one method against the others due to many factors involved, for example, the number of parameters, the depth of network, training images, patch size, number of channels and loss function, \etc To compare fairly, most of these determinants should be kept consistent. To further validate the performance of different deep algorithms, we conduct qualitative comparisons on diverse underwater images from different datasets in the next section.

\subsection{Qualitative Comparisons}
We present the visual results on UIEBD~\cite{libenchmark2019}, Haze-line~\cite{haze-line} and ULFID~\cite{ULFID} in Figures~\ref{fig:im_greenish}-\ref{fig:im_ULFID}. The ground-truth images for Haze-line~\cite{haze-line} and ULFID~\cite{ULFID} are not available; hence, we furnish the visual results only for both the datasets.  

\begin{itemize}
   \itemsep1em 
   \item \textbf{Greenish tone images:}  In Figure~\ref{fig:im_greenish}, we present the visual comparisons of greenish underwater images from UIEBD~\cite{libenchmark2019} for the state-of-the-art CNN-based and GAN-based methods. The GAN-based models aim to improve the perceptual quality, while CNN models are more focused on the PSNR values of the enhanced images. One can notice that the outputs of GAN methods are generally different in the tone as compared to CNN methods, as the later is more faithful to the original underwater image colors. This also contributes to the higher PSNR for the CNN methods compared to GAN methods, as shown in Table~\ref{table:UWE_Metric}. It is to be noted that in Figure~\ref{fig:im_greenish}, we only show one of the variants in case of the same algorithm for the limited space. 

     \item \textbf{Bluish tone images:} Figure~\ref{fig:im_bluish} shows the visual comparisons on  two bluish images from UIEBD~\cite{libenchmark2019} consisting of a ray and statues. The bluish tone is ubiquitous in underwater images and difficult to be completely removed by current algorithms.  DUIENet~\cite{libenchmark2019} and UWCNN~\cite{UWCNN2018} render the best outcomes; however, the results still have a bluish tone, especially in far distances (more severe backscatter). By contrast, the UWGAN~\cite{UWGAN2018} and DenseGAN~\cite{DenseGAN} introduces obvious artificial colors mainly inducing by the shortcomings of their unpaired training data.

    \item \textbf{Low and high backscatter images:} Backscatter is a challenging problem faced during the underwater imaginary. The leading causes of backscattering are the strobes or the internal flash, which lights up the particles in the water present between the subject and the camera lens. This phenomenon can also be observed behind the subject, lighting up the open water. With a dark background, backscattering is more natural to recognize. Here, we present two images in Figure~\ref{fig:im_low_high_backscatter} on low and high backscatter from~\cite{libenchmark2019}. The first image in Figure~\ref{fig:im_low_high_backscatter} is an example of low backscatter, while the bottom one is of high backscatter. We can visually observe that the URCNN~\cite{hou2018URCNN} has over-exposed the images while the UWGAN~\cite{UWGAN2018} created some artificial colors. In addition, the low backscatter is relatively easier to be removed than the high backscatter. For the high backscatter image, none of the methods can produce visually pleasing results and current methods even introduce annoying artifacts and color casts. It should also be regarded here that UWCNN~\cite{UWCNN2018} can produce good results if the model matches the type of water.  

     \item \textbf{Haze-line~\cite{haze-line} images:} The visual comparisons for underwater images from Haze-line dataset~\cite{haze-line} is provided in Figure~\ref{fig:im_HazeLine}. This dataset only provides the depth maps reconstructed from the stereo images; however, no ground-truth images are available for computing the evaluation metrics. The images in this dataset are challenging since most of the images have bluish tone and high backscatter. UWGAN~\cite{UWGAN2018} and DenseGAN~\cite{DenseGAN} provide visually promising results, but both have created false colors, and this is also the case with DUIENet~\cite{libenchmark2019} and  MCycleGAN~\cite{lu2019MCycleGAN} networks. It is obvious that all deep algorithms fall behind the performance of a conventional method~\cite{haze-line} which mismatches the progress of deep learning in other low-level visual tasks.

    \item \textbf{ULFID~\cite{ULFID} images:} As the last example, we show the images with severe degradations from ULFID~\cite{ULFID} in Figure~\ref{fig:im_ULFID}. The ground-truth images for this dataset are not feasible to evaluate the models; hence, we only present the visual results. Although the deep algorithms can remove the greenish tone from the images; however, all of them fail to furnish clear images and even amplify the noise. This dataset is an excellent example that the underwater image enhancement still requires concerted efforts to progress, and the noise in underwater images should be paid more attention in the future study.
    
\end{itemize}
\section{Future and Emerging Directions}
Underwater image enhancement is a classical research area and has improved a lot in recent years, mainly due to the rapid development of deep learning techniques. The performance is still lacking in many aspects when compared to other image enhancement techniques like image super-resolution, deblurring, and dehazing.  There is ample room to advancement the underwater image enhancement direction. Here, in the following paragraphs, we present the list of some of the potential future directions.
\begin{itemize}
    \itemsep1em 
    \item \textbf{Datasets:}
    Underwater image enhancement methods usually employ synthetic images for training due to lack of representative real-world underwater images and its corresponding ground-truth images. Although there are limited datasets available, which have underwater and their reference images; however, these datasets consist of a finite number of images and are typically used as test images rather than training the models. A true effort in this direction may improve the performance of underwater image enhancement models and also provide realistic feedback on the image quality of enhanced results by different methods.

    \item \textbf{Objective functions and evaluation metrics:}
    Current algorithms predominantly employ objective functions common to image enhancement techniques. Although these functions produce some favorable results; however, none of them incorporate the underwater physical model properties. Likewise, the available evaluation metrics to underwater images are limited and have failure cases, which keeps the field of underwater image enhancement at a standstill. For example, the visual results shown in Figures~\ref{fig:im_greenish}-\ref{fig:im_ULFID} do not match the quantitative results in Table~\ref{table:UWE_Metric}.  Therefore, more specialized objective functions and evaluation metrics are required to advance the underwater image enhancement research.

    \item \textbf{Prior knowledge:}
    The human perception of the scene depends on the extensive domain or prior knowledge. When experts describe the image quality, they don't solely rely on the content of the visuals; instead, they also use their domain knowledge. An exciting venue to explore is to augment the current techniques with prior or domain knowledge~\cite{wu2016ask}. This has shown an increase in the performance in areas like visual question answering and would likely help to improve underwater image enhancement.

    \item \textbf{Unsupervised learning:}
    Due to the lack of dataset, which has underwater images and their ground-truth images, many methods generate synthetic data to train their models. Although these models exhibit promising results for synthetic underwater scenes; however, they fail on real-world underwater images. To deal with the lack of data, a possible research direction could be unsupervised learning, also known as zero-shot or few-shot learning. This capability may lead to promising results, but the zero-shot problem itself is not trivial. A more realistic scenario would be to employ the present limited datasets, few-shot learning, where the network learns from a few available images. The development of unsupervised learning is an open research problem.

    \item \textbf{Real vs. Synthetic:}
    Existing algorithms use diverse physical (mathematical) models to generate underwater images. The distribution of the generated underwater scenes may not be conferred to the real-world scenes; therefore, the models trained on artificially produced datasets lack generalization capability. A more thorough and exhaustive effort is required to generate artificial datasets, and one solution may be to use GAN-based networks to transfer style from underwater images to the simulated scenes. Even though minimal work~\cite{li2018watergan} has been done in this direction, still there is a lot of scope of improvement.
\end{itemize}

\section{Conclusion} 
We presented the first comprehensive literature survey on CNNs and GANs for underwater image enhancement. To the best of our knowledge, we have included all the deep learning-based methods, which deal with underwater image enhancement, including those which are available on arxiv\footnote[9]{at the time of submission}. Moreover, we provided and reviewed the datasets, which can be used for training and testing the algorithms. We also discussed the details of the evaluation metrics with their limitations. Using all the metrics, we compared the performance on the benchmark dataset. We also presented the visual comparisons to illustrate the varying difficulty and the robustness of the algorithms. As a final step, we reviewed the limitations and provided future research areas to advance the underwater image enhancement. 

The deep learning-based underwater image enhancement methods still follow the development of deep learning ranging from CNNs to GANs. Most of the current models are the modifications of existing network architectures such as encoder-decoder network and CycleGAN. The significant difference is the training data (\ie, underwater images). Besides, there is no network architecture or loss function well-designed for underwater image enhancement tasks, which results in the unstable and visually unpleasing results. In most cases, the deep learning-based methods fall behind state-of-the-art conventional methods. More importantly, almost all models use synthetic data for networks' training. The synthetic training data limit the generalization of models. Thus, the development of deep learning-based underwater image enhancement has a long way to go. 

According to our survey, the underwater research progress is hindered by the lack of purposely built evaluation metrics and large training dataset. The current metrics are taken from the image enhancement while the training datasets are synthetically generated. One approach to develop evaluation metrics is to incorporate underwater image properties. Similarly, more realistic datasets can be created using the GANs.

\bibliographystyle{spmpsci}      % mathematics and physical sciences
\bibliography{ref}

\end{document}